\documentclass[10pt]{article}
\usepackage[preprint]{style}

\PassOptionsToPackage{numbers,sort&compress}{natbib}
\setcitestyle{numbers,sort&compress,square}

\usepackage{amssymb}
\usepackage{amsthm}
\usepackage{bm}
\usepackage{mathtools}
\usepackage{parskip}
\usepackage{graphicx}
\usepackage{enumitem}
\usepackage{csquotes}
\usepackage{url}
\usepackage{booktabs}
\usepackage{algorithm}
\usepackage{algorithmic}
\usepackage{tikz-cd}
\usepackage[many,listings]{tcolorbox}
\usepackage{float}
\usepackage{listings}
\usepackage{pythonhighlight}
\usepackage{mathrsfs}
\usepackage{listings}
\usepackage[dvipsnames]{xcolor}

\definecolor{mygreen}{RGB}{10,150,60}

\usepackage[
  colorlinks=true,
  linkcolor=MidnightBlue,
  urlcolor=RoyalBlue,
  citecolor=SeaGreen
]{hyperref}
\usepackage{cleveref}

\usepackage[many]{tcolorbox}

\newtcbox{\codebox}{
  on line,
  arc=5pt,
  boxsep=0pt,
  left=3pt,
  right=3pt,
  top=2pt,
  bottom=2pt,
  colback=white,      
  colframe=gray!50,     
  boxrule=0.5pt,
  enhanced,
  fontupper=\ttfamily\small\color{NavyBlue}, 
}

\newcommand{\R}{\mathbb{R}}

\newtheorem{theorem}{Theorem}[section]

\theoremstyle{definition}

\colorlet{pygray}{black!40!white}
\colorlet{pylightgray}{black!40!white}
\definecolor{pyblue}{RGB}{126,166,224}
\definecolor{pyred}{RGB}{234,107,102}

\lstdefinestyle{pythonstyle}{
    language=Python,
    basicstyle=\ttfamily\small,
    keywordstyle=\bfseries\color{black},
    stringstyle=\color{black},
    commentstyle=\color{pygray}\itshape,
    identifierstyle=\color{black},
    morekeywords=[1]{def},
    keywordstyle=[1]{\color{teal}\bfseries},
    morekeywords=[2]{if,else,for,while,return,import,from,as,with,try,except,finally,class,True,False,None,and,or,not,in,is,lambda,yield,break,continue,pass,range,int,float,list,dict,set,str,enumerate,zip,len},
    keywordstyle=[2]{\color{teal}\bfseries},
    morekeywords=[3]{Parameter,randn,rfft,einsum,cfloat,pad,irfft,unsqueeze,sum,cat,view,cdist,expand,arange,sin,cos,backward,step,square,abs,det,stack,column_stack,pi,parameters},
    keywordstyle=[3]{\color{pyred}},
    morekeywords=[4]{MLP,spectral_conv,pointwise_layer,integral_transform,encoder_decoder,NeuralOperator,Adam,train_iteration,Delauney,compute_weights},
    keywordstyle=[4]{\color{pyblue}},
    emph=[5]{Tensor},
    emphstyle=[5]{\color{pylightgray}},
    moredelim=**[s][\color{pylightgray}]{Tensor[}{]},
    deletekeywords=[2]{sum,abs},
    showstringspaces=false,
    breaklines=true,
    tabsize=4,
    keepspaces=true,
    columns=flexible,
    xleftmargin=4pt,
    xrightmargin=4pt,
    framexleftmargin=12pt,
    framexrightmargin=12pt,
    upquote=true,
    morestring=[b]',
    morestring=[b]"
}

\renewcommand{\thealgorithm}{\arabic{algorithm}}
\newtcblisting{mypython}[2][]{
    enhanced,
    float,
    floatplacement=t,
    colback=black!5!white,
    colframe=black!15!white,
    arc=5pt,
    boxrule=0pt,
    fonttitle=\small\color{black},
    listing only,
    listing options={style=pythonstyle},
    before title={\refstepcounter{algorithm}\label{#1}},
    title={Algorithm \thealgorithm: #2},
    top=-2pt,
    bottom=-2pt,
    left=2pt,
    right=2pt,
}

\setlist[itemize]{leftmargin=*}
\setlist[enumerate]{leftmargin=*}
\lstdefinestyle{float}{
  float=tp,
  floatplacement=tbp,
}

\crefname{lstlisting}{algorithm}{algorithm}
\Crefname{lstlisting}{Algorithm}{Algorithm}

\usepackage{dsfont}

\usepackage[frozencache]{minted}

\setminted{
  frame=lines,
  framesep=3mm,
  baselinestretch=1.0,
  bgcolor=gray!4,
  fontsize=\small,
  linenos,
  numbersep=2pt,
  breaklines,
  samepage=true,
  fontfamily=tt,
  tabsize=4,
  xleftmargin=10pt,
  style=friendly
}

\title{Fourier Neural Operators Explained: A Practical Perspective}

\author{\name Valentin Duruisseaux  \\
      \addr California Institute of Technology
      \AND
      \name Jean Kossaifi \\
      \addr NVIDIA
      \AND
      \name Anima Anandkumar \\
      \addr California Institute of Technology}

\newcommand{\mydef}[1]{\textcolor[rgb]{0.4,0.4,0.8}{\textbf{#1}}}

\begin{document}

\maketitle

\hfill  \\ 

\hfill 

\begin{abstract}
Partial differential equations (PDEs) govern a wide variety of dynamical processes in science and engineering, yet obtaining their numerical solutions often requires high-resolution discretizations and repeated evaluations of complex operators, leading to substantial computational costs. Neural operators have recently emerged as a powerful framework for learning mappings between function spaces directly from data, enabling efficient surrogate models for PDE systems. Among these architectures, the Fourier Neural Operator (FNO) has become the most influential and widely adopted due to its elegant spectral formulation, which captures global correlations through learnable transformations in Fourier space while remaining invariant to discretization and resolution. Despite their success, the practical use of FNOs is often hindered by an incomplete understanding among practitioners of their theoretical foundations, practical constraints, and implementation details, which can lead to their incorrect or unreliable application. This work presents a comprehensive and practice-oriented guide to FNOs, unifying their mathematical principles with implementation strategies. We provide an intuitive exposition of the concepts of operator theory and signal processing that underlie the FNO, detail its spectral parameterization and the computational design of all its components, and address common misunderstandings encountered in the literature. The exposition is closely integrated with the \texttt{NeuralOperator 2.0.0} library, offering modular state-of-the-art implementations that faithfully reflect the theory. By connecting rigorous foundations with practical insight, this guide aims to establish a clear and reliable framework for applying FNOs effectively across diverse scientific and engineering fields.
\end{abstract}

\clearpage 

\tableofcontents

\clearpage
\section{Introduction}

\vspace{2mm}

Many phenomena and dynamical systems of interest in science and engineering can be modeled as nonlinear partial differential equations (PDEs). Since closed-form solutions to these PDEs are typically not known, simulating the associated dynamics requires the use of numerical integration and time-stepping schemes which can become very expensive computationally as the desired simulation resolution increases. Indeed, to capture small-scale structures, fast-varying dynamics, and steep gradients, traditional numerical integrators must discretize space and time into extremely fine grids, which makes them prohibitively slow and resource-intensive.

Researchers have increasingly turned to machine learning to accelerate scientific simulations by replacing traditional numerical integrators with data-driven surrogates. Motivated by the remarkable success of deep learning approaches in numerous domains such as natural language processing, speech and audio processing, and computer vision, neural networks have been widely employed across diverse scientific fields to simulate dynamical systems. Once trained, these models can predict the evolutions of new dynamical systems directly from input data without explicitly solving the governing PDEs. By avoiding the need to discretize the domain finely and solve every equation from first principles, they can achieve substantial speedups over numerical solvers. 

Despite their advantages over traditional numerical solvers, neural networks face challenges in scientific computing. They are inherently designed to operate on finite-dimensional vectors such as pixel arrays or measurement vectors, which limits their ability to generalize across different discretizations and resolutions. Many natural phenomena, however, are more accurately described by continuous functions governed by PDEs. Examples include systems in fluid dynamics, thermodynamics, continuum mechanics, and electromagnetism, where physical quantities evolve continuously in space and time. In such cases, predicting the future state of a system, such as the evolution of a velocity field, involves mapping an input function (e.g. representing the initial condition) to the PDE solution function.

To overcome these limitations, machine learning approaches have been developed to learn solution operators of PDEs directly from available data~\citep{kovachki2023neural}, providing a principled framework for modeling continuum systems. Among these approaches, neural operators~\citep{azizzadenesheli2024neural, Berner2025Principles} have gained considerable attention due to their ability to process input functions provided with arbitrary discretizations and produce output functions that can be evaluated on any discretization or at any resolution. By design, neural operators produce outputs that can be evaluated at arbitrary coordinates and remain consistent across different discretizations, with any differences arising solely from discretization errors that vanish as the resolution is refined. This inherent flexibility enables learning of mappings between functions and capturing multi-scale phenomena in a wide range of applications where data are governed by continuum descriptions. Overall, these properties make neural operators particularly well-suited for scientific computing applications and in particular for learning solution operators of PDEs. 

Among neural operator architectures, the Fourier Neural Operator (FNO)~\citep{li2020fourier} has emerged as the most influential and widely adopted framework for learning operators from data, owing to its simplicity and high efficiency. The FNO parametrizes integral kernels in Fourier space, where convolution operations become simple pointwise multiplications. This spectral formulation allows the model to capture long-range dependencies efficiently while maintaining a global receptive field. By learning to manipulate and act on a compact set of low-frequency modes, the FNO achieves a data-efficient and resolution-invariant representation of the underlying operator. These properties have enabled the successful application of the FNO to a broad range of scientific problems.

Despite these numerous reported successes, several studies have presented inconsistent or inaccurate results regarding Fourier neural operators. These issues often stem from an incomplete understanding of their theoretical foundations and implementation details, as well as from inadequate hyperparameter tuning and limited or incorrect intuition about how individual hyperparameters affect performance. In practice, such difficulties usually arise from incorrect parameter configurations, distortions from data processing or handling, or a general misunderstanding of spectral methods and how to apply them.

We present a comprehensive and practical guide to FNOs. Its purpose is to bridge the gap between theory and practice by explaining the scientific computing and signal processing principles that underpin FNOs, as well as by detailing the design choices that govern their parametrization and performance. We provide clear explanations of their spectral formulation, kernel parameterization, and implementation details to promote both intuition and technical understanding. This guide is closely integrated with the \texttt{NeuralOperator 2.0.0} library, which offers modular and efficient implementations of neural operator architectures, allowing practitioners to connect theoretical concepts directly to their practical realization in \texttt{PyTorch} code. 

\vspace{3mm}

\paragraph{Addressing Common Difficulties and Misunderstandings.} In this guide, we also address several practical difficulties and pervasive misconceptions in the existing literature that have limited the effective use of FNOs. We also provide practical recommendations, to separate genuine limitations of FNOs from artifacts of data generation, preprocessing, and inadequate tuning, and to support their reliable deployment in scientific and engineering applications. More precisely,

\begin{itemize}
  \item  We discuss the number of Fourier modes, a key hyperparameter that is frequently misconfigured. Too few modes prevent the model from capturing essential fine-scale dynamics and lead to overly smooth predictions. Conversely, an excessively large number of modes introduces high-frequency energy through the nonlinearities that is not resolved by the discretization, and this unresolved energy can alias back into lower frequencies, severely degrading generalization across resolutions and super-resolution. In practice, this can manifest itself as aliased predictions with non-physical artifacts. We provide concrete diagnostics for inspecting power spectra, enforcing Nyquist constraints, and anticipating super-resolution use cases, and translate these into actionable guidelines for selecting architectures and preprocessing pipelines. \Cref{sec: Nyquist,sec: n_modes} examine the interplay between Fourier modes, aliasing, and super-resolution.
    \item Contrary to common belief, FNOs are not intrinsically restricted to periodic systems. FNOs can be used on non-periodic systems by constructing a periodic extension on an extended domain and applying the Fourier spectral layers to the periodic extended signal. For many data-driven applications, simple domain padding can be sufficient and highly effective. However, in physics-informed settings that require highly-accurate spectral derivatives at boundaries, padding introduces discontinuities which degrade spectral convergence and trigger Gibbs oscillatory artifacts. In such cases, more advanced extension approaches such as Fourier continuation techniques can be used to construct periodic extensions that preserve smoothness across the extended domain and significantly improve derivatives accuracy. For noisy signals, Fourier continuation can amplify noise and degrade derivatives quality. In this regime, spectrum optimization based extensions determine the added values by minimizing a discrete Sobolev norm, yielding smooth periodic continuations that are more robust to noise. \Cref{sec: Non-periodicity,sec: FNOs Non-periodic domains} discuss how, with appropriate extension strategies, spectral methods and FNOs can be highly effective on non-periodic domains.
         \item We emphasize that standard best practices in machine learning are indispensable for drawing meaningful conclusions. As with other deep learning approaches, the reliable use of FNOs requires systematic problem-dependent tuning of model and training hyperparameters. Many negative or inconclusive results reported in the literature can be traced to non-tuned or under-tuned architectures and optimization settings rather than to fundamental limitations of FNOs. In addition, the behavior of any data-driven approach, including FNOs, depends critically on how data are generated and processed. In particular, assumptions and numerical errors in the solvers used to construct datasets are inherited by the learned neural operator. Likewise, preprocessing steps such as normalization, filtering, and especially downsampling can substantially modify spatial correlations and power spectra. By redistributing energy across scales in a nonphysical way, these can lead to aliasing or create artifacts that are sometimes misinterpreted as intrinsic failures of FNOs. \Cref{sec: Dataset Generation and Data Manipulation} discusses data generation and preprocessing, and \Cref{sec: Hyperparameter Tuning} discusses the systematic hyperparameter tuning required for reliable FNO performance.

    \item We clarify that FNOs are not confined to single input fields, homogeneous domains, or maps between functions of identical dimensionality and resolution. With appropriate embeddings and resolution-invariant components, FNOs can fuse multiple input fields, incorporate scalar and low-dimensional parameters, integrate heterogeneous data defined on different geometries, and produce scalar, vector, functional, or categorical outputs. Furthermore, despite a common misconception, FNOs can naturally handle complex-valued data. \Cref{sec: Mapping of Interest,sec: Embeddings} discuss how FNOs can realize a large variety of mappings.

\end{itemize}

\clearpage 

\hfill 

\vspace{2mm}

\textbf{Outline of the Paper.}

\vspace{3mm}

 In \Cref{sec: Scientific Background}, we cover the analytical and numerical background that motivates operator learning in spectral form, with emphasis on function spaces, Fourier analysis, and practical aspects of dataset construction for scientific problems. More precisely, \vspace{3mm} 
  \begin{itemize}
    \item We introduce the basic notions of function spaces and  operator learning. We discuss how functions can be approximated as finite expansions  in suitable function bases and how this leads to finite-dimensional representations that support spectral formulations of neural operators. \vspace{3mm}
    \item We develop the elements of Fourier analysis needed to understand the mechanisms underlying FNOs. In particular, we first introduce the continuous, discrete, and fast Fourier transforms, and explain how frequency spectra can be used as a tool for analyzing energy across scales. We also discuss the Nyquist-Shannon sampling theorem and aliasing, making explicit how resolution, resolvable frequencies, and the choice of spectral truncation must be coordinated in FNO architectures.  \vspace{3mm}
    \item We discuss spectral methods such as spectral interpolation, spectral resampling, and spectral differentiation, and examine the periodicity assumption underlying them. We show that, while they can produce Gibbs-type artifacts when applied naively to non-periodic data, this is not a fundamental limitation. In particular, we present practical remedies based on padding, Fourier continuation, and spectrum optimization that construct smooth periodic extensions on enlarged domains, which allow Fourier-based architectures to remain fully applicable beyond strictly periodic settings. \vspace{3mm}
    \item We explain how to specify the mapping to approximate and clarify common misconceptions about the types of inputs and outputs that neural operators can handle. 
    \begin{itemize}
        \item On the input side, neural operators are not limited to a single function or homogeneous domains: they can jointly process multiple functions, incorporate scalar or low-dimensional quantities via embeddings, and fuse heterogeneous data defined on different geometries or modalities into a shared latent representation. 
        \item On the output side, they are not restricted to functions of the same dimension and resolution as the inputs: through appropriate resolution-invariant transformations, they can produce scalar, vector, or categorical quantities, map between function spaces of different dimensions, and evaluate outputs on arbitrary discretizations.  \end{itemize} \vspace{3mm}
    \item We describe how datasets can be assembled from numerical solvers and experiments, including multi-solver and multi-resolution settings, and how assumptions made to acquire the data typically translate to the learned model. We also discuss how different downsampling strategies can affect spatial structure and power spectra of the data in different ways, stressing how these choices influence the behavior of the learned operator. We also recommend visualization tools for fields, spectra, errors, and PDE residuals that enable qualitative assessment of model performance.

  \end{itemize}
  
 \hfill 

\clearpage 
 
In \Cref{sec: NOs}, we introduce neural operators and their advantages, describe the FNO structure and implementation in detail, and explain how it can be adapted to a broad class of tasks in scientific computing. We also list scientific applications in which FNOs have been deployed successfully. More precisely,\vspace{2mm} 
  \begin{itemize}
   \item We introduce neural operators as families of mappings approximating continuous operators, and contrast them with standard neural networks that operate on fixed-dimensional vectors. This paradigm leads to advantages for scientific modeling, such as the ability to generalize across discretizations, to learn entire families of PDE solutions, and to act as efficient surrogates that better respect the underlying PDE structure and often generalize more reliably. \vspace{2mm}
    \item We present the FNO and describe every single one of its components in great detail. In particular, we explain how each layer combines a learnable spectral convolution, local channel mixing, nonlinearities, and residual connections, and how choices such as the number of retained modes, depth, and feature width control expressivity and computational cost. We also examine the use of FNOs for super-resolution tasks, and provide practical guidelines for training and evaluation at different resolutions.\vspace{2mm}
    \item We present mechanisms for embedding input and output data for FNOs. We show how resolution can be controlled inside the architecture, to predict on grids finer than the inputs. We describe embedding strategies for scalar or constant parameters, such as sinusoidal embeddings, which lift them into rich multi-frequency signals, and discuss how to concatenate multiple input and output fields, potentially defined on different grids, into a shared resolution-invariant latent representation. We further outline strategies for mapping between inputs and outputs of different dimensionalities, for training on multi-resolution datasets without architectural changes, and for handling complex-valued data. \vspace{2mm}
    \item We address the common misconception that FNOs are restricted to periodic problems and explain how they can be applied to non-periodic settings using padding and Fourier continuation. \vspace{2mm}
    \item We list examples of scientific applications of neural operators, and discuss how they can be used for inverse design through accelerated large-scale design exploration and gradient-based optimization.\vspace{2mm}
    \item We also present the \texttt{NeuralOperator} PyTorch library which provides a state-of-the-art and modular implementation of numerous neural operator architectures.
  \end{itemize}

 \hfill  

\vspace{4mm}
 
In \Cref{sec: Hyperparameter Tuning}, we provide practical guidance for tuning FNO hyperparameters to obtain accurate, stable, and efficient models, in conjunction with general deep learning practices. More precisely,\vspace{2mm}
  \begin{itemize} \item We discuss extensively the choice of the number of Fourier modes, a very important hyperparameter which is often incorrectly tuned. Too few modes may hinder the FNO from learning essential fine-scale dynamics well, while too many modes can inject unresolved high-frequency energy that aliases back into lower frequency modes through nonlinearities, and severely compromise generalization to different resolutions. We provide concrete guidance for inspecting power spectra, enforcing Nyquist constraints, accounting for anisotropy, anticipating super-resolution use cases, and ultimately selecting a number of modes that balances small-scale fidelity, robustness, and computational cost. \vspace{2mm}
     \item We discuss how the number of layers and hidden channels in FNOs influence capacity and computational costs. We also examine additional hyperparameters of the \texttt{NeuralOperator} FNO implementation, including support for complex-valued data, alternative normalization strategies, choices of nonlinear activation, separable spectral convolutions, and explicit control over numerical precision. \vspace{2mm}
    \item We advocate small-scale overfitting experiments as a diagnostic tool, since fitting a tiny subset of the data can rapidly reveal inadequate capacity or training losses. We discuss how insights from these experiments can guide hyperparameter adjustments before committing to full training runs.
  \end{itemize}

\clearpage 

\hfill 
 
In \Cref{sec: Advanced Training Strategies}, we explore more advanced training strategies. More precisely,\vspace{2mm} 
  \begin{itemize}
    \item We present common loss functions that are appropriate for operator learning tasks, and discuss strategies to tailor them so they are better aligned with given scientific goals. We also discuss uncertainty quantification in predictions from neural operators. \vspace{2mm}
    \item We discuss autoregressive rollout strategies for time-dependent systems. We highlight their advantages (reduced learning complexity, data-efficiency) and discuss how to mitigate challenges that arise over long horizons (distribution shifts, error accumulation). We introduce the Recurrent Neural Operator (RNO), which augments FNOs with a recurrent structure to produce more stable long-range forecasts.\vspace{2mm}
    \item We investigate multi-resolution training schemes and present the incremental FNO (iFNO) framework where the FNO's expressivity and the data resolution are progressively increased, allowing for more data-efficient and stable training.\vspace{2mm}
    \item We motivate physics-informed approaches for regimes where high-quality simulation or experimental data are scarce, noisy, only available at coarse resolutions, or completely unavailable. In particular, we introduce the Physics-Informed Neural Operator (PINO), where a neural operator is trained with a combination of data losses and physics-based losses, and we compare approaches for computing the derivatives appearing in physics losses accurately and efficiently. We also present adaptive weighting strategies that automatically rebalance the different loss terms during training.

  \end{itemize}

 \hfill  

\vspace{8mm}

In \Cref{sec: Advanced Architectural Modifications}, we present architectural extensions of FNOs for more compressed, localized, or geometric settings that occur in demanding scientific applications. More precisely,\vspace{2mm} 
  \begin{itemize}
    \item We present the Tensor FNO (TFNO) which uses low-rank tensor decompositions to compress weight representations, reducing parameter counts and memory footprint, accelerating training, and improving generalization by 	reducing the risk of overfitting, while retaining expressive power.\vspace{2mm}
    \item We augment FNOs with localized differential and integral kernels that capture short-range structure while remaining well-defined in the continuum limit. We show how finite-difference-style differential layers and discrete-continuous integral kernels arise as limits of convolutions, describe their implementation in dedicated blocks, and discuss how these hybrid architectures balance local and global interactions while offering structured control over receptive field and computational cost. \vspace{2mm}
    \item We discuss how to generalize FNOs to structured non-Euclidean domains, building on harmonic analysis and group actions to define Fourier transforms that respect the symmetries of manifolds. We discuss the special case of the sphere, where spherical harmonics provide the basis for the Spherical FNO (SFNO), which has catalyzed recent state-of-the-art weather forecasting architectures.\vspace{2mm}
    \item We extend FNOs to geometrically complex settings by coupling them with geometric encoder-decoder architectures that operate on irregular point clouds. A geometric encoder maps fields on irregular geometries to a latent regular grid, an FNO operates there with efficient FFT-based spectral convolutions, and a geometric decoder maps predictions back to the physical domain. Within this framework, we present Geo-FNO, GINO, and OTNO.
    
  \end{itemize}

\clearpage 

\section{Scientific Computing Background}
\label{sec: Scientific Background}

Understanding how Fourier Neural Operators (FNOs) work requires familiarity with fundamental concepts from scientific computing, including numerical linear algebra, signal processing, and spectral analysis. Neural operators are built upon mathematical ideas from functional analysis and Fourier theory, which provide the foundation for representing and learning mappings between functions. This section offers a concise and intuitive overview of these core principles, focusing on the aspects most relevant to operator learning, while keeping the exposition at a high level to avoid unnecessary technical details. Together, these topics establish the mathematical and computational background necessary to understand how FNOs leverage spectral representations to approximate mappings between infinite-dimensional function spaces. 

We begin by introducing function spaces and operators, together with their representations through basis functions and expansions. We then review essential concepts from Fourier and spectral analysis, including the continuous and discrete Fourier transforms, the Nyquist--Shannon sampling theorem, spectral interpolation, and spectral differentiation, as well as the assumptions and limitations associated with periodicity in spectral methods. We refer the reader to the classical texts~\citep{trefethen1997numerical,boyd2001chebyshev,gottlieb1983numerical} for comprehensive treatments of numerical linear algebra and spectral methods, including theoretical foundations, implementation details, and practical insights relevant to the topics discussed in this section. Finally, we discuss practical aspects of dataset generation and manipulation for operator learning, including data sources, data cleaning, downsampling, and visualization, and refer the reader to~\citep{press2007numerical} for a more comprehensive overview of general scientific computing principles and data processing methodologies. \\

\subsection{Function Spaces and Operators}  \label{sec: Function Spaces and Operators}

\subsubsection{Definitions} \label{sec: Function Space Definitions}

\vspace{1mm}

A \mydef{function} is a mapping $f$ that assigns to each element \( x \) in a set \( X \) exactly one element \( f(x) \) in a set \( Y \). Classical neural networks aim to approximate functions between finite-dimensional Euclidean spaces, such as $\R^d$. More precisely, they take a vector $x$ as input and output another vector $f(x)$. Given a function $f$, we denote by $D_f$ its domain and by $f(x)$ its evaluation at coordinates $x \in D_f$.

Many problems in scientific computing involve representing complicated or partially observed functions in simpler forms. Approximating functions with well-chosen basic components allows one to analyze, compute, and interpret them efficiently. This idea parallels the representation of vectors in finite-dimensional spaces, where each vector can be written as a linear combination of basis vectors. Extending this concept to sets of functions leads to the notion of function spaces, function bases, and function expansions. 

A \mydef{function space} is a collection of functions that share certain properties and are equipped with a structure that allows mathematical analysis. This structure often comes from defining operations such as addition and scalar multiplication, together with a notion of distance, norm, or inner product. Examples include the space of continuous functions, the space of differentiable functions, and the space of polynomials of degree at most $N$. In such spaces, concepts such as convergence and orthogonality can be extended to functions, paving the way for analysis in infinite dimensions.

\mydef{Neural operators} are principled extensions of neural networks which aim to approximate \mydef{operators}, i.e. mappings on function spaces. An operator $T$ maps any input function $f$ from one function space \( \mathcal{F}_{in} \) to a function $T(f)$ in the same or a different function space \( \mathcal{F}_{out} \). Operators describe transformations of functions, including processes such as differentiation and integration. Dynamical systems can often be described using partial differential operators, and we can define the PDE solution operator as the mapping that takes an input function (e.g. the initial state or boundary conditions) to the corresponding solution of the PDE.

\paragraph{$L^p$ spaces of integrable functions.} A very important function space is the space of square-integrable functions, denoted by $L^2(D)$ for a domain $D \subseteq \mathbb{R}^d$. The $L^2$ \emph{norm} of a function $f$ is defined as
\begin{equation}
    \|f\|_2 = \left( \int_D |f(x)|^2 \, dx \right)^{1/2},
\end{equation}
and a function $f$ belongs to $L^2(D)$ if and only if $\|f\|_2 < \infty$. 

The space $L^2(D)$ is equipped with an \emph{inner product} given by
\begin{equation}
\langle f, g \rangle = \int_{D} f(x) g(x) \, dx,
\end{equation}
inducing the norm $\|f\|_2 = \sqrt{\langle f, f \rangle}$. This inner product extends geometric ideas such as length, angle, and orthogonality to spaces of functions (e.g. two functions $f$ and $g$ are said to be \emph{orthogonal} if $\langle f, g \rangle = 0$). The notion of the $L^2$ norm can be generalized naturally to the $L^p$ \emph{norm}, $\|f\|_p = \left( \int_D |f(x)|^p \, dx \right)^{1/p}$, for any $1 \leq p < \infty$ and the corresponding function space $L^p(D) = \left\{ f : D \to \R \ | \ \|f\|_p < \infty \right\}$. These function spaces form the backbone of functional analysis and the theory of PDEs.

\hfill 

\subsubsection{Basis Functions and Expansions}   \label{sec: Basis Functions and Expansions}

\vspace{2mm}

Just as vectors can be represented in terms of basis vectors, functions can be expressed as combinations of basis functions. A collection $\{\varphi_k\}_{k=0}^{\infty}$ of functions forms a \mydef{basis} of a function space (and its elements are referred to as \mydef{basis functions}) if any function $f$ in that space can be represented as
\begin{equation}
f(x) = \sum_{k=0}^{\infty} c_k \, \varphi_k(x),
\end{equation}
where the coefficients $c_k$ depend on $f$. In practice, one often considers finite approximations of the form
\begin{equation}
f(x) \approx \sum_{k=0}^{N} c_k \, \varphi_k(x),
\end{equation}
where the approximation becomes more accurate as $N$ increases. This idea underlies many analytical and numerical techniques, and the expansion typically converges as $N \to \infty$ under appropriate conditions on the basis functions and on $f$.

\vspace{2mm}

\paragraph{Example 1: Polynomials.} As a simple example, the polynomials $\{1, x, x^2, \dots, x^N\}$ form a basis for the space of polynomials of degree up to $N$. Many smooth functions can be locally approximated near a point $x_0$ by polynomials through the \emph{Taylor expansion}
\begin{equation}
f(x) \approx \sum_{k=0}^{N} \frac{f^{(k)}(x_0)}{k!} (x - x_0)^k.
\end{equation}
Although the polynomial basis is not orthogonal with respect to the standard $L^2$ inner product, orthogonalized versions such as the \emph{Legendre} or \emph{Chebyshev polynomials} are widely used in practice. 

\vspace{2mm}

\begin{figure}[h]  
\centering   
\includegraphics[width=\textwidth]{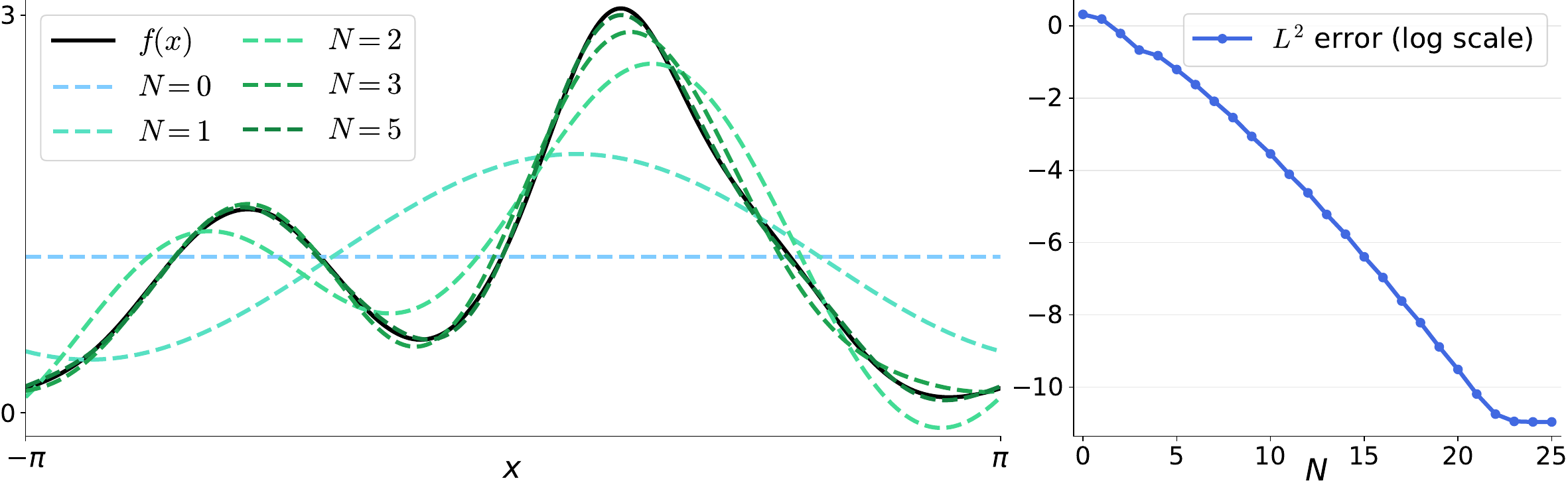}  \vspace{-7mm}
\caption{Example of Fourier series for the periodic function $f(x) = \cos(x) + \sin(x)^4 + 0.5\exp(\sin(2x))$ truncated at different values of $N$, next to the $L^2$ error of the truncated Fourier series.   
\label{fig: Fourier Series Example}} \vspace{7mm}
\end{figure}

\paragraph{Example 2: Trigonometric Functions.} Another important class of basis functions is given by the trigonometric system $\{\sin(nx), \cos(nx)\}_{n=0}^{\infty}$, which forms an orthogonal family of functions on the interval $[-\pi, \pi]$ with respect to the $L^2$ inner product. It can be shown that any periodic function $f(x)$ can be represented as a sum of sines and cosines, leading to the \mydef{Fourier series} expansion
\begin{equation}
f(x) = a_0 + \sum_{k=1}^{\infty} \big( a_k \cos(kx) + b_k \sin(kx) \big),
\end{equation}
where $a_k$ and $b_k$ are the Fourier coefficients of $f(x)$. The idea that any sufficiently regular periodic function can be expressed as an infinite superposition of sine and cosine functions forms the foundation of Fourier analysis, with deep implications in differential equations, signal processing, and data representation. \Cref{fig: Fourier Series Example} illustrates the convergence of the Fourier series for a periodic function as the truncation index $N$ increases.

\hfill

\subsection{Applied Spectral/Fourier Analysis} \label{sec: Applied Spectral/Fourier Analysis}

\vspace{2mm}

\subsubsection{The Continuous Fourier Transform} \label{sec: Continuous FT}

\vspace{2mm}

Fourier analysis extends the idea of representing periodic functions as sums of sines and cosines to more general functions. Its central principle is that a function can be decomposed into its constituent frequency components, revealing how oscillations at different scales contribute to its overall structure. This frequency-based representation is fundamental in signal processing and applied mathematics.

While the Fourier series applies to periodic functions on bounded intervals, the \mydef{Fourier transform} generalizes the concept to functions defined on the entire real line or, more generally, on $\R^d$. For a sufficiently well-behaved function $f$ on $\R^d$, the \mydef{continuous Fourier transform} is defined as
\begin{equation}
\hat{f}(\xi) = \frac{1}{(2\pi)^{d/2}} \int_{\R^d} f(x) \, e^{-i \, x \cdot \xi} \, dx,
\end{equation}
where $\xi \in \R^d$ denotes the frequency variable. The corresponding \mydef{inverse Fourier transform} $\mathcal{F}^{-1}$ recovers $f$ from its frequency representation,
\begin{equation}
f(x) = \mathcal{F}^{-1}(\hat{f})(x) = \frac{1}{(2\pi)^{d/2}} \int_{\R^d} \hat{f}(\xi) \, e^{i \, x \cdot \xi} \, d\xi.
\end{equation}

\paragraph{Convolutions.} An important operation in Fourier analysis is the \mydef{convolution}. For functions $f$ and $g$ on $ \R^d$, their convolution is defined by
\begin{equation}
(f * g)(x) = \int_{\R^d} f(y) \, g(x - y) \, dy.
\end{equation}

Convolution expresses how one function interacts with another by sliding across it, multiplying where they overlap, and integrating the result. In effect, it describes how one function smooths, filters, or modulates another, a principle that underlies phenomena ranging from signal processing and heat diffusion to the core operation of convolutional neural networks. A key property of the Fourier transform is that it converts convolutions in the spatial domain into pointwise products in the frequency domain. This result is known as the \mydef{Spectral Convolution Theorem}, and can be written as 
\begin{equation}\mathcal{F}(f * g) = \mathcal{F}(f) \cdot \mathcal{F}(g)  \qquad \text{or} \qquad  f * g = \mathcal{F}^{-1}\left(\mathcal{F}(f) \cdot \mathcal{F}(g)\right) .\end{equation}

\hfill 

\subsubsection{The Discrete Fourier Transform (DFT)}  \label{sec: DFT}

\vspace{2mm}

In practical applications, one typically works with discrete data rather than continuous functions. The \mydef{Discrete Fourier Transform (DFT)} provides a finite-dimensional analogue of the continuous Fourier transform. In the one-dimensional case, given a sequence of complex numbers $\{x_n\}_{n=0}^{N-1}$, the DFT and its inverse, the \mydef{Inverse Discrete Fourier Transform (IDFT)} are defined respectively as
\begin{equation} \label{eq: DFT IDFT}
X_k = \frac{1}{\sqrt{N}} \sum_{n=0}^{N-1} x_n \, e^{-2\pi i kn / N}, 
\qquad 
x_n = \frac{1}{\sqrt{N}} \sum_{k=0}^{N-1} X_k \, e^{2\pi i kn / N},
\end{equation}
for $k, n = 0, 1, \dots, N-1$. The DFT maps a discrete signal from the temporal or spatial domain to the frequency domain, enabling efficient analysis of the frequency content of the signal. In higher dimensions, the transform is applied along each dimension separately. 

Note that for the exposition here, we adopt the \emph{orthonormal} normalization convention using factors of $1/\sqrt{N}$ for both the DFT and IDFT, which renders them orthonormal (and analogously in the continuous setting). Alternative conventions commonly used in the literature and implementations include the \emph{forward} normalization (with a factor of $1/N$ in the DFT and none in the IDFT) and the \emph{backward} normalization (with a factor of $1/N$ in the IDFT and none in the DFT).

\paragraph{Power Spectrum.} Given the discrete Fourier transform coefficients $\{X_k\}_{k=0}^{N-1}$ of a discrete signal, we can compute and plot its \mydef{power spectrum}, which provides a quantitative measure of how the signal's energy is distributed across frequencies. The power spectrum is obtained via $P_k = |X_k|^2$ for $k = 0, \dots, N-1$. Each $P_k$ represents the contribution of the frequency component corresponding to index $k$ to the total signal energy. Since the DFT decomposes a signal into orthogonal frequency components, the total energy of the signal satisfies Parseval's identity, $\sum_{n=0}^{N-1} |x_n|^2 = \sum_{k=0}^{N-1} |X_k|^2 = \sum_{k=0}^{N-1} P_k$. This equality shows that the power spectrum partitions the total signal energy among the different frequencies.

\begin{figure}[h]  
\centering    \vspace{2mm}
\includegraphics[width=\textwidth]{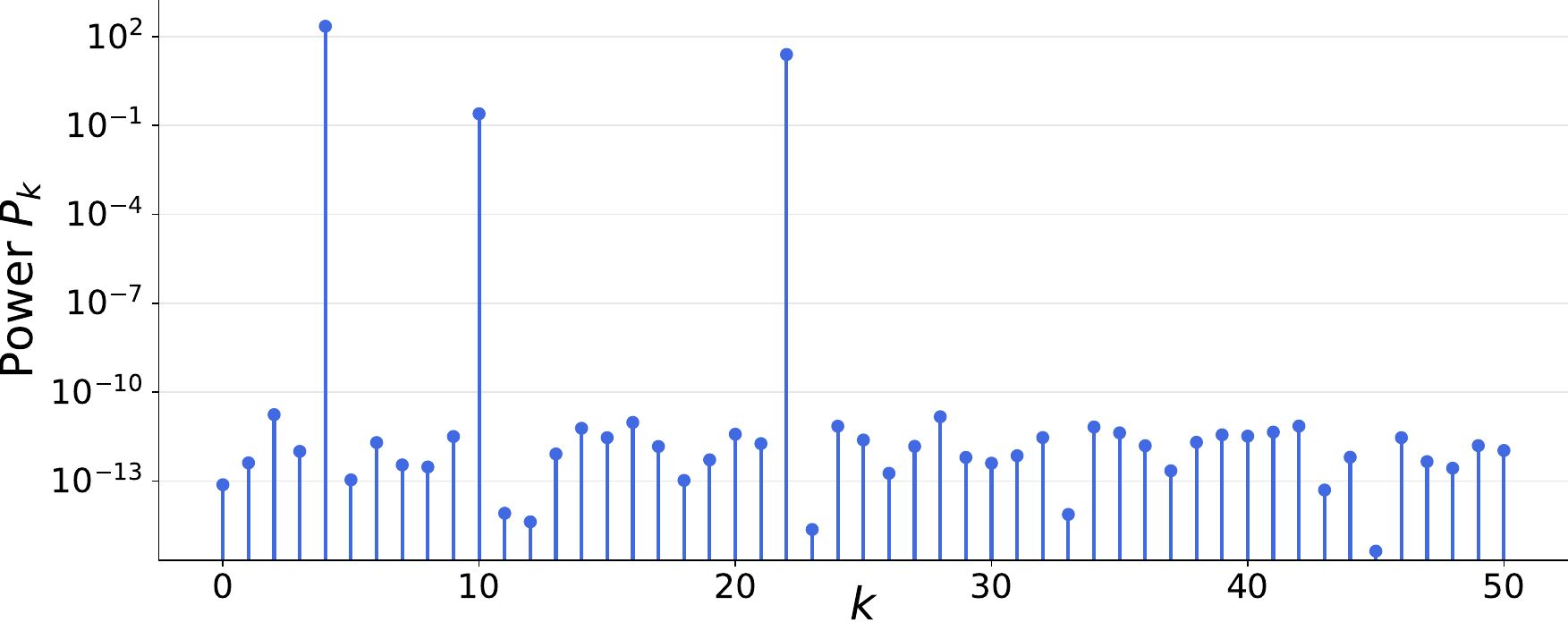}  \vspace{-7mm}
\caption{Power spectrum of the function $f(x) = 3\cos(4x) + 0.1\sin(10x) + \cos(22x)$. Distinct peaks appear at $k = 4$, $10$, and $22$, revealing the dominant periodic components at these frequencies. The amplitude of the peak at $k = 4$ is noticeably higher, reflecting the larger coefficient 3 compared to the weaker contribution from the $k = 10$ term with coefficient 0.1.}
\label{fig: Power Spectrum Example}  
\end{figure}

The power spectrum thus reveals which frequencies dominate a signal. Sharp peaks indicate strong periodic components at specific frequencies, while a broad spectrum corresponds to a signal that is more irregular or noise-like.  In practice, the power spectrum is often plotted as $P_k$ versus the corresponding frequency, providing an intuitive fingerprint of a signal's oscillatory content.  It is widely used in scientific computing to identify characteristic frequencies, detect periodicity, or analyze energy transfer across scales. \Cref{fig: Power Spectrum Example} displays the power spectrum for an example of a periodic function.

\clearpage 

\subsubsection{The Fast Fourier Transform (FFT)}
\label{sec: fft}

\vspace{2mm}

Although the DFT is conceptually straightforward, its direct computation requires $\mathcal{O}(N^2)$ arithmetic operations for a signal of length $N$, since each of the $N$ output coefficients involves a sum over $N$ input terms. The \mydef{Fast Fourier Transform (FFT)} is an algorithmic breakthrough that computes the same result in only $\mathcal{O}(N \log N)$ time by recursively exploiting the symmetries in the complex exponential factors of the DFT. The FFT made large-scale spectral analysis and numerical simulation computationally feasible and remains one of the most influential algorithms in scientific computing.

To illustrate the key idea of an FFT approach, consider a one-dimensional sequence $\{x_n\}_{n=0}^{N-1}$ with equidistant spacing, where $N$ is a power of two. The sequence $\{x_n\}_{n=0}^{N-1}$ can be split into its even components $x_{2\ell}$ and odd components $x_{2\ell+1}$. Substituting this decomposition into the formula \eqref{eq: DFT IDFT} for its DFT gives
\begin{equation}
X_k 
= \frac{1}{\sqrt{N}} \sum_{\ell=0}^{N/2-1} x_{2\ell} \, e^{-2\pi i k (2\ell) / N}
  + \frac{1}{\sqrt{N}} e^{-2\pi i k / N} \sum_{\ell=0}^{N/2-1} x_{2\ell+1} \, e^{-2\pi i k (2\ell) / N}.
\end{equation}
Define the length $N/2$ unitary DFTs of the even and odd subsequences by
\begin{equation}
X_k^{(1)} = \frac{1}{\sqrt{N/2}} \sum_{\ell=0}^{N/2-1} x_{2\ell} \, e^{-2\pi i k \ell / (N/2)},
\quad \ \ 
X_k^{(2)} = \frac{1}{\sqrt{N/2}} \sum_{\ell=0}^{N/2-1} x_{2\ell+1} \, e^{-2\pi i k \ell / (N/2)}.
\end{equation}
Then
\begin{equation}
X_k = \frac{1}{\sqrt{2}} \Big( X_k^{(1)} + e^{-2\pi i k / N} X_k^{(2)} \Big).
\end{equation}
Therefore, one DFT of size $N$ is reduced to two unitary DFTs of size $N/2$, plus $\mathcal{O}(N)$ multiplications by the phase factors $e^{-2\pi i k / N}$. This process is applied recursively to the smaller signals of length $N/2$. As a result, the number $\text{Ops}(N)$ of operations required to compute an FFT of length $N$ satisfies the recursive relation

\begin{equation}
\text{Ops}(N) = 2 \ \text{Ops}(N/2) + \mathcal{O}(N),
\end{equation}
which yields the desired computational complexity $\text{Ops}(N) = \mathcal{O}(N \log N)$. 

Multiple variants of the FFT exist to further optimize computations and memory usage. Multidimensional FFTs can be obtained by applying one-dimensional FFTs along each dimension independently, and reduce the $\mathcal{O}\left((N_1\ldots N_d)^2\right)$ computational cost of the multi-dimensional DFT to $\mathcal{O}\left((N_1\ldots N_d) \sum_{m=1}^d \log{N_m}\right)$.

The IDFT can be accelerated in the same way, leading to the \mydef{Inverse Fast Fourier Transform (IFFT)}. 

\vspace{3mm}

\paragraph{Implementation Details.} 

When applying FFT and IFFT operations, the number of discrete Fourier modes matches the number of points at which the physical signal is sampled uniformly, along each transformed dimension. Concretely, along any transformed axis of length $N$, the FFT maps an $N$-point signal to $N$ complex Fourier coefficients, and the IFFT maps these $N$ coefficients back to an $N$-point signal (and analogously, dimension-by-dimension, in the multidimensional setting). By adding higher-frequency modes with zero coefficients, the length-$N$ spectrum can be embedded into a longer length-$M$ spectrum, and the IFFT returns an $M$-point signal on a denser grid over the same physical domain: the original $N$ Fourier modes specify a function, and zero-padding simply asks the IFFT to evaluate that function at more sample locations. See \Cref{sec: Spectral Interpolation} for more details on spectral interpolation and resampling. 

One must also be aware that libraries differ in how they order the discrete frequency modes in the returned tensor. In \texttt{PyTorch}, the FFT lists the nonnegative modes first and then the negative ones, namely
\begin{equation}
\begin{cases} \quad
\big[\,X_{0},\,X_{1},\,\dots,\,X_{N/2-1},\,X_{-N/2},\,X_{-N/2+1},\,\dots,\,X_{-1}\,\big], \ \ &   N \ \text{even},\\[1mm] \quad
\big[\,X_{0},\,X_{1},\,\dots,\,X_{(N-1)/2},\,X_{-(N-1)/2},\,\dots,\,X_{-1}\,\big], & N \ \text{odd}.
\end{cases}
\end{equation}
If a centered ordering is preferred (negative modes first, then $0$, then positive modes), one can apply \texttt{torch.fft.fftshift} to the output of the \texttt{PyTorch} FFT. To recover the non-centered ordering afterwards, one can apply the inverse permutation, \texttt{torch.fft.ifftshift}.

\clearpage

\subsubsection{The Nyquist--Shannon Sampling Theorem} \label{sec: Nyquist}

\vspace{2mm}

The \mydef{Nyquist--Shannon Sampling Theorem} establishes the conditions under which a continuous-time signal can be perfectly reconstructed from its discrete samples, ensuring that no information is lost.  

\begin{figure}[h]  
\vspace{2mm}
\centering   
\includegraphics[width=\textwidth]{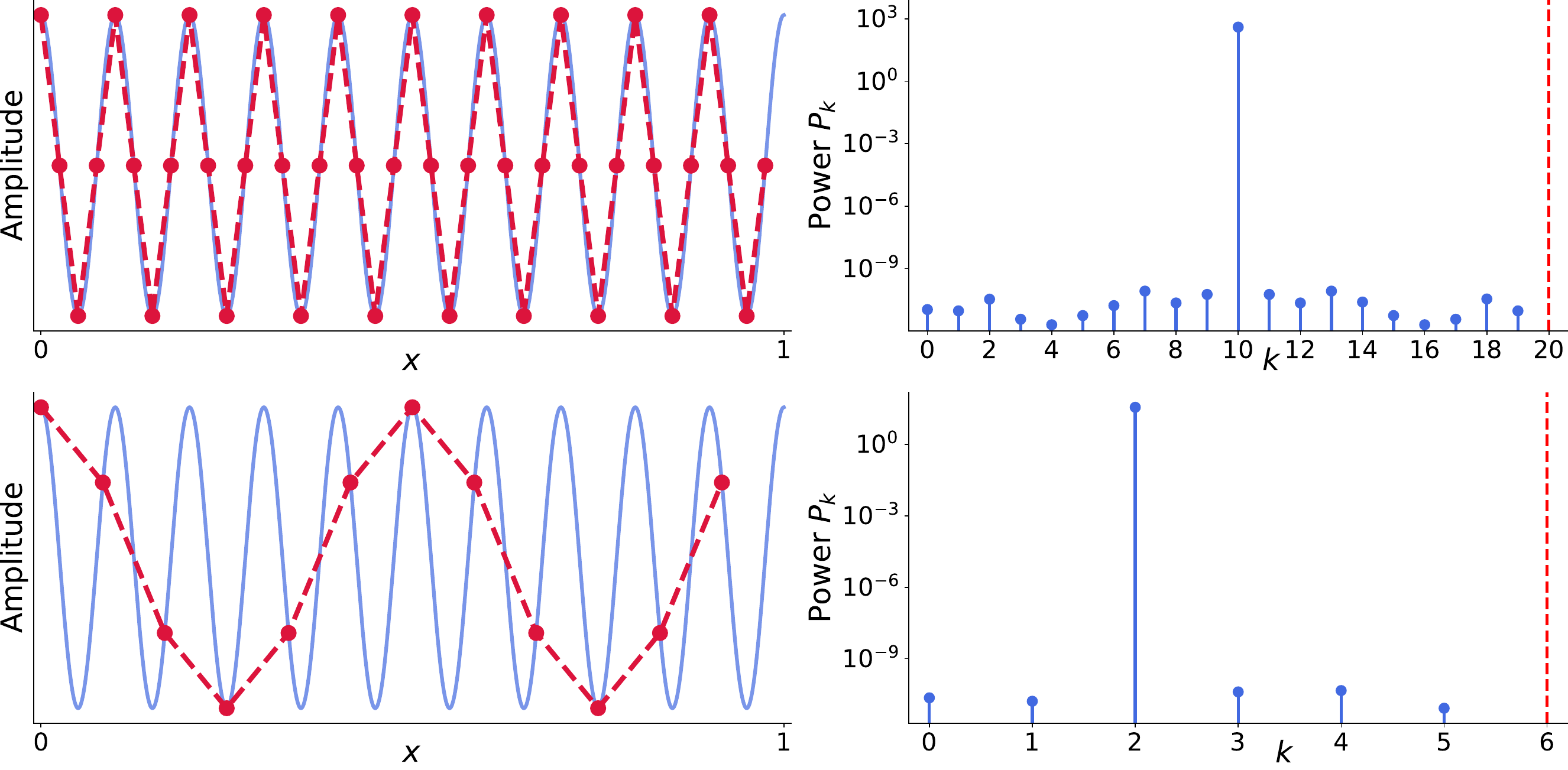}  \vspace{-6mm}
\caption{Illustration of aliasing in the time and frequency domains. The continuous signal $f(x) = \cos(20\pi x)$ (in blue) is shown together with its sampled versions at two different sampling rates (in red). When the sampling rate exceeds the Nyquist limit (top), the samples accurately capture the waveform, and the power spectrum shows a single peak at the true frequency. When the sampling rate is below the Nyquist limit (bottom), the samples misrepresent the signal, producing an apparent lower-frequency oscillation in the time domain and a corresponding aliased peak in the frequency spectrum.}
\label{fig: Aliasing}\vspace{4mm}
\end{figure}

Let $f(t)$ be a continuous signal whose Fourier transform $\hat{f}(\xi)$ is supported within a finite frequency band $[-\xi_{\max}, \xi_{\max}]$.  
If the signal is sampled at regular intervals with sampling frequency $f_s = 1 / \Delta t$, the Nyquist--Shannon theorem states that perfect reconstruction of $f(t)$ from its samples is possible provided that
\begin{equation}
f_s \geq 2 \, \xi_{\max}.
\end{equation}
The critical frequency $f_{\mathrm{Nyquist}} = f_s / 2$, called the \mydef{Nyquist frequency}, represents the highest frequency that can be accurately represented when sampling at rate $f_s$.  
If the signal contains frequencies above \(f_{\rm Nyquist}\) $f_{\mathrm{Nyquist}}$,the high-frequency components of this signal will be misrepresented as lower frequencies, a phenomenon known as \mydef{aliasing}. When aliasing occurs, high frequencies above the Nyquist limit $f_{\mathrm{Nyquist}}$ are effectively folded-back into the low-frequency range $[-f_{\mathrm{Nyquist}}, f_{\mathrm{Nyquist}}]$, and contaminate the spectrum by producing spurious or distorted low-frequency artifacts that were not present in the original signal. Aliasing is illustrated with a simple example in \Cref{fig: Aliasing}.

\paragraph{Relation to the DFT.} The DFT provides a natural framework for understanding this effect. It can be viewed as evaluating the Fourier transform of a sampled, periodic version of the original continuous signal.  
Because sampling induces periodicity in the frequency domain, the DFT spectrum inherently represents the signal only within the primary frequency band $[-f_{\mathrm{Nyquist}}, f_{\mathrm{Nyquist}}]$.  
All spectral content beyond this interval is indistinguishable from its aliased counterpart within it.  
Hence, the DFT's frequency bins correspond precisely to the discrete set of resolvable frequencies in this range, and there is no meaningful information beyond the Nyquist limit.  
This symmetry is also reflected in the DFT of real-valued signals, where the positive-frequency and negative-frequency components are complex conjugates, ensuring that the frequency content is fully captured up to $f_{\mathrm{Nyquist}}$.

In practical terms, the Nyquist--Shannon theorem dictates the minimum sampling rate required to capture a signal's true spectral content without distortion. This principle ensures that the discrete representation of a function or measurement preserves the integrity of its frequency spectrum. \\

\paragraph{Importance for Fourier Neural Operators.} In the FNO architecture, only a finite number of Fourier modes are retained for the spectral convolution operation. The Nyquist limit implies that we should be careful when setting the number of Fourier modes. Including too many modes relative to the resolution can cause the model to learn from artificial or aliased frequency components that do not correspond to the true underlying function. Consequently, when the trained FNO is evaluated at a higher resolution, these spurious high-frequency modes can yield unstable or physically inconsistent predictions. Respecting the Nyquist limit therefore ensures both numerical stability and physical fidelity in Fourier-based spectral methods, including FNOs. \\

\hfill

\subsubsection{Spectral Interpolation} \label{sec: Spectral Interpolation}

\vspace{3mm}

\mydef{Spectral interpolation}, or \mydef{spectral resampling}, provides a principled way to change the spatial resolution or discretization of a discretely sampled signal via its frequency representation. The key idea is to transform a discrete signal into the frequency domain to obtain its representation as an expansion in trigonometric functions, and then evaluate that expansion using the inverse Fourier transform at a different set of points. This approach leverages the Fourier representation to produce smooth and globally consistent interpolants, in contrast to local interpolation schemes.

Consider discrete samples $\{x_n\}_{n=1}^{N}$ in $[0,1]$ of a smooth one-dimensional periodic function $f(x)$. The DFT provides an approximation of the Fourier coefficients $\{X_k\}$ of $f$, given by
\begin{equation}
X_k = \frac{1}{\sqrt{N}} \sum_{n=1}^{N} f(x_n) \, e^{-2\pi i k x_n}.
\end{equation} 
Once these coefficients are known, the function can be estimated at any other set of points $\{y_j\}_{j=1}^{M}$ through the inverse DFT,
\begin{equation}
f(y_j) \approx  \frac{1}{\sqrt{N}} \sum_{k=-N/2}^{N/2-1} X_k \, e^{2\pi i k y_j}.
\end{equation}
This defines a smooth, periodic trigonometric interpolant that exactly reproduces the original data at the sample points $\{x_n\}$ and provides a natural way to interpolate at arbitrary locations $\{y_j\}$, whether uniformly spaced or not.  In this sense, the Fourier coefficients define a global representation of the function, from which interpolation becomes evaluation of a continuous spectral expansion rather than polynomial fitting. \\

\paragraph{Spectral Interpolation on Regular Grids.} When the data $\{x_n\}_{n=0}^{N-1}$ are sampled on a uniform grid, the computation of the DFT and its inverse can be performed efficiently using the FFT and IFFT. Suppose we wish to resample the signal onto a finer grid of size $M > N$. After computing the FFT of the signal to obtain its frequency coefficients $\{X_k\}_{k=0}^{N-1}$, one can zero-pad the array of Fourier coefficients in the frequency domain so that its total length becomes $M$. This operation needs to be done in accordance with the ordering of the modes returned by the FFT. As discussed in \Cref{sec: fft}, in \texttt{PyTorch}, the FFT lists the nonnegative modes first and then the negative ones, as
\begin{equation}
\begin{cases} \quad
\big[\,X_{0},\,X_{1},\,\dots,\,X_{N/2-1},\,X_{-N/2},\,X_{-N/2+1},\,\dots,\,X_{-1}\,\big], \ \ &   N \ \text{even},\\[1mm] \quad
\big[\,X_{0},\,X_{1},\,\dots,\,X_{(N-1)/2},\,X_{-(N-1)/2},\,\dots,\,X_{-1}\,\big], & N \ \text{odd}.
\end{cases}
\end{equation}

Thus, in \texttt{PyTorch}, one would insert $(M - N)$ zeros into the center of the spectrum, between the positive and negative frequency components. 

\vspace{2mm}

\begin{figure}[t]  
\centering   
\includegraphics[width=\textwidth]{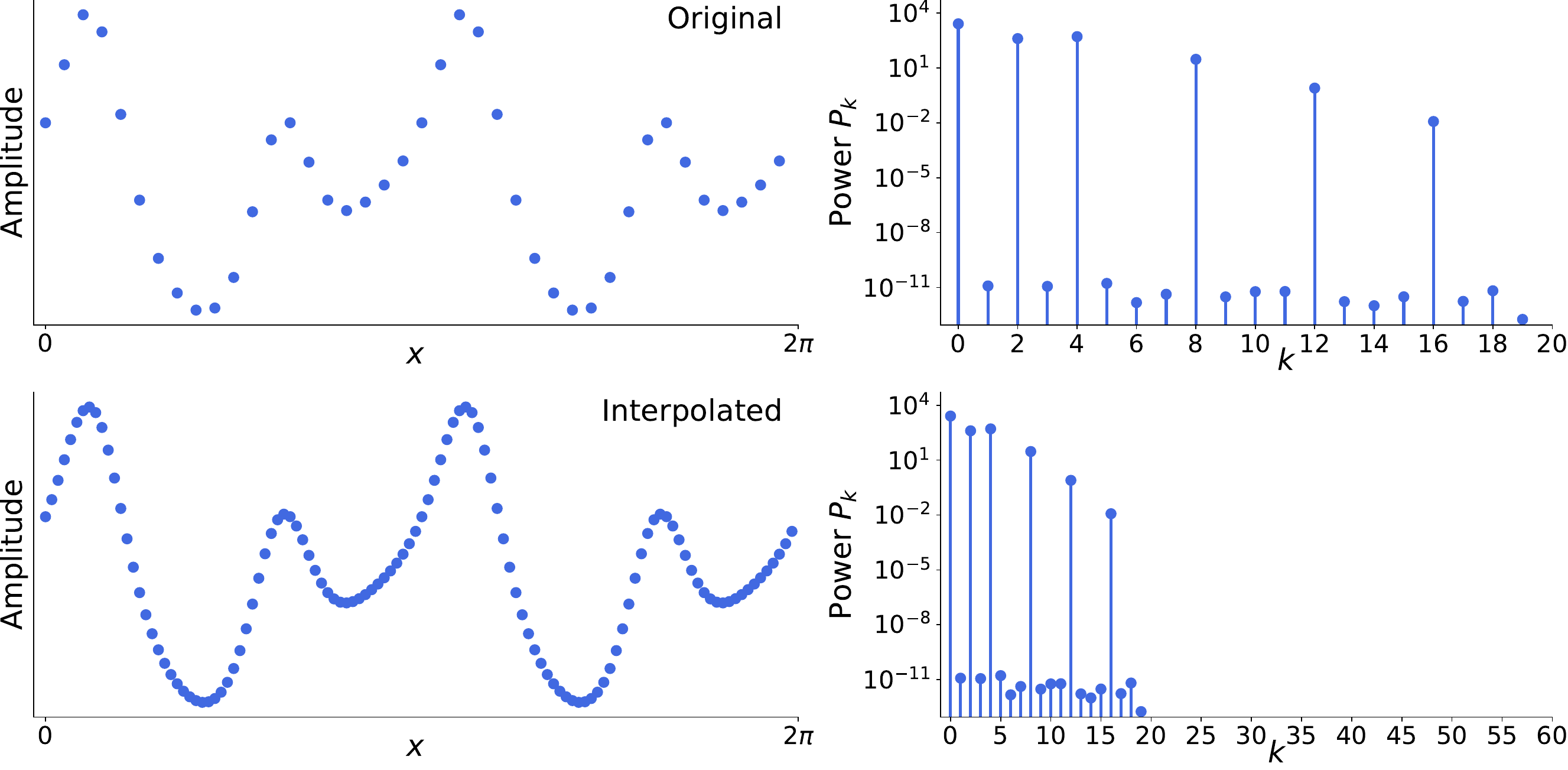}  \vspace{-6mm}
\caption{Spectral interpolation. The periodic signal $f(x) = \cos(2x) + \exp(\sin(4x))$ is sampled at $40$ points and its spectrum is computed (top). The signal is then reconstructed at a finer resolution with $120$ points by inserting zeros at the higher frequencies in the spectrum before performing the inverse DFT (bottom).}
\label{fig: Spectral Interpolation}  \vspace{6mm} 
\end{figure}

This process preserves the physically meaningful low-frequency content of the signal without introducing artificial high-frequency modes.  Finally, applying the inverse FFT of size $M$ to the padded spectrum reconstructs the signal at $M$ uniformly spaced points over the same spatial domain. The resulting signal represents the same periodic, band-limited function evaluated at a denser set of sample points, effectively increasing spatial resolution without altering the underlying frequency content. This interpolation process is illustrated in \Cref{fig: Spectral Interpolation}. 

\vspace{2mm}

Spectral interpolation on regular grids is implemented in \texttt{NeuralOperator 2.0.0} for three-dimensional signals in \codebox{neuralop/layers/resample.py}. For one-dimensional signals, spectral interpolation can be implemented as follows in \texttt{PyTorch}:

\vspace{1mm}

\begin{minted}[fontsize=\small, bgcolor=gray!5, frame=single, linenos]{python}
fft_coeffs = torch.fft.fft(signal, norm="forward")  # Compute FFT

# Pad zeros in the middle (between positive and negative frequencies)
pos_freqs = fft_coeffs[: N // 2 + 1]  # Non-negative frequencies
neg_freqs = fft_coeffs[N // 2 + 1 :]  # Negative frequencies
fft_coeffs_padded = torch.cat([pos_freqs, torch.zeros(N_new - N), neg_freqs])

interpolated_signal = torch.fft.ifft(fft_coeffs_padded, norm="forward").real  # Compute IFFT
\end{minted}

\vspace{4mm}

Zero-padding in the frequency domain therefore adds no new information but simply refines the sampling of the function with the same frequency content. This approach is widely used for smooth upsampling in spectral methods and FNOs, where increasing resolution corresponds to evaluating the learned spectral representation on a finer grid.

\clearpage

\subsubsection{Spectral Differentiation} \label{sec: Spectral Differentiation}

\vspace{2.5mm}

\mydef{Spectral differentiation}, also known as \mydef{Fourier differentiation}, provides a highly accurate and efficient way to compute derivatives of smooth, periodic functions by working in the frequency domain rather than in physical space.  It is particularly effective when the function is well-resolved on a uniform grid, as differentiation becomes an exact algebraic operation on the Fourier coefficients.  
This method achieves near machine precision for analytic functions, while requiring only a single forward and inverse DFT (or FFT on regular grids) for any differentiation order.

Consider a smooth periodic function $f : [0, 2\pi] \to \R$ sampled at the equispaced points $x_n = 2\pi n/N$, with DFT and IDFT given by
\begin{equation}
\widehat{f}_k = \frac{1}{\sqrt{N}} \sum_{n=0}^{N-1} f(x_n) \, e^{-i k x_n}, \qquad  \quad 
f(x_n) = \frac{1}{\sqrt{N}} \sum_{k=-N/2}^{N/2-1} \widehat{f}_k \, e^{i k x_n},
\end{equation}
for even $N$ (the odd-$N$ case is analogous). Differentiating term-by-term yields
\begin{equation} \label{eq: spectral derivative}
\partial_x^m f(x_n) = \frac{1}{\sqrt{N}} \sum_{k=-N/2}^{N/2-1} (i k)^m \, \widehat{f}_k \, e^{i k x_n}.
\end{equation}
Thus, differentiation in physical space corresponds to multiplication by powers of $i k$ in Fourier space,
\begin{equation}
\mathcal{F}\{\partial_x^m f\}(k) = (i k)^m \, \widehat{f}_k.
\end{equation}
Therefore, to compute the $m$-th derivative of $f$, one simply multiplies the Fourier coefficients of the function $f$ by $(i k)^m$ and applies the inverse DFT as in equation \eqref{eq: spectral derivative}. 

This approach naturally extends to multidimensional signals by applying the DFT along each dimension independently. A key advantage of spectral differentiation is its simplicity and high efficiency on regular grids, with a computational cost independent of differentiation order.  

A general spectral differentiation class with numerous options and routines, \codebox{FourierDiff}, is implemented in \texttt{NeuralOperator 2.0.0} in \codebox{neuralop/losses/differentiation.py}. A tutorial on how to use this class is provided at \href{https://neuraloperator.github.io/dev/auto\_examples/layers/plot_fourier_diff.html}{neuraloperator.github.io/dev/auto\_examples/layers/plot\_fourier\_diff.html}.

\vspace{2mm}

\paragraph{Spectral Numerical Solvers.} These properties of the discrete Fourier transform, which convert differentiation into multiplication and enable efficient computation through the FFT, form the basis of \emph{spectral solvers}.  In such methods, differential equations are reformulated in the frequency domain, where derivatives become algebraic operations and numerical accuracy can be achieved with relatively few modes. This approach leads to highly precise and highly efficient numerical schemes, especially for smooth and periodic problems, and represents one of the most elegant uses of Fourier analysis in computation.

\vspace{7mm}

\subsubsection{On the Periodic Assumption of Fourier Spectral Methods} \label{sec: Non-periodicity}

\vspace{2.5mm}

\begin{figure}[h]  
\centering   
\includegraphics[width=\textwidth]{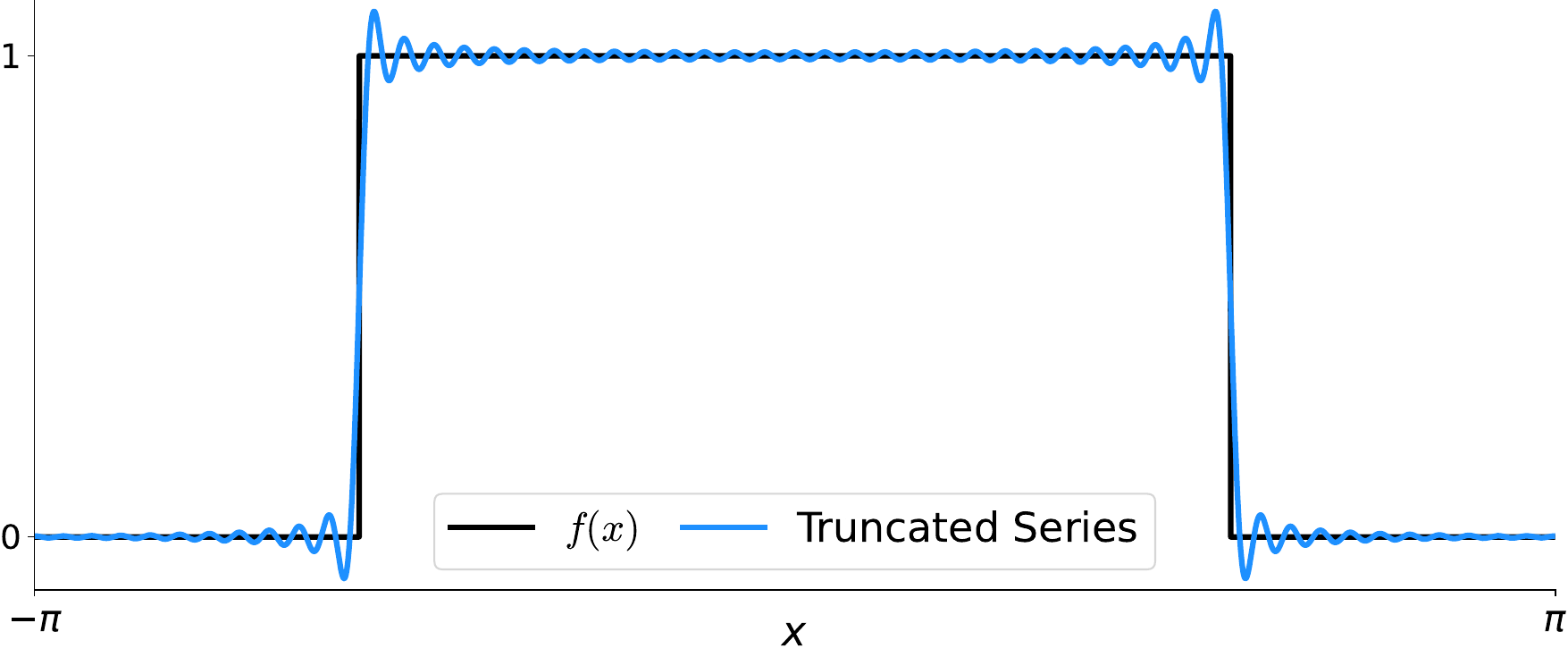}  \vspace{-6mm}
\caption{Illustration of the Gibbs phenomenon on a step function. The \(N=50\) truncated Fourier series reproduces the constant regions accurately but displays persistent overshoot and undershoot with ringing near the discontinuities.   
\label{fig: Fourier Series Example Gibbs}}  \vspace{3mm}
\end{figure}

Fourier spectral methods are inherently designed for periodic functions. Their mathematical foundation relies on Fourier expansions, which assume the function repeats identically beyond the domain boundaries. When this periodicity assumption does not hold, the implied periodic extension introduces a discontinuity at the boundaries. Applying spectral methods directly to such non-periodic data can thus produce artifacts such as the \emph{Gibbs phenomenon}, characterized by spurious oscillations and a significant loss of convergence accuracy in the Fourier series approximation. Figure~\ref{fig: Fourier Series Example Gibbs} illustrates the Gibbs phenomenon for a discontinuous step function. Although the truncated Fourier series matches the constant values well, it develops oscillatory ringing near the discontinuities. A discontinuity cannot be represented exactly with any finite number of modes, since a jump contains contributions from arbitrarily high frequencies and therefore requires an infinite Fourier series for exact reconstruction. Accordingly, the sharper the transition, the more high-frequency Fourier modes are needed to resolve it with a truncated series. Increasing $N$ confines the oscillations to a progressively smaller neighborhood of the jump, but the overshoot persists, so the approximation improves pointwise away from the discontinuity while still failing to converge uniformly at the jump.

\paragraph{Padding.} One straightforward approach to circumvent this issue and apply spectral methods to non-periodic signals is to embed the original non-periodic signal within a larger domain and define its extension so that it becomes periodic on the extended domain. Spectral methods can then be applied to the signal on the larger domain.

\emph{Zero-padding}, which appends zeros to both ends of the signal, can be helpful and sufficient in certain settings, but can also introduce artificial discontinuities at the edges. When the Fourier transform is applied to such data, these discontinuities can still produce \emph{Gibbs oscillations} that distort the spectral representation and reduce accuracy. 

An alternative strategy is \emph{mirror-padding}, where the signal is extended by reflecting it about one or both endpoints. This is typically better and guarantees continuity of the extended signal, but now introduces discontinuities in its derivatives unless the function happens to be locally constant at the reflection points. Because the convergence rate of Fourier series depends on smoothness, the lack of differentiability at these mirror points leads to slower spectral convergence and residual artifacts in the frequency spectrum. 

\begin{figure}[h]   \vspace{5mm}
\centering   
\includegraphics[width=0.9\textwidth]{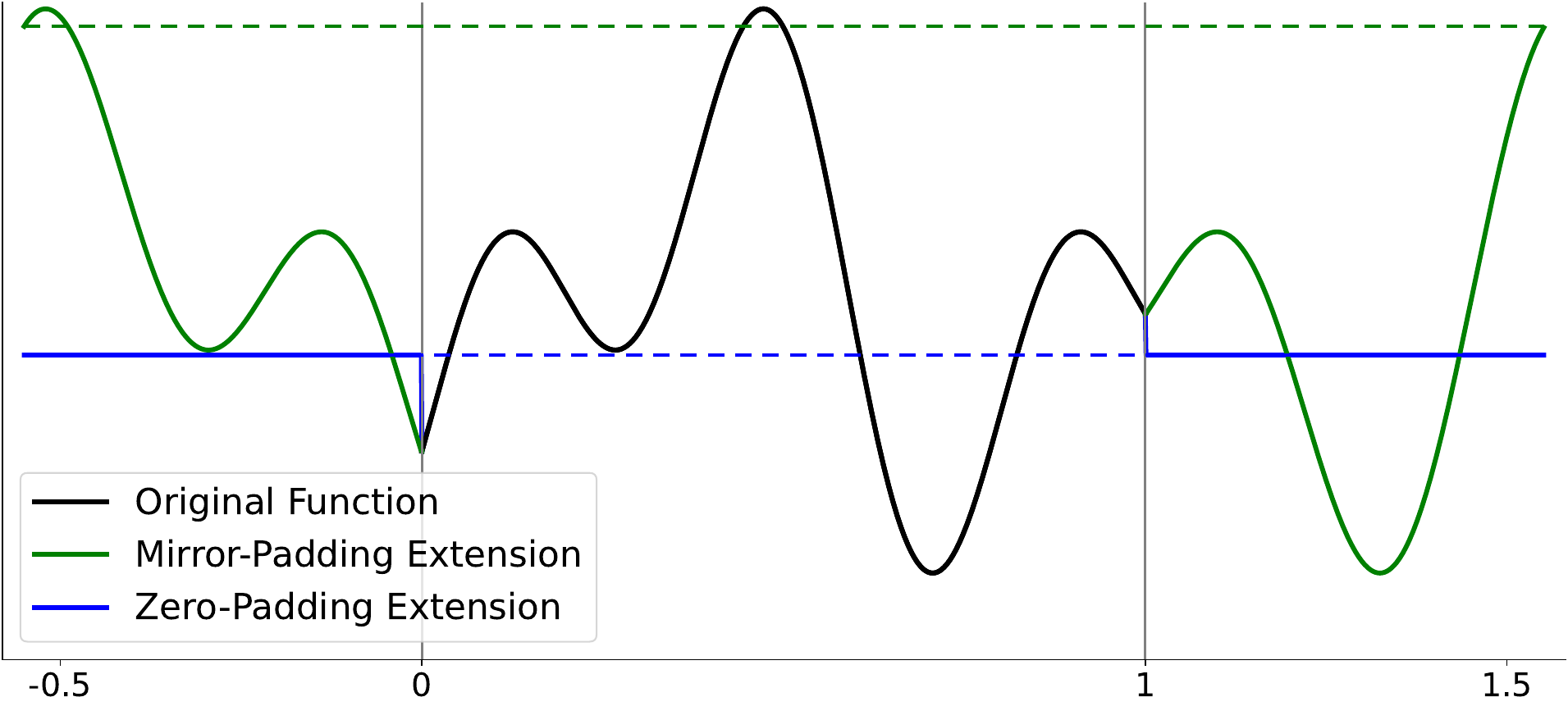} 
\caption{Example of periodic extensions of the function $f(x) = \sin(16x) - \cos(8x) + 0.4$ obtained using zero-padding and mirror-padding. 
\label{fig: Padding} } \vspace{6mm}
\end{figure}

Zero-padding and mirror-padding are illustrated in \Cref{fig: Padding}.  

\paragraph{Fourier Continuation.} A more principled solution is to construct a smooth, periodic extension of the non-periodic signal.  
\mydef{Fourier continuation} methods~\citep{Fourier_continuation_2011, fontana2020Fourier, bruno2022two} address this problem by extending a non-periodic function to a periodic one over an enlarged domain using smooth polynomial or spline-based interpolation.  
The resulting continuation preserves both the continuity and differentiability of the function across the artificial boundary, allowing Fourier-based techniques to achieve convergence even for originally non-periodic data. In this way, Fourier continuation provides a bridge between the mathematical elegance of spectral methods and the practical need to handle general, non-periodic problems reliably.

\begin{figure}[t]  \vspace{6mm}
\centering   \
\includegraphics[width=0.93\textwidth]{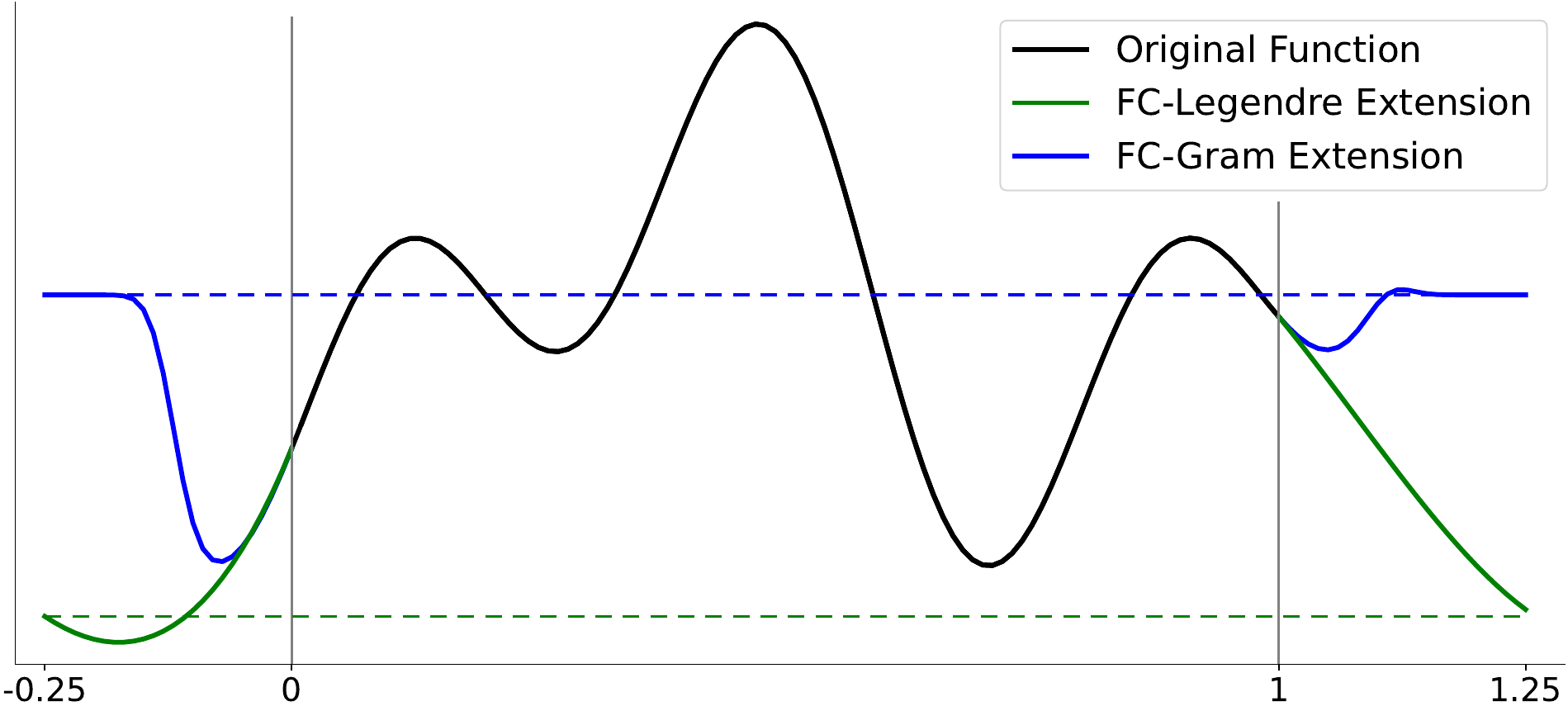} 
\caption{Example of periodic extensions of the function $f(x) = \sin(16x) - \cos(8x)$ obtained using the Fourier continuation methods to extend it from the interval $[0,1]$ to $[-0.25,1.25]$. 
\label{fig: FC Examples} }  \vspace{4mm}
\end{figure}

We refer the reader to \citep{ganeshram2025fcpinohighprecisionphysicsinformed} for a detailed description of two such Fourier continuation methods, \textsc{FC--Legendre} and \textsc{FC--Gram}. In summary, these two continuation methods proceed as follows.  Let \(f:[a,b]\to\mathbb{R}\) be given on a grid of \(n\) points, with equidistant nodes $x_k$ on $[a,b]$, and denote $f_k = f(x_k)$. The goal is to produce a new sequence~\(\tilde f\) of length \((n+c)\) which is exactly periodic of period \((n+c)\) and agrees with \(f\) on the original interval. Define the left and right boundary vectors of width \(d\),
\begin{align}
f_\ell = (f_0, f_1, \ldots, f_{d - 1})^\top, \quad f_r = (f_{n - d}, \ldots, f_{n-1})^\top.
\end{align}
\textsc{FC--Legendre} and \textsc{FC--Gram} construct an extension signal $\textcolor{NavyBlue}{f_{\text{ext}}} \in \mathbb{R}^c$ using only the information contained in the boundary vectors $f_{\ell}$ and $ f_r$ of width \(d\), and the periodic extended signal $\tilde f$ is then obtained by concatenating the original signal $f$ and the extension $\textcolor{NavyBlue}{f_{\text{ext}}}$ on both sides, 
\begin{align}
\tilde f & =  (\textcolor{NavyBlue}{f_{\text{ext},c/2},\dots,f_{\text{ext},c-1}},\,f_0,\dots,f_{n-1},\,\textcolor{NavyBlue}{f_{\text{ext},0},\dots,f_{\text{ext},c/2-1}}).
\end{align}
For functions in higher dimensions, the one-dimensional extension is applied along each direction iteratively using the same extension matrices, i.e. first along the first dimension (left/right), then along the second dimension (top/bottom), and so on. 

\vspace{2mm}

In \textsc{FC--Legendre}, the left and right boundary vectors are concatenated into the vector
$y = (f_{r}^\top, f_{\ell}^\top)^\top \in \mathbb{R}^{2d}$, and the extension $\textcolor{NavyBlue}{f_{\text{ext}}} \in \mathbb{R}^c$ is obtained via $ E y = \textcolor{NavyBlue}{f_{\text{ext}}} \in \mathbb{R}^c$. Here, $E \in \mathbb{R}^{c \times 2d}$ is an extension matrix, built from evaluations of shifted Legendre polynomials of degree \((2d-1)\), to perform polynomial interpolation between the right and left boundaries. In \textsc{FC--Gram}, the extension values $\textcolor{NavyBlue}{f_{\text{ext}}}$ are formed by projecting the left and right boundary vectors $f_\ell, f_r$ onto Gram bases via matrices \(Q_\ell\), \(Q_r\), and then blending their respective continuations using the continuation matrices \(A_\ell\), \(A_r\), i.e., $\textcolor{NavyBlue}{f_{\text{ext}}} = A_\ell Q_\ell^\top f_\ell + A_r Q_r^\top f_r.$

\textsc{FC--Legendre} constructs the extension by fitting a single Legendre polynomial of degree \( (2d -1) \) spanning the entire extended interval between the left and right boundaries. In contrast, \textsc{FC--Gram} takes a more local approach, projecting boundary data onto lower-order polynomial bases of degree \(d\) separately on each boundary and then smoothly blending the extensions outside the domain. Examples of \textsc{FC--Legendre} and  \textsc{FC--Gram} extensions are displayed in \Cref{fig: FC Examples}. 

A Fourier continuation class \codebox{FourierContinuation} and subclasses \codebox{FCLegendre} and \codebox{FCGram} are implemented in \texttt{NeuralOperator 2.0.0} in \codebox{neuralop/layers/fourier\_continuation.py}. A tutorial on how to use these classes is provided at \href{https://neuraloperator.github.io/dev/auto_examples/layers/plot_fourier_continuation.html}{neuraloperator.github.io/dev/auto\_examples/layers/plot\_fourier\_continuation.html}.

\begin{figure}[h]  \vspace{-2mm}
\centering   
\includegraphics[width=0.92\textwidth]{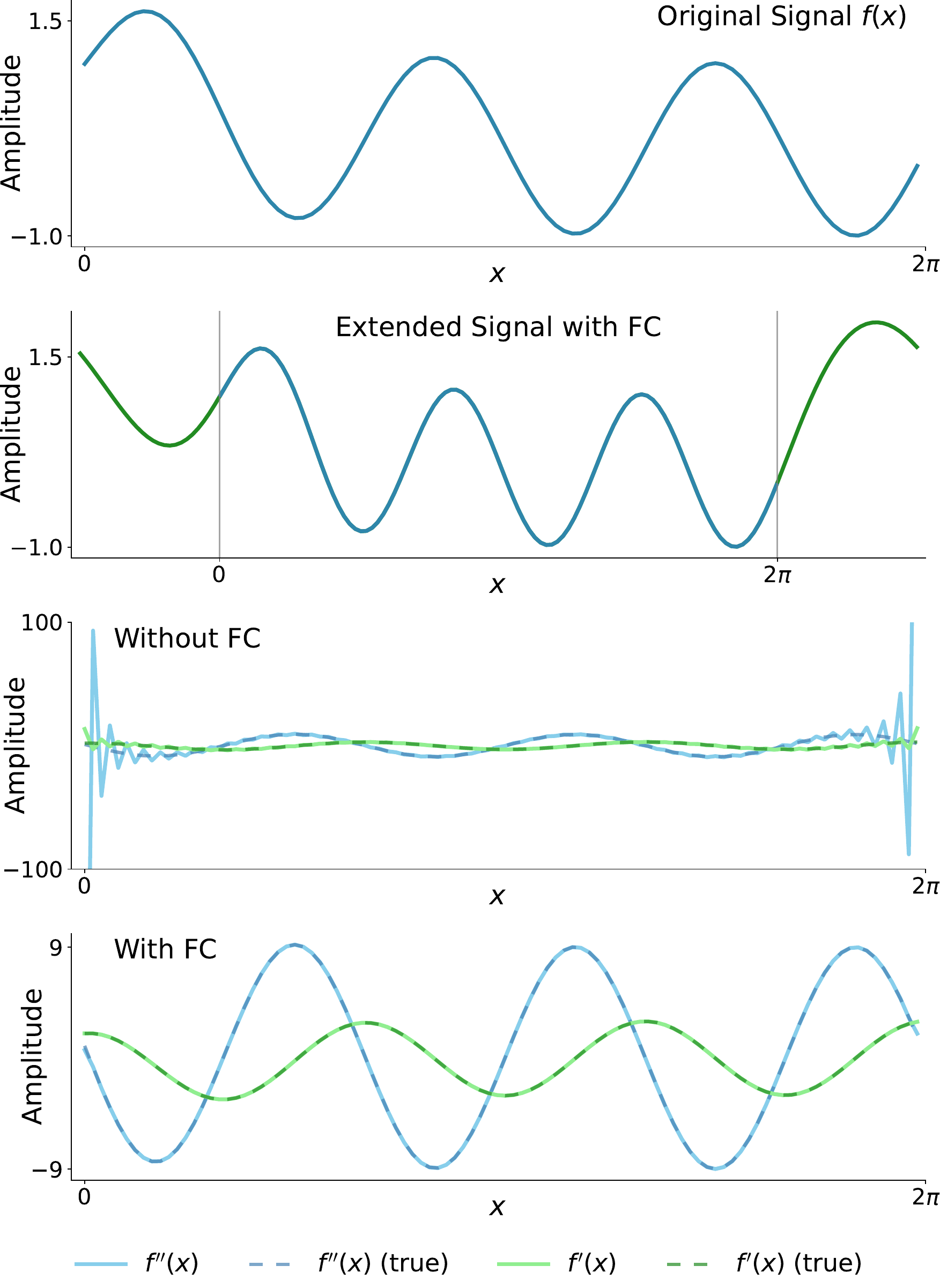}  \vspace{-0mm}
\caption{Spectral differentiation of the non-periodic signal $f(x) = e^{-x} + \sin(3x)$. 
The third and fourth panels display the first and second derivatives obtained using spectral differentiation without (third panel) and with (fourth panel) Fourier continuation. The dashed curves indicate the analytical derivatives. Without Fourier continuation, strong oscillatory artifacts appear near the boundaries, typical of the Gibbs phenomenon. }
\label{fig: Spectral Differentiation}   \vspace{-18mm}
\end{figure}

Fourier continuation combined with higher-precision arithmetic is highly effective for computing accurate spectral derivatives of non-periodic signals. In this approach, spectral differentiation is performed on the extended periodic domain produced by the Fourier continuation procedure. The resulting improvement in accuracy for spectral derivatives of non-periodic data is illustrated in \Cref{fig: Spectral Differentiation}. It should be noted that spectral differentiation and Fourier continuation are highly accurate mainly for noise-free signals. In the presence of noise, differentiation becomes ill-conditioned, so high-frequency noise is strongly amplified and accuracy degrades. To limit this effect, spectral differentiation is typically combined with low-pass filtering, which suppresses noise but also reduces the accuracy of the computed derivatives.

\paragraph{Extension through Spectrum Optimization.}  A smooth periodic extension of a given non-periodic signal can also be constructed by optimizing its spectral content on the extended domain. The grid is extended by adding a small number of unknown points to the left and right of the original interval, and the values at these new grid points are determined by minimizing the discrete Sobolev spectral \(H^s\) norm
\begin{equation}
\|u\|_{H^s}^2 \coloneqq \sum_{k} \bigl(1 + |k|^{2}\bigr)^s\, \bigl|\hat{u} (k)\bigr|^2 ,
\end{equation}
where $\hat{u} = \mathcal{F}(u)$. This norm measures the energy of \(u\) in all derivatives up to order \(s\), since in Fourier space the \(j\)-th derivative corresponds to multiplication by \((ik)^j\), and the factor \(\bigl(1 + |k|^{2}\bigr)^s\) therefore emphasizes high-frequency content and high-order derivatives.

\begin{figure}[t]  \vspace{7mm}
\centering   \
\includegraphics[width=\textwidth]{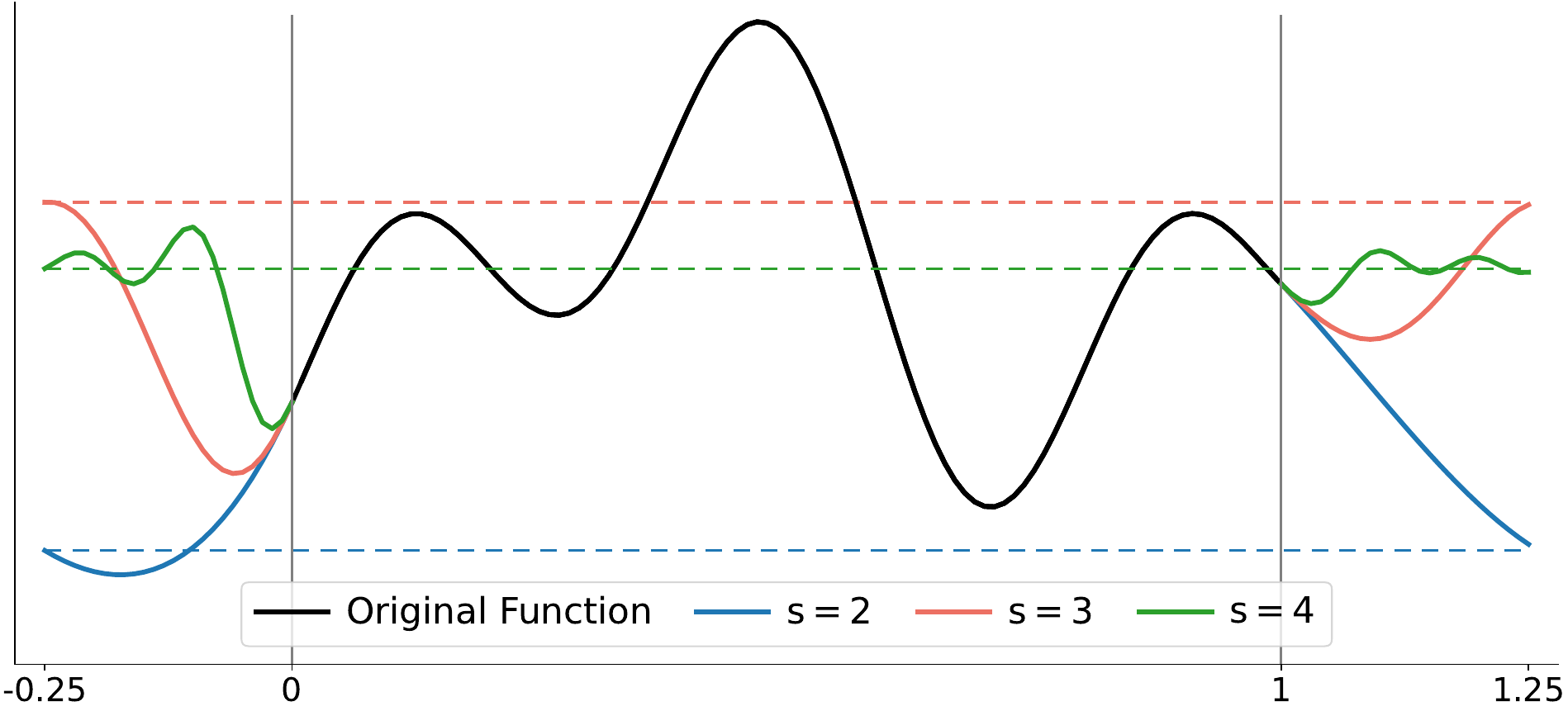} 
\caption{Example of periodic extensions of the function $f(x) = \sin(16x) - \cos(8x)$ obtained using spectrum optimization with different values of $s$ for the Sobolev spectral $H^s$ norm, to extend it from the interval $[0,1]$ to $[-0.25,1.25]$. 
\label{fig: spectral extension} }  \vspace{12mm}
\end{figure}

Concretely, consider as before a function \(f:[a,b]\to\mathbb{R}\)  given on a grid of \(n\) equidistant nodes \(x_k\) with samples \(f_k = f(x_k)\), and let \(\tilde f \in \mathbb{R}^{n+c}\) denote the periodic extended sequence defined by
\[
\tilde f = \bigl(\textcolor{NavyBlue}{f_{\text{ext},c/2},\dots,f_{\text{ext},c-1}},\,f_0,\dots,f_{n-1},\,\textcolor{NavyBlue}{f_{\text{ext},0},\dots,f_{\text{ext},c/2-1}}\bigr),
\]
where the extension \(\textcolor{NavyBlue}{f_{\text{ext}}} \in \mathbb{R}^c\) is obtained by minimizing \(\|\tilde f\|_{H^k}^2\) . Intuitively, this selects among all periodic continuations the one that is spectrally smooth, in the sense of having minimal energy in its low-order and high-order derivatives, while exactly preserving the original data in the interior. Examples of periodic extensions using spectrum optimization with different values of $s$ for the Sobolev spectral $H^s$ norm are displayed in \Cref{fig: spectral extension}.

In practice, the discrete Fourier transform is linear and the spectral \(H^s\) functional is quadratic in \(\tilde f\), so the resulting optimization problem for \(\textcolor{NavyBlue}{f_{\text{ext}}}\) is an ordinary linear least-squares problem that admits a closed-form solution and can be implemented with computational cost comparable to \textsc{FC--Legendre} and \textsc{FC--Gram}. 

An important consequence of the spectral $H^s$ regularization is that high-frequency components are penalized more heavily and systematically suppressed, so the constructed extension tends to be more robust to noise than  \textsc{FC--Legendre} and \textsc{FC--Gram}, especially when the noise manifests predominantly in the high-frequency part of the spectrum.

\clearpage

\clearpage 

\newpage

\subsection{Dataset Generation and Data Manipulation}
\label{sec: Dataset Generation and Data Manipulation}
\vspace{3mm}

\subsubsection{Defining the Mapping of Interest}
\label{sec: Mapping of Interest}

\vspace{3mm}

Neural operators are principled extensions of neural networks that aim to approximate operators, that is, mappings between function spaces rather than finite-dimensional vectors. Formally, an operator maps an input function \( f \in \mathcal{F}_{\text{in}} \) to an output function \( g \in \mathcal{F}_{\text{out}} \). This abstraction is particularly relevant in scientific computing and physics-informed learning, where many dynamical systems can be expressed as operators mapping functions to PDE solutions.

For operator learning to be mathematically meaningful, the target mapping must be well-posed in the following sense: each admissible input function \( f \in \mathcal{F}_{\text{in}} \) must correspond to a unique output \( g  \in \mathcal{F}_{\text{out}} \). If the mapping is not unique, i.e. multiple valid outputs exist for the same input, a deterministic neural operator cannot represent it consistently. In such cases, the model tends to average across possible outputs, which often leads to blurred or unstable predictions and poor generalization. When non-uniqueness is inherent to the problem, alternative probabilistic formulations or ensemble-based neural operators are required to capture the distribution of possible outputs.

\vspace{1mm}

Although neural operators are theoretically general, several misunderstandings persist regarding their flexibility and range of application, particularly in the case of Fourier Neural Operators. With suitable embeddings and resolution-invariant transformations, neural operators can address a broad range of tasks that extend well beyond their original goal of function to function regression. A detailed discussion of embedding architectures and resolution-invariant representations is provided in \Cref{sec: Embeddings}, which elaborates on how these mechanisms enhance the flexibility of neural operators. Here, we refute some of the most common misconceptions regarding the inputs and outputs of neural operators.

\vspace{3mm}

\textbf{Misconceptions about the inputs:} \\
\textit{``Neural operators can only process a single input function, cannot incorporate scalar values or discrete finite-dimensional inputs, and cannot handle inputs with heterogeneous domains or data types.''}  

\vspace{0mm}

These are all misconceptions! \\
Neural operators can process multiple input functions and quantities jointly. Each input can be encoded using a separate resolution-invariant embedding strategy and fused through concatenation, allowing the model to learn complex cross-variable relationships. Neural operators can also incorporate scalar or low-dimensional quantities such as physical constants, geometric descriptors, or control variables by embedding them into latent spaces and combining them with functional representations. Furthermore, through appropriate domain embeddings, neural operators can integrate heterogeneous data defined on different geometries, coordinate systems, or modalities, for example by combining structured and unstructured spatial data within a single latent representation. The key requirement is that the neural operator receives as input all quantities that vary across simulations in the dataset.

\vspace{7mm}

\textbf{Misconceptions about the outputs:} \\
\textit{``Neural operators can only output functions, require identical input and output resolutions, and can only map between functions of the same dimensionality.''}  

\vspace{0mm}

These are all misconceptions! \\
Neural operators can produce scalar, vector, or categorical outputs by applying suitable resolution-invariant transformations (such as projections, pooling, or integration operations) to intermediate or final representations. This flexibility enables applications such as function to scalar regression, classification based on functional inputs, and optimization tasks where the target quantity is not itself a function. Neural operators can also learn mappings across function spaces of different dimensions, for example transforming one-dimensional signals into two-dimensional fields or mapping spatial functions to temporal responses. Finally, since neural operators act in function spaces rather than on discretized grids, they can evaluate outputs at arbitrary discretizations independently of the input discretization. This resolution invariance fundamentally distinguishes operator learning and neural operators from function learning and conventional neural networks.

\vspace{12mm}

\subsubsection{Dataset Sources and Data Generation}
\label{sec: Dataset Sources and Data Generation}

\vspace{3.5mm}

An important advantage of neural operators is their resolution invariance, which allows them to learn from data generated at different spatial and temporal discretizations. This property makes it easier to combine datasets from heterogeneous sources, such as solvers with distinct grid resolutions or experimental measurements collected at varying sampling densities, within a single unified framework. Consequently, the overall dataset can be enriched both in size and in diversity without requiring strict alignment across resolutions or domains. As a result, training data for neural operators can originate from a wide range of sources, including experimental measurements, numerical simulations from a single solver, numerical simulations from multiple solvers, or combinations of those. 

The simplest case involves data generated from a single numerical solver that approximates the governing equations of a physical system. Each simulation produces input-output pairs that serve as samples of the underlying operator which can be used for training the neural operators. When the data are obtained from a numerical solver, it is crucial to keep in mind that the resulting dataset reflects the assumptions, simplifications, and numerical errors inherent to the solver. A neural operator trained exclusively on such synthetic data effectively learns to reproduce the behavior of the solver rather than the true underlying physical process. As a result, the trained model will inherit the solver's biases and cannot, in general, be expected to outperform it without access to additional information. 

The fidelity and generalization of neural operators can sometimes be improved by broadening or enriching the available training data. One strategy is to combine data from multiple solvers that rely on different numerical schemes, discretization strategies, or modeling assumptions, allowing the neural operator to infer a more robust and universal functional relationship. Neural operators can also be trained using multi-resolution strategies, where data sampled at different discretizations are jointly incorporated during training. This enables the use of solvers with varying levels of fidelity, from coarse low-cost simulations to fine high-accuracy ones, and can substantially reduce the need for large amounts of high-resolution high-fidelity data while still achieving accurate and generalizable operator learning. Another approach is to embed physical knowledge directly into the training process through physics-based constraints or regularization terms that enforce known conservation laws or symmetries. Alternatively, supplementing synthetic data with experimental observations can help anchor the learned operator to physical reality and correct biases introduced by synthetic simulations.

\hfill \\ 

\subsubsection{Dataset Validation and Preprocessing}
\label{sec: Data Cleaning}

\vspace{3mm}

High-quality and well-curated data are essential for the successful training of neural operators. Inconsistencies, noise, or biases in the dataset can significantly degrade their performance and stability. The preparation of data can therefore sometimes involve a careful sequence of cleaning, normalization, denoising, filtering, and augmentation steps, all of which aim to improve data consistency, enhance model generalization, and reduce the risk of overfitting or learning spurious correlations.

\vspace{4mm}

\paragraph{Data Cleaning.} The cleaning stage focuses on detecting and correcting corrupt, missing, or inconsistent samples that may arise from numerical instabilities, incomplete experiments, or measurement errors. In simulated data this may involve removing samples with non-physical values, inconsistent boundary conditions, or solver failures, whereas for experimental data it often requires outlier detection, interpolation of missing measurements, or correction of calibration drifts. Maintaining consistent conventions across all data sources is critical, particularly when combining datasets generated by different solvers or measurement systems.

\vspace{4mm}

\paragraph{Normalization.} Normalization plays a crucial role in stabilizing the training process. Features can be normalized globally across the dataset or locally within each sample. Common strategies include min-max scaling, standardization by the mean and standard deviation, or normalization with respect to characteristic physical quantities such as reference velocity, pressure, or temperature. Proper normalization ensures that all variables contribute proportionally to the loss function and that the optimization process is not dominated by those with large numerical magnitudes. When multiple data sources are combined, normalization also helps align their statistical distributions, which improves stability during joint training.

\vspace{4mm}

\paragraph{Data Augmentation.} Data augmentation is another important aspect of data preparation. By artificially increasing the diversity of the training data, augmentation can improve model robustness and generalization. Transformations that preserve the underlying physics, such as spatial translations, rotations, reflections, or temporal shifts, can be applied when the system exhibits the corresponding symmetries. Synthetic perturbations may also be introduced by adding controlled stochastic noise, allowing the model to become more resilient to small variations. Neural operators can exploit their resolution invariance for data augmentation by leveraging multi-resolution training. Through appropriate downsampling strategies, high-resolution data can be systematically reduced to generate additional lower-resolution samples that capture the same underlying physical behavior. These augmented datasets expose the model to inputs and outputs at multiple discretizations, improving robustness to changes in resolution and enhancing generalization. This approach effectively increases dataset diversity without requiring additional expensive high-fidelity simulations, allowing the model to learn consistent representations across scales.

\vspace{4mm}

\paragraph{Bias Correction.} When datasets are assembled from heterogeneous sources, such as a combination of simulations from different solvers and experimental data, systematic biases may arise due to differences in numerical schemes, modeling assumptions, or measurement procedures. Careful bias correction and consistency checks are essential to ensure that the model does not learn solver-specific artifacts but rather captures the true underlying physical relationships.

\paragraph{Challenges with Experimental Data.} Real-world measurements are typically noisy, incomplete, or affected by calibration uncertainties, which can obscure the deterministic mapping between inputs and outputs and introduce effective non-uniqueness in the data. Addressing this issue requires the use of denoising strategies, uncertainty quantification, or probabilistic formulations of neural operators so that the model can represent variability while maintaining stability and physical consistency. Denoising can be performed before training through spatial or temporal smoothing, low-pass filtering, or wavelet-based thresholding.

\hfill 

\subsubsection{Downsampling Strategies}
\label{sec: downsampling}

\vspace{2mm}

When downsampling data, the chosen strategy can have major implications for data quality and spectral integrity. Each method discards information differently, altering both spatial correlations and the frequency spectrum of the underlying fields. The choice of approach must therefore reflect the characteristics of the problem, the physical nature of the quantities involved, and the intended downstream tasks.

Projecting high-resolution data onto a coarser grid inevitably removes high-frequency information. In spectral terms, downsampling acts as a frequency truncation or folding process. When the sampling rate falls below the Nyquist limit, unresolved high-frequency components are aliased into lower frequencies, introducing spurious oscillations and distorting the true energy distribution across scales. Such distortions alter the physical interpretation of the data, as the power spectrum no longer reflects the genuine hierarchy of spatial or temporal modes. In neural operator learning, these artifacts can propagate through the model, leading to inaccurate reconstructions, unphysical oscillations, and degraded super-resolution performance. Controlling these effects is therefore essential to preserve the integrity of both the data and the learned operator.

\begin{figure}[h]  
\centering   
\includegraphics[width=\textwidth]{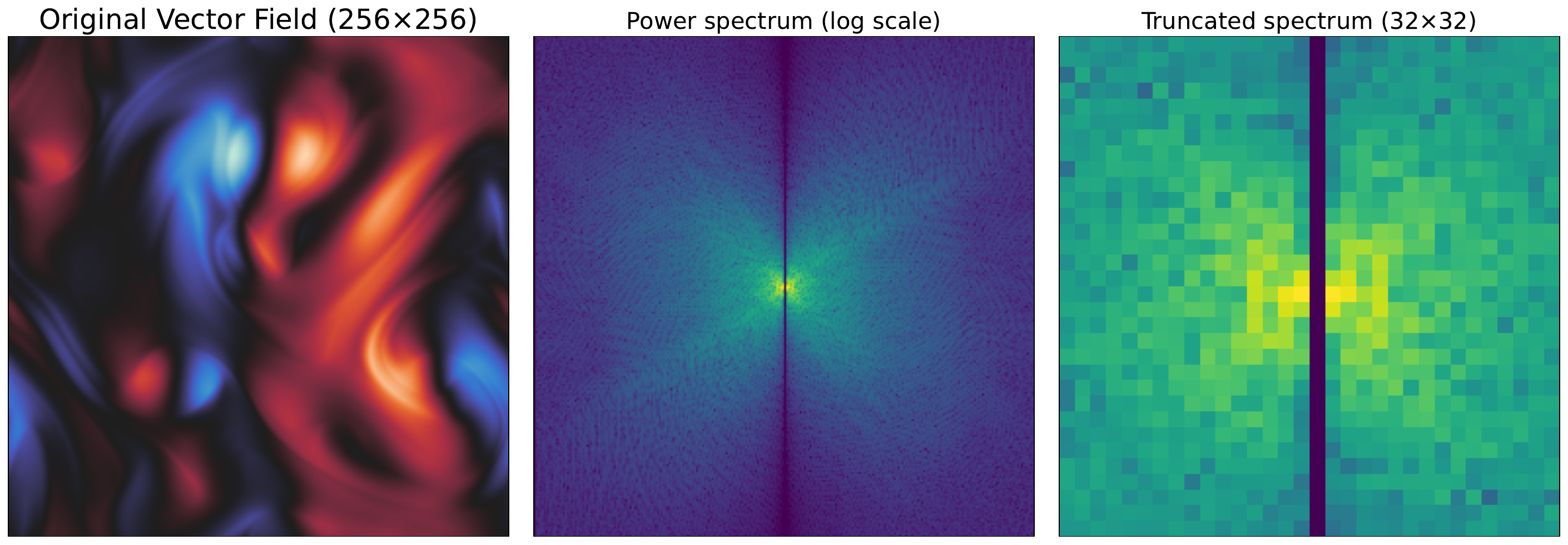}
\includegraphics[width=\textwidth]{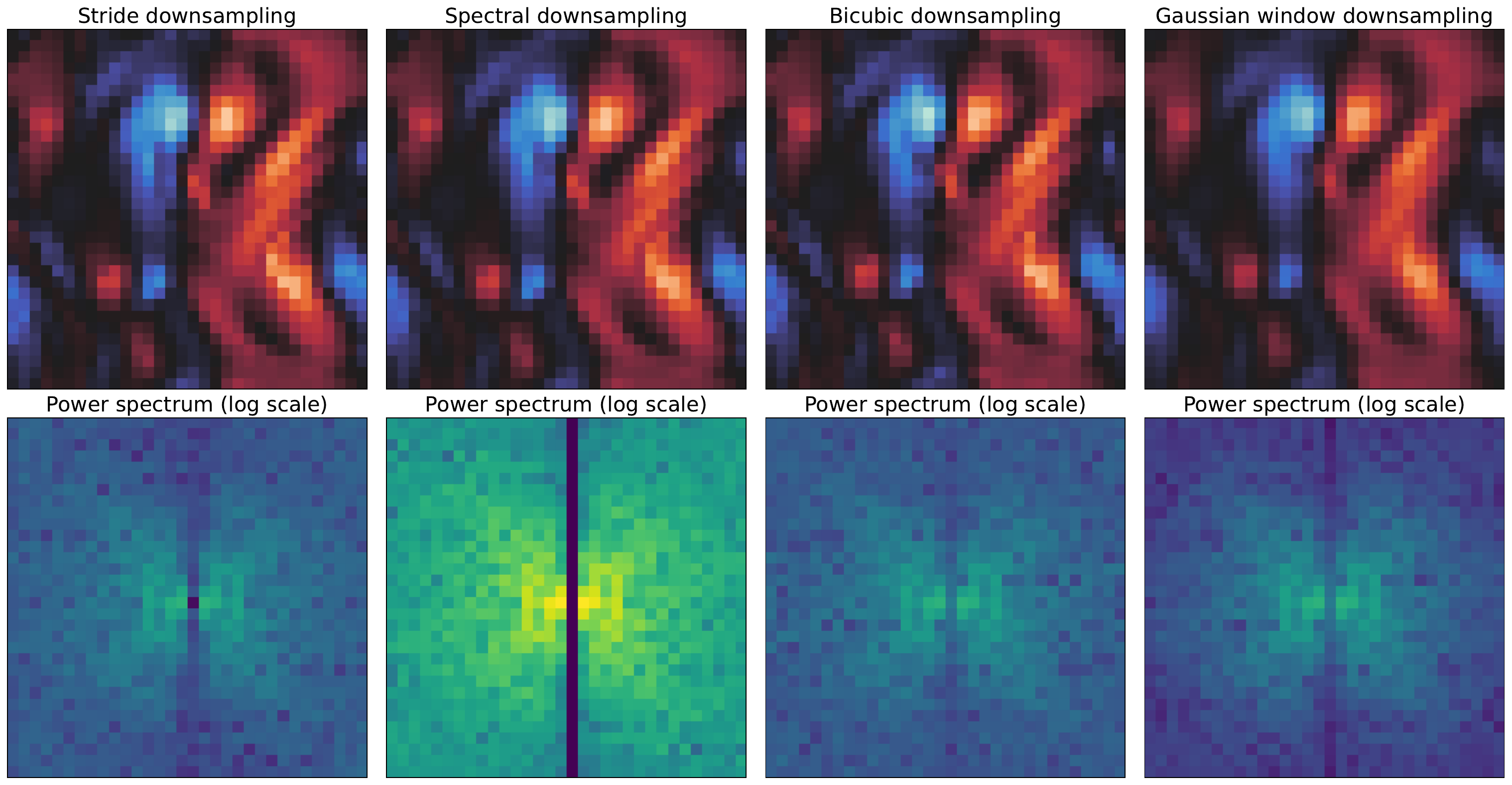}
\vspace{-5mm}
\caption{Visualizing downsampling effects. The first row presents the 256$\times$256 $x$-component of a turbulent vector field, its complete power spectrum, and the truncated 32$\times$32 central power spectrum. The second and third rows compare four downsampling methods: stride, spectral truncation, bicubic interpolation, and Gaussian filtering, highlighting their distinct effects on spatial coherence and spectral energy preservation.}
\label{fig: downsampling}   \vspace{4mm}
\end{figure}

The simplest downsampling method is \emph{stride-based downsampling}, where every \( s \)-th sample along each dimension is retained. Although computationally efficient, this approach is highly prone to aliasing, since it removes data points without smoothing. The abrupt loss of high-frequency content can introduce discontinuities, phase shifts, and spurious patterns, especially in regions with steep gradients or oscillatory behavior. Such distortions can mislead the model by providing incomplete or spectrally inconsistent samples.

In \emph{spectral downsampling}, the field is transformed into the Fourier domain, high-frequency modes are truncated according to the target resolution, and the remaining modes are used to reconstruct the signal on a lower-resolution grid. This is equivalent to applying an ideal \emph{low-pass filter} before doing spectral interpolation onto a lower-resolution grid, producing smooth, alias-free fields that preserve the correct energy distribution across modes. However, this method assumes periodicity and may introduce boundary artifacts when applied to non-periodic data naively (see \Cref{sec: Non-periodicity} for a discussion of non-periodicity in the context of spectral methods). Low-pass filtering can also precede stride-based downsampling to reduce aliasing, as removing frequencies beyond what the target resolution can resolve improves spectral fidelity and smoothness. 

Among alternative approaches, \emph{interpolation} methods such as bilinear, bicubic, or spline interpolation ensure smoothness but attenuate fine-scale variability. \emph{Pooling}, by contrast, aggregates information from local neighborhoods: mean pooling acts as a low-pass filter that retains large-scale trends, while max pooling emphasizes localized extremes but can distort global patterns. \emph{Wavelet-based downsampling} offers a structured multiscale alternative, by decomposing the signal into localized basis functions with both spatial and frequency resolution, allowing selective retention of coarse coefficients while discarding fine details. This produces a hierarchical representation that supports reconstruction across scales and separates meaningful structures from noise, making it particularly effective for systems with inhomogeneous features such as turbulence or fracture processes. \emph{Window-function-based downsampling} approaches apply spatial weighting before resampling, combining local averaging with spectral smoothing. Narrow windows enhance spatial precision but filter weakly, while broader windows improve low-pass characteristics at the cost of spatial detail. Such methods are especially useful for finite or non-periodic domains.

Overall, downsampling is a modeling decision that can strongly affect the information content and spectral characteristics of the data. Each strategy distorts information differently, as illustrated in \Cref{fig: downsampling}, and the optimal choice is problem dependent. While simple stride-based approaches may suffice for simpler analyses, spectral or wavelet-based methods may sometimes be required to preserve physical structure and spectral accuracy. A carefully designed downsampling pipeline, aligned with the physics and objectives of the problem, is essential for achieving robust, generalizable neural operators across resolutions.

\clearpage 

\subsubsection{Data Visualization}
\label{sec: Data Visualization}

\vspace{3.5mm}

Data visualization plays a central role in the analysis, validation, and interpretation of models. It is essential not only for understanding the structure of the training data but also for assessing the fidelity, stability, and generalization of the learned operator. Well-designed visualizations help identify systematic errors, evaluate physical consistency, and reveal how effectively the model captures patterns across multiple scales.

\vspace{3.5mm}

\paragraph{Field Visualization.} Streamline and quiver plots are useful for illustrating flow direction and coherence, while contour plots, surface plots, and colormaps can represent magnitudes or derived quantities. The choice of colors can affect interpretability, and ideally it should ensure consistent contrast and avoid perceptual distortions. In some cases, separating magnitude and directional information provides clearer insight into structural and energetic features of the field. When quantities span several orders of magnitude, logarithmic or normalized scales can be employed to enhance contrast and make subtle variations visible. For models trained on multi-resolution data, visual comparisons of predictions and errors across discretizations are useful for assessing how well resolution-invariance is maintained. 

\vspace{3.5mm}

\paragraph{Power Spectrum Visualization.} Examining the power spectrum of the input and output fields can also be a very valuable diagnostic tool. The power spectrum characterizes how energy is distributed across spatial or temporal frequencies, revealing the range of scales represented in the data. Comparing the power spectra of predicted and reference fields indicates whether the model reproduces both large-scale and fine-scale features, and can also guide hyperparameter selection, for example the number of Fourier modes in Fourier Neural Operators.

\vspace{3.5mm}

\paragraph{Error Visualization.} Error visualization, from absolute or relative differences, complements field analysis by providing spatial insight into where the predictions deviate from reference data the most. Displaying these maps in linear and logarithmic scales reveals both localized and global errors, while overlaying error contours on physical fields helps identify whether prediction inaccuracies occur near sharp gradients, boundaries, or small scale structures. Differences in spectral content between predicted and reference fields may expose systematic tendencies such as excessive smoothing of high frequency components or amplification of numerical noise, offering guidance for improving the model architecture, data processing, or learning strategy.

\vspace{3.5mm}

\paragraph{PDE Residuals Visualizations.} Visualization can also be extended to physical consistency checks through the evaluation of PDE residuals and conservation law errors. Plotting the spatial distribution of PDE residuals reveals regions where the learned operator violates governing equations such as mass continuity, energy conservation, or momentum balance. These residual maps can be visualized alongside physical or error fields to uncover correlations between model inaccuracies and physical inconsistencies. Examining residuals across time, space, or training stages provides direct evidence of whether the model has internalized the correct physical behavior or merely replicates empirical patterns in the training data.

\clearpage

\section{Neural Operators} \label{sec: NOs}

\subsection{Introduction to Neural Operators} \label{sec: Intro to NOs}

\paragraph{Neural Networks vs. Neural Operators.} Classical neural networks approximate functions between finite-dimensional Euclidean spaces, which map input vectors to output vectors. In contrast, neural operators are principled extensions of neural networks aiming to approximate operators, which are transformations acting on functions to produce new functions, a more general and complex task. This is especially relevant in scientific computing, where many problems can be expressed as operators acting on functions. Examples include dynamical systems governed by PDEs, in which the operator represents the PDE solution map that associates an input function, such as an initial condition or boundary condition, with the corresponding solution function. By adopting this continuous perspective, operator learning provides the flexibility to model and learn mappings between functions, thereby overcoming the limitations imposed by fixed discretizations.

This transition from discrete to continuous problems necessitates a fundamental shift in how data is represented and processed. While the objective is to learn mappings between functions, in practice the true continuous functions are not available in numerical form. Instead, only discrete samples or discretized representations of these functions can be accessed. Consequently, the learning task must bridge the gap between finite-dimensional data and infinite-dimensional function spaces. 

What distinguishes neural operators from traditional neural networks is their ability to generalize across different discretizations of the same underlying function. A single neural operator is designed to accept inputs sampled arbitrarily and to produce outputs that can be queried continuously at any spatial or temporal location. In contrast, conventional neural networks operate on fixed input and output grids, making them dependent on the chosen resolution and unsuitable for tasks requiring resolution invariance or cross-discretization consistency. This property gives neural operators a distinct advantage in scientific computing, where numerical data often come from heterogeneous meshes or dynamically adaptive discretizations.

\begin{figure}[h]  
\centering   
\includegraphics[width=\textwidth]{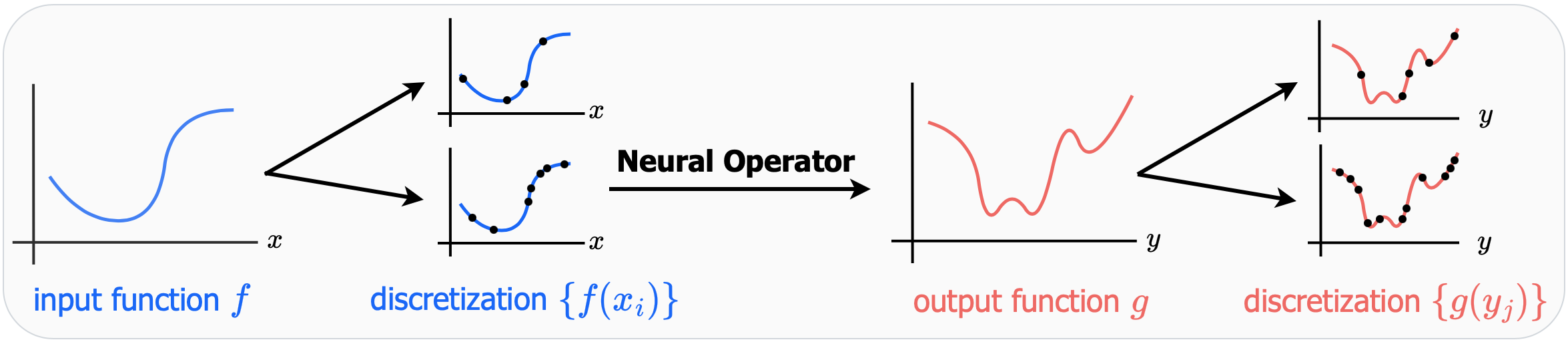}
\vspace{-6.5mm}
\caption{Illustration of the neural operator paradigm. The input is a function $f$. The same neural operator can take as input any discretization $\{f(x_i)\}$ of this function. The neural operator produces an output function $g$, which can then be evaluated at any discretization $\{g(y_j)\}$.}
\label{fig: NNs to NOs}  \vspace{1mm}
\end{figure}

\paragraph{Extending Neural Networks to Neural Operators.} An effective strategy for constructing neural operators is to design simple, modular layers that can be flexibly composed to form expressive operator models. Each layer is defined as an operator acting on functions, ensuring that all intermediate representations throughout the model remain infinite-dimensional functions rather than finite-dimensional vectors. Importantly, the output of every neural operator layer should be resolution-invariant, allowing the overall model to process information consistently across discretizations.  

A common strategy to design neural operator layers is to extend familiar neural network layers, such as convolutions, attention mechanisms, or linear transformations, to the continuous operator setting. The strategy is to first identify the continuous transformation that a given neural network layer implicitly approximates and to reformulate it as an operator acting in function space. The next step is to devise discretization strategies that approximate this continuous operator while preserving resolution invariance, consistency across sampling schemes, and other desired properties. This perspective not only provides a principled foundation for extending deep learning architectures to infinite-dimensional settings but also guides the development of new operator layers that balance expressiveness, efficiency, and theoretical rigor. 

More details on principled methodologies for extending neural networks into neural operators, as well as many examples of common neural network layers adapted to the neural operator setting, can be found in~\citep{Berner2025Principles}.

\hfill

\subsection{Advantages of Neural Operators} \label{sec: Advantages NOs}

Neural operators represent a paradigm shift in scientific computing, offering advantages that address fundamental limitations of both traditional neural networks and numerical integrators. Their mathematical foundation in operator theory, combined with practical benefits like resolution invariance, makes them uniquely suited for the challenges of scientific computing. We present here some of the key advantages of neural operators, which work together to enable new capabilities in scientific computing that were previously impossible or computationally prohibitive. As the field continues to develop, these advantages will likely expand further, opening new possibilities for scientific discovery and engineering applications.

\vspace{4mm}

\paragraph{Mathematical Well-Posedness.} A fundamental advantage of neural operators lies in their mathematical foundation in operator theory. In scientific computing, the problems we aim to solve are inherently operator learning tasks, i.e. the ground-truth mappings of interest are operators. Neural operators are designed to approximate these true operators directly rather than learning discretized approximations. This alignment with the underlying mathematical structure ensures that the learned mapping respects the continuous nature of the problem, and perhaps generalizes better to new unseen settings.

\vspace{4mm}

\paragraph{Universal Operator Approximation.} Fourier Neural Operators (and other neural operators) are universal operator approximators, in the sense that any sufficiently smooth operator can be approximated to arbitrary accuracy using a Fourier Neural Operator (see Theorem 5 in \cite{kovachki2021universal}). This further highlights the versatility of neural operators, as a single architectural framework can be applied across a wide range of scientific domains, including fluid dynamics, materials science, and other areas governed by complex physical processes. Note, however, that just as with universal function approximation theorems for neural networks, these results are only theoretical guarantees. In practice, discrepancies may arise due to discretization errors or complex optimization landscapes, which can lead to suboptimal training and limited accuracy.

\vspace{4mm}

\paragraph{Function Representation and Output Discretization Flexibility.} A significant advantage of neural operators is their ability to output continuous functions that can be queried at arbitrary discretizations. As a result, neural operators can be evaluated at any resolution without retraining, allowing for instance zero-shot super-resolution. This can be valuable when high-resolution predictions are needed but training data are available only at lower resolutions. Neural operator predictions also remain consistent across resolutions, as they correspond to evaluations of the same underlying function at different resolutions, thereby eliminating artifacts that typically arise from discrete-to-continuous interpolation. Additionally, the functional form of the output can also be exploited to compute derivatives, integrals, and other quantities with improved numerical stability, avoiding the accuracy limitations that arise from fixed-resolution representations.

\vspace{4mm}

\paragraph{Solving Parametrized PDEs Simultaneously.} Traditional numerical methods are typically designed to solve individual instances of PDEs under fixed parameters, boundary conditions, and initial conditions. Neural operators, in contrast, are capable of learning solution operators that generalize across entire families of PDEs, enabling a single model to represent a continuum of problem configurations. This flexibility allows neural operators to accommodate variations in physical parameters such as viscosity, temperature, material properties, and many others, while also generalizing across diverse boundary conditions and time-dependent forcing. Furthermore, they can adapt to different domain geometries, making them particularly suitable for tasks such as shape optimization and design. Beyond single-physics scenarios, neural operators can capture coupled behaviors in multi-physics systems, providing a unified framework for modeling interactions between distinct physical processes. 

\vspace{4mm}

\paragraph{Resolution Invariance and Convergence.} Neural operators possess the distinctive ability to be evaluated at arbitrary resolutions, offering several important advantages for scientific computing. They exhibit discretization invariance, producing consistent predictions across input discretizations, whether these are provided on regular grids, unstructured meshes, or point clouds. As the input resolution increases, neural operator predictions converge towards a continuous solution function. This property, often referred to as resolution convergence, ensures that the learned operator maintains physical and numerical fidelity across scales. Moreover, neural operators naturally capture multi-scale behaviors, enabling a single model to represent phenomena spanning fine-grained local dynamics and global system patterns.

\vspace{4mm}

\paragraph{Data Efficiency and Training Advantages.} Neural operators offer the capability to learn effectively from datasets that span multiple resolutions, providing substantial advantages in both efficiency and generalization. Their ability to handle mixed-resolution training allows models to integrate data sampled at different spatial or temporal resolutions, maximizing the use of available computational resources and experimental data. Training can be structured using a curriculum learning approach, where the model is first exposed to low-resolution samples to accelerate convergence and then fine-tuned with high-resolution data to refine accuracy. This progressive strategy not only reduces computational demands but can also enhance stability during optimization. Moreover, data augmentation can be achieved by incorporating data from multiple resolutions of the same physical system, enriching the effective size and diversity of the training set. Finally, neural operators support transfer learning across resolutions, allowing models trained at one scale to be adapted efficiently to another with minimal retraining. These properties make neural operators particularly powerful for large-scale scientific problems where data availability, resolution variability, and computational efficiency are critical considerations.

\vspace{4mm}

\paragraph{Inference Speed and Inverse Design.} Once trained, neural operators can generate high-resolution outputs at a fraction of the computational cost of traditional numerical solvers, achieving speedups that are typically between 100$\times$ and 1,000,000$\times$. These capabilities make neural operators an efficient and scalable alternative for high-fidelity simulations in complex physical systems. In addition, most neural operator architectures are differentiable. Fast inference and full differentiability make neural operators a powerful foundation for inverse design tasks. Their ability to generate high-fidelity predictions within milliseconds enables exhaustive exploration of large parameter spaces that would otherwise be computationally prohibitive using traditional solvers. This efficiency allows millions of evaluations to be performed in practical time frames, supporting global searches across design parameters and facilitating comprehensive mapping of complex design landscapes. At the same time, the differentiable nature of neural operators allows gradient-based optimization to be seamlessly integrated into the design process. Gradients of target quantities with respect to input parameters can be computed through backpropagation, enabling direct and efficient navigation toward optimal solutions. This combination of rapid inference and differentiability bridges global exploration and local refinement, providing a scalable framework for data-driven optimization and accelerating the discovery of optimal designs in high-dimensional spaces.

\vspace{4mm}

\paragraph{Practical Implementation Benefits.}  Neural operators also provide several practical advantages that make them well-suited for large-scale scientific and engineering applications. By operating directly in function spaces rather than on dense grids, they achieve remarkable memory efficiency, allowing high-resolution inputs and outputs to be processed without incurring prohibitive storage or computational costs. Their intrinsic function-to-function mapping structure naturally supports parallelization, enabling efficient computation across spatial and temporal dimensions and facilitating deployment on modern hardware. Furthermore, the continuous representation of learned operators imparts a high degree of robustness to noise and discretization artifacts, improving stability and generalization across diverse datasets. Finally, neural operators can sometimes exhibit a higher degree of interpretability, as distinct components of the learned operator can correspond to meaningful physical processes or structures within the system.

\clearpage

\subsection{The \texttt{NeuralOperator} Library} \label{sec: NO Library}

\hfill 

\begin{figure*}[!h] \vspace{-1.2mm}
\centering
\includegraphics[width=0.6\textwidth]{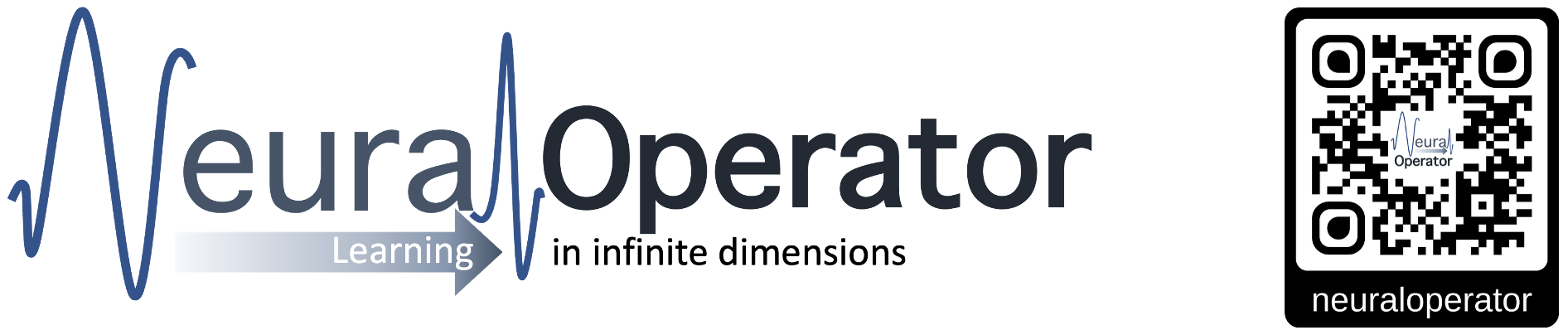}  \vspace{-1.2mm}
\end{figure*}

The \texttt{NeuralOperator} library~\citep{kossaifi2024neural}, open-sourced at \href{https://github.com/neuraloperator/neuraloperator}{github.com/neuraloperator/neuraloperator}, provides the official and most comprehensive implementation of modern neural operator architectures. Part of the official \texttt{PyTorch} Ecosystem\footnote{\url{https://landscape.pytorch.org/}}\! and released under the MIT license, it offers a coherent, extensible, and reproducible environment for constructing, training, and deploying operator learning models. The library translates the mathematical theory of neural operators into a practical computational framework, combining conceptual rigor with accessible software design. Throughout this paper, we use \texttt{NeuralOperator 2.0.0} as a primary reference for exposition, illustrating the implementation of the various components of neural operators.

At its foundation, \texttt{NeuralOperator} extends the scope of deep learning beyond finite-dimensional representations to neural operators that learn mappings between functions. Once trained, a neural operator should be able to act on inputs represented at any resolution or discretization. This principle of discretization invariance is built into the library at every level: from spectral transforms and integral kernels to normalization routines and loss definitions. It enables models trained at one resolution to generalize naturally to others.

The design of \texttt{NeuralOperator} follows three central principles: clarity, modularity, and scientific reliability. All components inherit from the standard \texttt{torch.nn.Module} class, allowing operator learning to integrate seamlessly into existing \texttt{PyTorch} workflows. The architecture is deliberately modular, encouraging the combination of learnable integral transforms, pointwise mappings, and positional encoding into new configurations. This structure supports both rapid experimentation and systematic analysis while preserving theoretical consistency. The same framework accommodates operators defined on Euclidean domains, graphs, manifolds, and spheres, demonstrating the generality of the underlying abstractions.

The library unifies, within a single platform, canonical and state-of-the-art neural operator architectures. The FNO~\cite{li2020fourier} forms its core. The Tensor FNO~\cite{Kossaifi2023MGTFNO} extends this model by introducing low-rank tensor decompositions for parameter efficiency, while the Spherical FNO~\citep{bonev2023spherical} adapts spectral convolutions to data defined on the sphere via spherical harmonic transforms. Further variants such as the Geometry-Informed Neural Operator (GINO)~\citep{li2020neural,li2024geometry,lin2025mGNO} and the Physics-Informed Neural Operator (PINO)~\citep{li2024physics} incorporate geometric priors and physical constraints into the operator formulation. By consolidating these diverse models into a consistent and rigorously tested framework, \texttt{NeuralOperator} serves both as an authoritative reference implementation and as a platform for advancing operator learning research.

A defining strength of the library is its accessibility. Each major architecture is accompanied by example training scripts (in \codebox{scripts}) and ready-to-use configurations, enabling users to reproduce benchmark results or adapt them to new problems with minimal effort. The documentation accompanying the library, available at \href{https://neuraloperator.github.io/dev/index.html}{neuraloperator.github.io/dev/index.html}, is comprehensive and pedagogically structured. It contains a detailed API reference and an extensive user guide. A collection of tutorials illustrating the effect of the different layers and demonstrating how neural operators can be applied to scientific datasets is provided at \href{https://neuraloperator.github.io/dev/auto_examples/index.html}{neuraloperator.github.io/dev/auto\_examples/index.html}. Together, they provide a complete learning path from introductory examples to advanced customization, and have made the library a central resource for both new users and experienced researchers seeking to understand and extend operator-based models.

\texttt{NeuralOperator} is designed not only as a research reference but also as a practical toolkit. It provides state-of-the-art implementations of major neural operator architectures together with numerous auxiliary functions, utilities, and options that simplify experimentation and deployment. Users are encouraged to adopt the library, as it offers a mature and actively maintained foundation for operator learning, capable of serving both as a benchmark baseline and as a flexible starting point for new model development. The project also welcomes community contributions, including new architectures, datasets, or optimization strategies, to continue advancing the field and expanding the library's capabilities through open collaboration.

\clearpage

\subsection{The Fourier Neural Operator (FNO)} \label{sec: The FNO}

\vspace{3mm}

\subsubsection{Overall Architecture} \label{sec: FNO general def}

\vspace{3mm}

\begin{figure*}[h] \vspace{4mm}
\centering
\includegraphics[width=\textwidth]{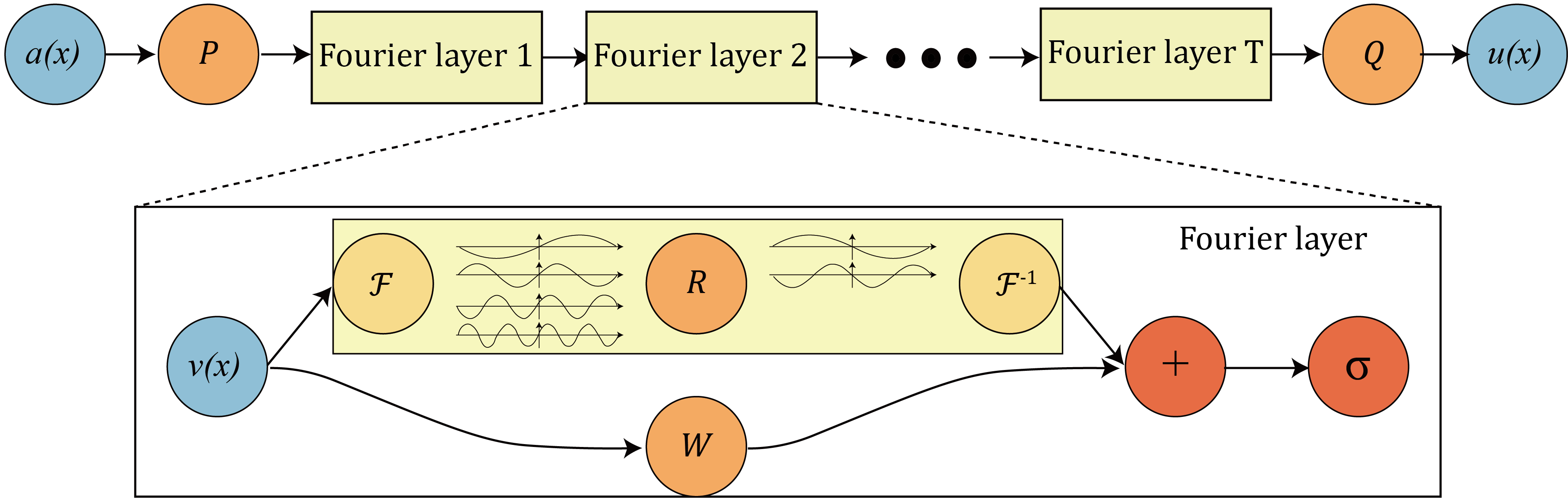} 
\caption{The Fourier Neural Operator (FNO) architecture (extracted from~\citep{li2020fourier}). \label{fig: FNO}  } \vspace{7mm}
\end{figure*}

The original Fourier Neural Operator (FNO)~\citep{li2020fourier}, depicted in Figure~\ref{fig: FNO}, composes Fourier integral operator layers with pointwise nonlinear activation functions $\sigma$ to approximate nonlinear operators as
\begin{equation} \label{eq: FNO}
   \mathcal{Q} \ \circ \  \sigma (\mathcal{W}_{L} + \mathcal{K}_{L}) \ \circ \  \cdots \ \circ \  \sigma(\mathcal{W}_1 + \mathcal{K}_1) \ \circ \  \mathcal{P}.
\end{equation}

In this expression, 
\begin{itemize}
    \item $\mathcal{K_\ell}$ denote spectral convolutions (see \Cref{sec: SpectralConv})
    \item \(\mathcal{P}\) and \(\mathcal{Q} \) are pointwise lifting and projection neural networks which encode the lower-dimensional function into a higher-dimensional space and project it back to the original space, respectively. These are implemented as multilayer perceptrons (MLPs) acting on the channel dimension (see \Cref{sec: ChannelMLP}).
    \item \(\mathcal{W}_\ell \) are linear transforms acting on the channel dimension, implemented as MLPs (see \Cref{sec: ChannelMLP}).
    \item \(\sigma\) denotes a nonlinear activation function.
\end{itemize}

\vspace{2mm}

The trainable parameters of the FNO consist of the parameters in $\mathcal{P}, \mathcal{Q}, \mathcal{W}_\ell, \mathcal{K}_\ell$. On uniform meshes, the Fourier transform $\mathcal{F}$ can be implemented efficiently using the fast Fourier transform.  \\

Overall, the spectral convolutions in the Fourier layers enable nonlocal mixing of information across spatial modes, thereby allowing for global feature coupling. The linear transforms acting on the channel dimension can influence and represent higher-frequency variations in the data, thereby complementing the spectral convolutions by enriching the representation with fine-scale, high-frequency details that may be lost in purely spectral filtering. Finally, lifting and projection layers expand the input into a high-dimensional latent space to capture complex feature interactions and then compress it back to the target dimensionality, enabling expressive intermediate representations while preserving correspondence with the physical input and output fields.

When composed with nonlinear activation functions, these layers collectively provide a highly expressive framework capable of capturing both local and global nonlinear dynamics across multiple scales.

\paragraph{Universal Approximation Theorem.} It has been shown that FNOs possess a universal approximation property: they can approximate any sufficiently regular operator to arbitrary accuracy~\citep{kovachki2021universal}.  To state this result rigorously, one needs a few concepts from functional analysis, which we summarize briefly below.

We consider the periodic domain $\mathbb{T}^d = [0, 2\pi]^d$. A function $a : \mathbb{T}^d \to \mathbb{R}^m$ belongs to the Sobolev space $H^s(\mathbb{T}^d ;\mathbb{R}^m)$ if both $a$ and its derivatives up to order $s$ are square-integrable. The same idea can be generalized to non-integer values of $s$. Intuitively, larger $s$ corresponds to smoother functions, while smaller or negative $s$ allows for rougher ones. For a function $a : \mathbb{T}^d \to \mathbb{R}^m$, the Sobolev norm of order $s \ge 0$ is defined as
\begin{equation}
\| a \|_{H^s}^2 = \sum_{k \in \mathbb{Z}^d} (1 + |k|^2)^s \, \| \widehat{a}(k) \|_2^2,
\end{equation}
where $\widehat{a}(k)$ denotes the Fourier coefficient of $a$ at frequency $k$.  This norm measures both the size of the function and the smoothness of its oscillations. For integer $s$, it is equivalent to combining the $L^2$ norms of all derivatives up to order $s$.

\vspace{3mm}

\begin{theorem}[Universal Approximation Theorem for FNOs {\citep{kovachki2021universal}}]
Let $\mathcal{G} : H^s(\mathbb{T}^d; \mathbb{R}^{d_a}) \to H^{s'}(\mathbb{T}^d; \mathbb{R}^{d_u})$ be a continuous operator, and let $K \subset H^s(\mathbb{T}^d; \mathbb{R}^{d_a})$ be compact. Then for every $\varepsilon > 0$, there exists a Fourier Neural Operator $\mathcal{N}$ such that
\begin{equation}
    \sup_{a \in K} \| \mathcal{G}(a) - \mathcal{N}(a) \|_{H^{s'}} \le \varepsilon.
\end{equation}

Moreover, this result extends to operators on Lipschitz domains $\Omega \subset \mathbb{R}^d$ (i.e. $\partial \Omega$ can be locally described by graphs of Lipschitz continuous functions)  by periodic extension of inputs and restriction of outputs.
\end{theorem}

\vspace{2mm}

This theorem generalizes the classical universal approximation result for neural networks to mappings between function spaces.  FNOs act on functions by alternating global spectral convolutional layers with pointwise nonlinearities.  These layers can represent integral transforms of arbitrary complexity, allowing FNOs to approximate any continuous operator between Sobolev spaces. As a result, the theorem establishes that FNOs can represent any such operator to arbitrary precision, which provides the theoretical foundation for using FNOs in data-driven learning of solution operators for PDEs and other infinite-dimensional mappings.

The theorem guarantees expressivity but not necessarily efficiency, and in the worst case, the model size may grow exponentially with the inverse of the desired accuracy for rough operators.  
However, for operators arising from PDEs with smooth structure, subsequent results in \citep{kovachki2021universal} show that FNOs achieve polynomial or logarithmic complexity in the approximation error.  Thus, the universal approximation theorem for FNOs provides a rigorous justification 
for their practical success in learning complex physical systems.

This result can be extended to the $C^1$ setting. Under suitable regularity assumptions, FNOs with smooth activation functions can simultaneously approximate a continuously differentiable operator and its Fr\'echet derivative uniformly on compact subsets of the input function space \cite{yao2026derivativesFNO}. This shows that FNOs can in principle approximate not only operator outputs but also their sensitivities with respect to input functions.

\hfill 

\paragraph{Resolution-invariance.} To enjoy resolution invariance, the FNO represents functions in the frequency domain using a truncated Fourier basis expansion. In this representation, each function is expressed as a finite sum of Fourier modes, and the model learns a transformation on the coefficients of these modes, rather than on the discrete spatial samples themselves. As a result, the learned mapping is defined over the continuous Fourier basis which is independent of the underlying discretization. 

Increasing the resolution in the physical domain simply provides a more accurate sampling of the same truncated Fourier expansion, without changing the learned parameters. 

The output can be reconstructed at any resolution from its frequency coefficients through the inverse FFT, or queried at arbitrary locations by replacing the inverse FFT with a (slower) inverse DFT. Thus, the FNO can be trained on coarse grids and directly applied to finer ones, achieving super-resolution, while maintaining consistent predictions across discretizations as it is just evaluating the same function at higher resolution.

\hfill

\subsubsection{Spectral Convolution: \codebox{SpectralConv}}  \label{sec: SpectralConv}

\vspace{2mm}

The spectral convolution layer is the central mechanism of the FNO, transforming a global integral operator into a simple and learnable multiplication in the frequency domain. This formulation provides an elegant balance between theoretical rigor and computational efficiency, allowing the operator to act globally while remaining compact and data-efficient. By operating directly on the spectral representation, the layer captures long-range correlations and complex nonlinear dependencies across spatial scales (and temporal scales as well when time is treated as an extra spatial dimension). Its parameters act on the underlying frequency content rather than on discrete spatial grids, which makes the operation inherently resolution-invariant: once trained, the same model can generalize seamlessly to inputs of different resolutions without retraining. This combination of global connectivity, interpretability, and scalability makes the spectral convolution a fundamental and powerful component of the FNO architecture.

\paragraph{Construction.} The starting point is the kernel integral operator $\mathcal{K}$, which acts on a latent representation $v$ of the input function by aggregating non-local information across the spatial domain:
\begin{equation}
    (\mathcal{K} v)(x) = \int_D \kappa_\phi(x, y)\, v(y)\, dy ,
\end{equation}
where $\kappa_\phi$ is a learnable kernel. In particular, when the kernel depends only on the difference of its arguments, $\kappa_\phi(x,y) = \kappa_\phi(x - y)$, the operator becomes a convolution. 
\begin{equation}
    (\mathcal{K} v)(x) \ = \  \int_D \kappa_\phi(x - y)\, v(y)\, dy \ = \  \kappa_\phi \ * \ v \ .
\end{equation}
Recall from \Cref{sec: Continuous FT} the spectral convolution theorem $f * g = \mathcal{F}^{-1}\left(\mathcal{F}(f) \cdot \mathcal{F}(g)\right) $ which converts convolutions in the spatial domain into pointwise products in the frequency domain. As a result, we can rewrite the kernel integral operator as
\begin{equation}
    (\mathcal{K} v)(x) = \mathcal{F}^{-1}\big( \mathcal{F}(\kappa_\phi) \cdot \mathcal{F}(v) \big)(x)
\end{equation}
This identity provides the foundation for the spectral convolution: rather than evaluating the kernel integral operator directly in physical space, the operation is performed as a simple product in the frequency domain. Instead of learning the kernel $\kappa_\phi$ itself, the FNO learns its Fourier coefficients, and the spectral convolution operation is therefore written as
\begin{equation}
    (\mathcal{K}_\phi   v)(x) = \mathcal{F}^{-1}\big( R_\phi \cdot \mathcal{F}(v) \big)(x),
\end{equation}
where $R_\phi$ is a complex-valued learnable tensor that acts linearly on the spectral representation of $v$. 

Once the operator is expressed in the spectral domain, the Fourier representation of the signal naturally decomposes it into independent frequency modes. The FNO does not treat these modes as isolated but leverages their joint structure by allowing controlled interactions between the feature channels associated with each frequency. This is achieved through a learnable linear transformation in Fourier space. For every mode $k$, the model learns a complex-valued matrix $R_\phi(k) \in \mathbb{C}^{d_v \times d_v}$, where $d_v$ is the number of feature channels. This matrix governs how information is mixed across channels within that particular frequency, enabling the model to capture how different physical quantities or latent variables interact at a given spatial scale. Conceptually, $R_\phi(k)$ serves as a spectral mixing operator: it determines how much energy to emphasize or suppress in each mode and how to redistribute correlations among channels. In this way, the spectral convolution does not merely scale frequency components, but actively learns how to reshape the flow of information across scales and variables in the frequency domain.

The collection of matrices $\{ R_\phi(k) \}$ forms a tensor $R_\phi$ that defines how the operator acts in the frequency domain. However, the FNO does not learn weights for all possible modes. Instead, it truncates the representation to a subset of the lowest-frequency components, $|k_m| \leq K_{m}$ along each spatial dimension $m$, for some integer frequency cutoff $K_{m}$. This truncation encodes a physically meaningful inductive bias. In many systems governed by PDEs, most of the signal's structure and long-range correlations are concentrated in low frequencies, while higher frequencies mainly contain fine-scale noise or local fluctuations. By focusing the learning process on these dominant components, the spectral convolution achieves a compact and data-efficient parameterization. The neglected higher frequencies are not lost entirely but can be reconstructed indirectly through nonlinearities, residual connections, and channel-mixing layers of the model.

\paragraph{In the discrete setting.} When the spatial domain $D$ is discretized into $n$ grid points, the latent function $v$ becomes an array $v \in \mathbb{R}^{n \times d_v}$. The Fourier transform $\mathcal{F}(v)$ is computed efficiently via the FFT. The core operation of the spectral convolution is then a weighted contraction between the transformed features and the learned coefficients:
\begin{equation}
    [(R_\phi \cdot \mathcal{F}(v))(k)]_{l} \ = \  \sum_{j=1}^{d_v} \ [R_{\phi}(k)]_{ l,j} \ [\mathcal{F}(v)(k)]_{j},
\end{equation}
where $k$ denotes the specific frequency component, and $l$ the channel index. Here, the summation is performed over feature channels (index $j$), and the operation is restricted to the retained modes $|k_m| \leq K_{m}$ along each spatial dimension $m$. 

Through this operation, at each frequency, the complex matrix $R_\phi(k)$ mixes the information across feature channels, effectively coupling physical variables. The operation is linear in the frequency domain but non-local in physical space, meaning that every spatial location is influenced by all others through these shared modes. After this transformation, the modified spectral coefficients are mapped back to the spatial domain using the inverse FFT:
\begin{equation}
    (K_\phi v)(x) = \mathcal{F}^{-1}\!\big( R_\phi \cdot \mathcal{F}(v) \big)(x).
\end{equation}

This inverse transform reassembles the globally mixed frequency components into a spatial representation. The result is a dense and global transformation that propagates information efficiently across the entire domain. Every point in the field is indirectly coupled to every other through the Fourier basis, achieving global receptive fields without explicitly computing pairwise interactions. \\

\vspace{6mm}

\paragraph{Implementation.} The spectral convolution is implemented in \texttt{NeuralOperator 2.0.0} as the \codebox{SpectralConv} class in \codebox{neuralop/layers/spectral\_convolutions.py}. 

\vspace{2mm}

We provide here a simplified implementation of the \codebox{SpectralConv} forward in two dimensions:

\vspace{2.5mm}

\begin{minted}[fontsize=\footnotesize, bgcolor=gray!5, frame=single, linenos]{python}
def SpectralConv_forward(x, weight):

    # 1. Compute the Fourier transform of the input along spatial dimensions
    x_fft = torch.fft.fftn(x, dim=(-2, -1))

    # 2. Shift the order of the frequency components so the zero mode is in the center of the spectrum
    x_fft = torch.fft.fftshift(x_fft, dim=(-2, -1))

    # 3. Identify and select the central low-frequency region where learnable weights act
    center_x, center_y = Nx // 2, Ny // 2
    start_x, end_x = center_x - n_modes[0] // 2, center_x + (n_modes[0] + 1) // 2
    start_y, end_y = center_y - n_modes[1] // 2, center_y + (n_modes[1] + 1) // 2
    x_selected = x_fft[:, :, start_x:end_x, start_y:end_y]

    # 4. Initialize the output tensor in the frequency domain
    out_fft = torch.zeros((batch_size, weight.shape[1], Nx, Ny), dtype=torch.cfloat)

    # 5. Apply the learned complex-valued weights to the selected modes
    out_fft[:, :, start_x:end_x, start_y:end_y] = torch.einsum('bixy,ioxy->boxy', x_selected, weight)

    # 6. Shift frequencies back so the zero-frequency component returns to the corner
    out_fft = torch.fft.ifftshift(out_fft, dim=(-2, -1))

    # 7. Apply the inverse Fourier transform to reconstruct the output in the spatial domain
    return torch.fft.ifftn(out_fft, dim=(-2, -1)).real
\end{minted}

\clearpage 

In practical implementations, the spectral convolution layer is significantly more advanced than the simplified pseudocode presented above. Several design optimizations improve computational efficiency, numerical stability, and overall memory usage. When the input field is real-valued, the implementation typically employs the real-valued Fourier transform functions \texttt{torch.fft.rfftn} and \texttt{torch.fft.irfftn} along the last spatial dimension, instead of \texttt{torch.fft.fftn} and \texttt{torch.fft.ifftn}. This takes advantage of the Hermitian symmetry of the Fourier spectrum, which makes the negative frequencies redundant (since they contain exactly the same information as the positive frequencies). As a result, only the non-redundant half of the frequency domain is stored and processed, effectively reducing both computation and memory requirements by a factor of two. The contraction between the transformed features and the learnable weights is also optimized. Instead of performing a full dense tensor multiplication, efficient \texttt{einsum} expressions can be used, and the weights can be stored using factorized representations that reduce the number of parameters and computational cost while preserving the ability to model complex spectral interactions (see \Cref{sec: tfno}). Additional implementation details, such as separable contractions, mixed-precision arithmetic, and resolution scaling, make the spectral convolution both stable and scalable across diverse problem settings while maintaining high accuracy and efficiency.

The main parameter of the \codebox{FNO} class that governs the behavior of the \codebox{SpectralConv} layer is \codebox{n\_modes}, which defines the number of retained Fourier modes along each spatial dimension. It controls the spectral resolution of the operator by determining how many low-frequency components are learned and used during convolution. To preserve the validity of the frequency representation, \codebox{n\_modes} should not exceed half the spatial resolution, in accordance with the Nyquist--Shannon sampling theorem (see \Cref{sec: Nyquist}). More details about tuning \codebox{n\_modes} are provided in \Cref{sec: n_modes}.

\vspace{4mm}

One additional parameter is \codebox{separable} (by default \texttt{False}), which determines whether the \codebox{SpectralConv} operator mixes channels in Fourier space: \vspace{1mm}
\begin{itemize}
    \item When it is set to \texttt{False}, the layer leverages a complex weight tensor of shape $C_{\text{hidden}} \times C_{\text{hidden}}$ for each mode. This weight tensor acts as a frequency-specific linear map across channels, allowing the spectral convolution to mix channels in one way for a specific frequency and another way for a different frequency.
    \item When \codebox{separable} is \texttt{True}, rather than learning a full $C_{\text{hidden}} \times C_{\text{hidden}}$ channel-mixing matrix at each Fourier mode, the layer learns a single complex value for each channel and retained frequency. Concretely, each channel is Fourier transformed, every retained Fourier coefficient in that channel is scaled by its own learned complex number, and the result is inverse transformed back to physical space. This substantially reduces the number of learnable parameters and the computational cost, but it also removes the ability of the spectral convolution to exchange information across channels, so any necessary channel mixing must be handled elsewhere in the architecture.
\end{itemize}

\vspace{3mm}

To improve efficiency, one can also leverage the \codebox{factorization} and \codebox{rank} arguments of the \codebox{FNO} class to represent the weight tensor using low-rank factorizations, as discussed in more detail in \Cref{sec: tfno}.

\hfill

\paragraph{Parameter Count.} For each mode $k$, there is a learnable dense complex-valued matrix $R_\phi(k)$ of shape $C_{\text{hidden}} \times C_{\text{hidden}}$ in the spectral convolution layer. Collectively, these matrices form a tensor $R_\phi$ with $C_{\text{hidden}}^2 \prod_{m=1}^d K_m$ entries, where $K_m$ denotes the number of retained modes along dimension $m$ (specified by \codebox{n\_modes}). Each complex-valued coefficient is parameterized by two real-valued parameters. Adding a bias term with $C_{\text{hidden}}$ parameters, the total number of trainable parameters in a \codebox{SpectralConv} layer is
\begin{equation}
\eta C_{\text{hidden}}^2 \prod_{m=1}^d K_m + C_{\text{hidden}},
\end{equation}
where $\eta=1$ for real-valued data and $\eta=2$ for complex-valued data. 

\codebox{SpectralConv} contains the majority of the learnable parameters in the FNO, since we will see in the subsequent subsections that the other components individually contain less than $4  C_{\text{hidden}}^2$ trainable parameters.

\clearpage

\paragraph{Inference Computational Cost.} A forward pass through \codebox{SpectralConv} consists of (i) a forward FFT, (ii) a learned channel-mixing contraction on a truncated low-frequency block, and (iii) an inverse FFT. Let the input have shape $(B, C_{\text{hidden}}, N_1,\dots,N_d)$ and define the total number of spatiotemporal points $N \ = \ \prod_{m=1}^d N_m $. The FFT and inverse FFT each scale quasi-linearly in the grid size, i.e.
\begin{equation}
\mathcal{O} \big(B\, C_{\text{hidden}}\, N \log N\big),
\end{equation}
up to constant factors and the real-FFT redundancy reduction. The learnable part acts only on the retained Fourier modes. If $K_m$ modes are kept along dimension $j$ (with the last dimension understood after the real-FFT adjustment), then for each retained mode $k$ the layer applies a dense complex matrix $R_\phi(k)$ of shape $C_{\text{hidden}} \times C_{\text{hidden}}$ to mix channels, yielding contraction cost
\begin{equation}
\mathcal{O} \big(B \ C_{\text{hidden}}^2 \,\prod_{m=1}^d K_m\big).
\end{equation}
Overall, the forward-pass complexity is
\begin{equation}
\mathcal{O} \Big(B\,C_{\text{hidden}}\, N \log N \;+\; B \ C_{\text{hidden}}^2 \ \prod_{m=1}^d K_m \Big).
\end{equation}

\hfill \\ 

\subsubsection{Channel Mixing: \codebox{ChannelMLP}}  \label{sec: ChannelMLP}

\vspace{3mm}

The \codebox{ChannelMLP} applies a multilayer perceptron (MLP) along the channel axis while sharing parameters across all spatiotemporal positions. The \codebox{ChannelMLP} module is agnostic to spatiotemporal resolution and supports inputs with one or more spatiotemporal axes. It implements the linear transforms \(\mathcal{W}_l \) acting on the channel dimension in \texttt{NeuralOperator 2.0.0} in \codebox{neuralop/layers/channel\_mlp.py} using one-dimensional convolutions of kernel size one which act as position-wise linear maps over channels and therefore do not couple spatiotemporal locations.

In more detail, the \codebox{ChannelMLP} can be understood as a small channel-wise MLP applied identically to every spatiotemporal position. Each spatiotemporal point is treated as a separate sample of a small feed-forward network that mixes information only across channels. In practice, this is implemented efficiently using one-dimensional convolutions with kernel size equal to one, which correspond to shared linear projections across all spatiotemporal sites. The kernel size of one ensures low computational overhead while maintaining expressive channel mixing. The first layer maps $C_{\text{in}}$ input channels to $C_{\text{hidden}}$ hidden channels, then the intermediate layers preserve the number of hidden channels, and the final layer maps the $C_{\text{hidden}}$ hidden channels to $C_{\text{out}}$ output channels. Nonlinear activations are inserted between all internal layers, followed optionally by dropout regularization. Conceptually, the \codebox{ChannelMLP} acts as a position-wise MLP that enriches the channel representation without introducing spatiotemporal coupling, providing an efficient and scalable mechanism for feature transformation in high-dimensional data. \\

\paragraph{Importance within the FNO.} While the \codebox{SpectralConv} layer operates on low-frequency components by truncating or weighting a limited set of Fourier modes, the \codebox{ChannelMLP} acts directly in the channel domain and applies identical transformations at every spatiotemporal location. As a result, it is not constrained to the lower part of the frequency spectrum. Instead, it can influence and represent both low-frequency and high-frequency variations in the data. This property allows \codebox{ChannelMLP} to complement spectral layers by enriching the representation with fine-scale, high-frequency details that may be lost in purely spectral filtering. Consequently, combining spectral convolutions with channel-wise MLPs enables a balanced treatment of global smooth structures and local high-frequency corrections within the FNO. \\

\paragraph{Implementation.} The module operates on tensors with shape $(B, C_{\text{in}}, N_1, \ldots, N_d)$, where $B$ is the batch size, $C_{\text{in}}$ the number of input channels, and $N_1, \ldots, N_d$ the spatiotemporal dimensions. Each spatiotemporal location can be viewed as a vector of channel values that is processed independently from other locations. The output tensor retains the same spatiotemporal dimensions while replacing the channel dimension with $C_{\text{out}}$, producing a tensor of shape $(B, C_{\text{out}}, N_1, \ldots, N_d)$. Note that the transformation depends only on the channel content and not on the spatiotemporal resolution.

During the forward pass, the input tensor is temporarily reshaped so that all spatiotemporal positions are flattened into a single dimension. This allows the $1 \times 1$ convolutions to operate as pointwise channel transformations. The model then applies each convolutional layer sequentially, followed by the chosen nonlinearity on all but the last layer. If dropout is activated, it is applied after every layer. After processing, the tensor is reshaped back to its original spatiotemporal form.  \\

The main parameters of the \codebox{FNO} class which can be used to modify the channel-mixing layers are
\begin{itemize}
\item \codebox{use\_channel\_mlp} specifies whether to use the MLPs \(\mathcal{W}_l \) on the channel axis. \texttt{True} by default.
  \item \codebox{hidden\_channels} in the \codebox{FNO} class specifies the number of input and output channels $C_{\text{in}} = C_{\text{out}}$ of the channel-mixing layers, and \codebox{channel\_mlp\_expansion} sets $C_{\text{hidden}} = \text{channel\_mlp\_expansion} \times C_{\text{in}}$ for internal layers. By default, \codebox{channel\_mlp\_expansion} is set to 0.5.
    \item \codebox{channel\_mlp\_dropout} is a probability in $[0,1]$ specifying the dropout probability. By default 0.
  \item \codebox{non\_linearity} sets the nonlinearity function used between internal layers. 
\end{itemize}

\hfill

A simplified pseudocode for the \codebox{ChannelMLP} forward is provided below 

\vspace{2mm}

\begin{minted}[fontsize=\footnotesize, bgcolor=gray!5, frame=single, linenos]{python}
## Sequence of channel-mixing layers
    self.mix_layers = nn.ModuleList()
    
    # First layer: input --> hidden
    self.mix_layers.append(nn.Conv1d(self.in_channels, self.hidden_channels, kernel_size=1))
    
    # Internal hidden layers (only if more than 2 total layers)
    for _ in range(n_layers - 2):
        self.mix_layers.append(nn.Conv1d(self.hidden_channels, self.hidden_channels, kernel_size=1))
        
    # Final layer: hidden --> output
    self.mix_layers.append(nn.Conv1d(self.hidden_channels, self.out_channels, kernel_size=1))

            
## Forward pass of ChannelMLP acting on tensor x
    # Flatten spatiotemporal dimensions if input has more than 3 dimensions
    # (batch, channels, s1, s2, ...) --> (batch, channels, -1)
    size = list(x.shape)
    if x.ndim > 3:
        x = x.reshape((*size[:2], -1))
        reshaped = True
    
    # Apply each layer sequentially
    for i, channel_mixing_layer in enumerate(self.mix_layers):
        x = channel_mixing_layer(x)  
        if i < self.n_layers - 1:
            x = self.non_linearity(x)
        if self.dropout is not None:
            x = self.dropout[i](x)
    
    # Restore the original spatiotemporal shape if reshaped earlier
    if reshaped:
        x = x.reshape((size[0], self.out_channels, *size[2:]))
\end{minted}

\clearpage

\paragraph{Parameter Count.} We begin by analyzing the number of trainable parameters in a general \codebox{ChannelMLP} layer. Let $C_{\text{in}}$, $C_{\text{out}}$, and $C_h$ represent the input, output, and hidden channel sizes, respectively, and let $L$ denote the number of layers. The total number of trainable parameters can be expressed as
\begin{equation}
\begin{cases}
C_{\text{out}}(C_{\text{in}} + 1), & L = 1,\\[6pt]
C_h(C_{\text{in}} + 1) + C_{\text{out}}(C_h + 1), & L = 2,\\[6pt]
C_h(C_{\text{in}} + 1) + (L - 2)C_h(C_h + 1) + C_{\text{out}}(C_h + 1), & L \ge 3~.
\end{cases}
\end{equation}
As discussed earlier, the channel-mixing layer is realized as a two-layer \codebox{ChannelMLP} where the numbers of channels are $C_{\text{in}} = C_{\text{out}} = C_{\text{hidden}}$ and $C_h = 0.5 C_{\text{hidden}}$ by default. Thus, each channel-mixing layer contains approximately $C_{\text{hidden}}^2 + 1.5C_{\text{hidden}}$ trainable parameters, typically between 1,000 and 100,000.

\hfill 

\paragraph{Inference Computational Cost.} The \codebox{ChannelMLP} in \texttt{NeuralOperator 2.0.0} is a position-wise MLP implemented via $1\times 1$ convolutions, so it applies the same channel transformation independently at every spatiotemporal location. Consider an input of shape $(B, C_{\text{in}}, N_1,\ldots,N_d)$, and let $N \ = \ \prod_{m=1}^d N_m$ be the number of spatiotemporal points. For an $L$-layer \codebox{ChannelMLP} with widths $C_0=C_{\text{in}}, C_1,\ldots,C_L=C_{\text{out}}$, layer $\ell$ performs a matrix multiplication of size $C_{\ell}\times C_{\ell+1}$ at each of the $BN$ positions, giving \begin{equation} \mathcal{O}\big(B\,N\,C_{\ell}\,C_{\ell+1}\big)\end{equation} cost per layer, and total inference cost
\begin{equation}
\mathcal{O}\Big(B\,N \ \sum_{\ell=0}^{L-1} C_{\ell}\,C_{\ell+1}\Big).
\end{equation}
The channel-mixing layer in the FNO is realized as a two-layer \codebox{ChannelMLP} where the numbers of channels are $C_{\text{in}} = C_{\text{out}} = C_{\text{hidden}}$ and $C_h = 0.5 C_{\text{hidden}}$ by default. Thus, its inference cost scales as $\mathcal{O}\big(B \ N \ C_{\text{hidden}}^2\big)$.

\hfill \\ 

\subsubsection{Skip Connections} \label{sec: skip connections}

\vspace{3mm} 

\mydef{Skip connections} (also known as \mydef{residual connections}) have become a foundational element of modern deep learning architectures~\citep{He2016}, functioning both as optimization stabilizers and as mechanisms for preserving representational fidelity across model depth. By providing a direct route from the input to the output for the flow of activations and gradients, they mitigate the vanishing-gradient and exploding-gradient problems that commonly hinder the training of models as they get deeper. This residual formulation enables each layer to focus on learning incremental refinements to its input rather than reconstructing the entire mapping from scratch. As a result, architectures with skip connections exhibit faster convergence, improved numerical stability, and enhanced generalization, while maintaining the ability to propagate fine-scale information throughout the model. 

\vspace{1.5mm}

In the context of FNOs, spectral convolutions operate in the Fourier domain by learning filters on a truncated set of low-frequency modes. This design efficiently captures large-scale and smooth structures but inherently restricts the representation of high-frequency content. As a consequence, fine-scale spatiotemporal information, such as sharp gradients or localized features, may be attenuated or lost through spectral truncation. Skip connections address this limitation by providing complementary pathways that preserve and propagate unfiltered high-frequency components throughout the model. In particular, the channel-wise skip connection mechanisms act directly in the spatiotemporal domain, transmitting detailed local variations that spectral convolutions overlook. The resulting interplay between frequency-limited spectral layers and full-band skip paths enables the architecture to simultaneously model smooth global dynamics and intricate high-frequency behaviors, yielding a more balanced and expressive representation across scales.

\vspace{1.5mm}

Beyond preserving information content, skip connections also improve the numerical and optimization properties of deep operator models. By providing shorter gradient routes between early and late layers, they facilitate stable training even in very deep architectures. The residual formulation ensures that each block learns small, incremental refinements to the input representation rather than full mappings, which reduces the risk of overfitting and accelerates convergence. When combined with soft-gating, the model can adaptively control the magnitude of these refinements, further enhancing stability during optimization. \\

\paragraph{Implementation.} Each FNO layer incorporates two complementary types of skip connections: one associated with the spectral pathway called \codebox{fno\_skips}, and a second one associated with the \codebox{ChannelMLP}.

Skip connections are implemented in \texttt{NeuralOperator 2.0.0} in \codebox{neuralop/layers/skip\_connections.py} as (potentially) learnable residual modules $\mathcal{S}_k$, each combined additively with its associated transformation $\mathcal{A}_k$, 
\begin{equation}
    x_{k+1} =  \mathcal{A}_k(x_k) + \mathcal{S}_k(x_k).
\end{equation}

Three variants of $\mathcal{S}_k$ are provided, corresponding to increasing levels of flexibility:
\vspace{2mm}
\begin{itemize}
    \item \textbf{Identity:} $ \mathcal{S}_{\text{id}}(x) = x$. This simple residual pathway preserves the original signal exactly, ensuring perfect gradient flow and enabling the layer to learn corrections relative to the identity mapping. 

    \vspace{2mm}
    
    \item \textbf{Linear:} $   \mathcal{S}_{\text{linear}}(x) = W * x$,   where $W \in \mathbb{R}^{C_{\text{out}} \times C_{\text{in}} \times 1}$ represents a learnable $1 \times 1$ convolution that allows adaptive channel mixing without altering spatiotemporal resolution.
      \vspace{2mm}
      
    \item \textbf{Soft-gating:} $   \mathcal{S}_{\text{gate}}(x) = w \odot x + b$, where $w$ and $b$ are learnable parameters applied channel-wise, and $\odot$ denotes element-wise multiplication. The gating mechanism dynamically modulates the strength of the residual signal per channel, allowing the model to control the relative contribution of the skip connection and transformed pathways during training.
\end{itemize}

\vspace{3mm}

The type of skip connection used in both the channel-mixing MLP and the spectral FNO layers can be specified as arguments of the \codebox{FNO} class through the parameters \codebox{channel\_mlp\_skip} and \codebox{fno\_skip}, which accept the options \texttt{"linear"}, \texttt{"identity"}, \texttt{"soft-gating"}, or \texttt{None}.

\vspace{3mm}

\paragraph{Parameter Counts.}  The \texttt{``linear''} skip connections are implemented using \texttt{torch.nn.Conv1d} modules with a kernel size of 1, resulting in $C_{\text{hidden}}^2$ learnable parameters. In contrast, the \texttt{``soft-gating''} skip connections are far more lightweight, containing only $C_{\text{hidden}}$ parameters.

\hfill \\

\subsubsection{Fourier Layers: \codebox{FNOBlock}} \label{sec: FNOBlock}

\vspace{3mm}

Each \codebox{FNOBlock} constitutes one computational layer of the FNO, combining spectral convolution, channel-wise nonlinear mixing, and residual skip connections into a unified forward transformation. Conceptually, each \codebox{FNOBlock} performs a composite nonlinear update consisting of a spectral transformation followed by a channel-wise mixing step. 

The \codebox{FNOBlock} has been slightly modified since the original FNO paper, with the addition of skip-connections and normalization layers. It is now implemented in \texttt{NeuralOperator 2.0.0} as
\begin{equation}
\begin{aligned}
x_{k+\frac{1}{2}} &= \! \sigma_k \left(\mathcal{N}_k\left[\mathcal{F}_k(x_k)\right]  + \mathcal{S}^{\text{FNO}}_k(x_k)\right), \\[4pt]
x_{k+1} &= \sigma_k\left( \mathcal{N}_k \left[   \mathcal{W}_k(x_{k+\frac{1}{2}}) + \mathcal{S}^{\text{MLP}}_k(x_k)       \right]  \right) ,
\end{aligned}
\end{equation}
where $\mathcal{F}_k$ denotes the spectral convolution operating in the Fourier domain, $\mathcal{W}_k$ the channel-wise MLP acting in the physical domain, and $\mathcal{S}^{\text{FNO}}_k$ and $\mathcal{S}^{\text{MLP}}_k$ the corresponding skip connections for each pathway. The operator $\mathcal{N}_k$  represents a possible normalization, while $\sigma_k$ denotes the nonlinearity.

This two-stage formulation highlights the complementary roles of the spectral and channel components: the spectral convolution $\mathcal{F}_k$ captures long-range, low-frequency correlations in the Fourier domain, while the channel-mixing MLP $\mathcal{W}_k$ restores and enriches high-frequency details directly in the spatial domain. The inclusion of skip connections $\mathcal{S}^{\text{FNO}}_k$ and $\mathcal{S}^{\text{MLP}}_k$ ensures efficient gradient flow and information preservation across layers, enabling stable optimization and improved expressivity of the learned operator. \\

\begin{figure}[h]  
\centering   \
\includegraphics[width=\textwidth]{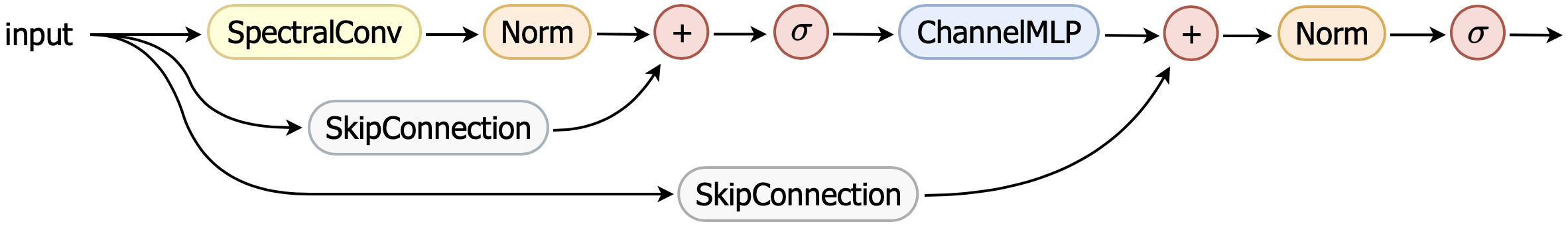}  \vspace{-3mm}
\caption{The \codebox{FNOBlock} as implemented in \texttt{NeuralOperator 2.0.0}. } 
\label{fig: FNOBlock} \vspace{8mm}
\end{figure}

The \codebox{FNOBlock} is displayed in \Cref{fig: FNOBlock}. 
Alternative combinations of components are available through the \codebox{preactivation} argument of the \codebox{FNO} class, following investigations carried out in~\cite{Kossaifi2023MGTFNO}.

\vspace{3mm}

We present below a simplified implementation of the \codebox{FNOBlock} forward from  \texttt{NeuralOperator 2.0.0}, which can be found in \codebox{neuralop/layers/fno\_block.py}:

\vspace{1mm}

\begin{minted}[fontsize=\small, bgcolor=gray!5, frame=single, linenos]{python}
## Forward pass of the i-th FNOBlock layer

    # Compute skip connections if defined
    if self.fno_skips is not None:
        x_skip_fno = self.fno_skips[i](x)
    if self.use_channel_mlp and self.channel_mlp_skips is not None:
        x_skip_mlp = self.channel_mlp_skips[i](x)
        
    x_fno = self.convs[i](x)  # Spectral convolution

    # Apply normalization after spectral convolution
    if self.norm is not None:   
        x_fno = self.norm[self.n_norms * i](x_fno)

    # Add spectral skip connection
    x = x_fno + x_skip_fno if self.fno_skips is not None else x_fno   
    
    x = self.non_linearity(x)   # Apply nonlinearity

    # Channel-wise MLP transformation
    if self.use_channel_mlp:
        x_mlp = self.channel_mlp[i](x)
        x = x_mlp + x_skip_mlp if self.channel_mlp_skips is not None else x_mlp
    
    # Apply normalization after channel mixing
    if self.norm is not None:   
        x = self.norm[self.n_norms * i + 1](x)

    return self.non_linearity(x)  # Apply nonlinearity
\end{minted}

\clearpage

\subsubsection{Lifting and Projection Layers} \label{sec: Lifting and Projection}

\vspace{3mm}

The FNO architecture employs two additional \codebox{ChannelMLP} modules to perform the lifting and projection operations at the beginning and end of the model, respectively. 

The lifting layer maps the input field from its original channel dimension $C_{\text{in}}$ into a higher-dimensional latent representation $C_{\text{hidden}}$, thereby expanding the representational capacity of the model before applying the sequence of Fourier layers. The projection layer performs the inverse operation, mapping the final latent representation back to the desired output dimensionality \(C_{\text{out}} \). 

This expansion enriches the internal feature space by allowing the model to represent a broader set of channel interactions before applying the sequence of Fourier layers. Conceptually, this resembles how convolutional neural networks (CNNs) transform an RGB image with 3 color channels into a hierarchy of feature maps with progressively higher channel dimensionality such as 64, 128, or 256 channels in intermediate layers. This enables the model to capture richer spatial and semantic relationships, which are later compressed back into the original 3-channel RGB space for output. Similarly, the lifting layer in the \codebox{FNO} expands the input into a high-dimensional latent representation to capture complex correlations, while the projection layer reduces it back to the desired output dimensionality, ensuring that the model learns expressive intermediate representations without losing correspondence to the physical input and output fields. \\ 

\paragraph{Implementation.} Both lifting and projection layers are implemented using the \codebox{ChannelMLP} class, ensuring parameter sharing across all spatiotemporal locations and resolution invariance. 

By default, these are implemented as two-layer \codebox{ChannelMLP} mapping the input channels to the hidden dimension \((C_{\text{in}} \rightarrow C_{\text{hidden}})\)  and  the hidden representation to the output channels \((C_{\text{hidden}} \rightarrow C_{\text{out}})\), respectively. For the number of channels in the internal layer(s) of these \codebox{ChannelMLP}, the implementation uses $(\text{lifting\_channel\_ratio} \times \text{hidden\_channels} )$ and $(\text{projection\_channel\_ratio} \times \text{hidden\_channels} )$, where \codebox{lifting\_channel\_ratio}, \codebox{projection\_channel\_ratio}, \codebox{hidden\_channels} can be specified as arguments of the \codebox{FNO} class.

\hfill 

\paragraph{Parameter Counts.}  As discussed above, the lifting and projection layers are implemented as two-layer \codebox{ChannelMLP} modules. Typically, $C_{\text{in}}$ and $C_{\text{out}}$ are smaller than the intermediate width $C_h$, which by default is set to $2C_{\text{hidden}}$, where $C_{\text{hidden}}$ denotes the user-defined \codebox{hidden\_channels}. Using the same analysis as in \Cref{sec: ChannelMLP}, this means that the total number of trainable parameters in the lifting or projection layer remains below $4C_{\text{hidden}}^{2}$, usually ranging between 1,000 and 100,000 parameters.

\hfill \\

\subsection{Processing and Embedding Input/Output Data in FNOs}\label{sec: Embeddings}

\vspace{2.5mm}

Neural operators are highly flexible models capable of addressing a wide range of problems that extend far beyond function-to-function mappings of the same dimensionality and resolution. As discussed in \Cref{sec: Mapping of Interest}, several misunderstandings persist regarding their flexibility and range of application, and in particular this is the case for FNOs. With suitable embeddings and resolution-invariant transformations, neural operators can handle a diverse set of tasks well beyond their original function-to-function formulation. 

\vspace{3mm}

\subsubsection{Output Functions Discretized at Higher-Resolution} \label{sec: Output Functions Discretized at Higher-Resolution}

\vspace{3mm}

By design, FNOs generate output functions that can be evaluated at arbitrary resolutions. In \texttt{NeuralOperator 2.0.0}, the \codebox{FNO} class by default produces outputs at the same spatiotemporal resolution as the input. However, this behavior can be modified using the \codebox{resolution\_scaling\_factor} argument, which controls how the resolution of the computational domain evolves across the layers of the FNO.

The \codebox{resolution\_scaling\_factor} parameter provides a flexible mechanism to adjust the resolution at different layers of the model. It supports two modes of operation: (1) when set to \texttt{None}, the resolution remains constant throughout the model; (2) when a list of scaling factors \([s_0, s_1, \ldots]\) is supplied, each layer \(i\) applies its corresponding value \(s_i\). This feature enables precise control over how spatial resolution changes through the model, facilitating the design of architectures that can operate across multiple scales or adapt to problems defined at varying resolutions. 

For example, setting \codebox{resolution\_scaling\_factor} to \texttt{[1, 8, 0.25]} for an input of resolution 64 results in the following progression: the resolution remains \(64 \times 1 = 64\) after the first layer, increases to \(64 \times 1 \times 8 = 512\) after the second layer, and decreases to \(64 \times 1 \times 8 \times 0.25 = 128\) after the final layer, which becomes the output resolution. This provides a simple yet powerful way to control resolution within FNO architectures, allowing smooth transitions between coarse and fine representations directly within the operator's structure.

\hfill \\

\subsubsection{Sinusoidal Embeddings for Constant Parameters} \label{sec: sinusoidal embeddings}

\vspace{3mm}

Many scientific and engineering problems involve models that depend on fixed scalar parameters, such as material properties or physical constants. When these parameters are used as scalar inputs, they offer limited expressive power: a model must learn to infer their global influence from a single constant value. To address this limitation, sinusoidal embeddings can be used to map constant parameters into rich, high-dimensional representations that better interact with the model's internal dynamics.

\mydef{Sinusoidal embeddings} transform parameters into periodic signals composed of sine and cosine functions at multiple frequencies. This transformation, often referred to as \mydef{spectral lifting}, expands the scalar parameter into a basis of oscillatory functions. The resulting representation captures information across several frequency bands, providing the model with a structured and possibly more interpretable way to express nonlinear effects associated with these parameters.

We illustrate here two common formulations for embedding a constant parameter \( p \in \mathbb{R} \) as a concatenation of sinusoidal signals on $[0,2\pi]$ as opposed to a constant signal, both illustrated in \Cref{fig: embeddings}: 
\vspace{2mm}

\begin{itemize}
    \item \textbf{Amplitude Modulation:} the parameter controls the amplitude of each sinusoidal mode, i.e.,
\begin{equation}
[ p\sin(x), \ p\cos(x), \ p \sin(2x), \ p \cos(2x), \ \ldots, \ p \sin(Lx), \ p \cos(Lx)],
\end{equation}
for some integer \(L>0\). Each sine-cosine pair introduces a new oscillatory mode that enriches the representation of the parameter $p$. This expansion allows the model to distinguish between parameters that differ subtly, by encoding them in a smooth manner that spans multiple spectral scales. This approach effectively scales the frequency spectrum uniformly.
\vspace{2mm}
\item \textbf{Frequency Modulation:} the parameter influences the oscillation rate itself, i.e., 
\begin{equation}
[ \sin(cpx), \ \cos(cpx), \  \sin(2cpx), \  \cos(2cpx), \ \ldots, \  \sin(Lcpx), \  \cos(Lcpx)],
\end{equation}
for some scaling constant $c>0$ and integer \(L>0\).
This method changes the frequency content of the embedded signal, allowing different parameter values to generate distinct spectral patterns.  

\end{itemize}

\begin{figure}[t]  
\centering   
\includegraphics[width=\textwidth]{figs/plot_embeddings.pdf}  
\caption{Comparison of amplitude and frequency modulation patterns for sinusoidal embeddings. 
The left panel illustrates amplitude-modulated functions, where the amplitude varies with cosine and sine harmonics scaled by a fixed parameter $p = 1.5$. The right panel illustrates frequency-modulated functions, where the frequency varies with the same parameter, resulting in increasingly dense oscillations.} 
\label{fig: embeddings} \vspace{3mm}
\end{figure}

\hfill 

Different variants of sinusoidal embeddings define how frequencies are distributed across the embedding dimensions. With sinusoidal embeddings, care must be taken to ensure that the frequencies introduced remain within the Nyquist limit of the discretized domain. As discussed in detail in \Cref{sec: Nyquist}, for a domain with \(N\) points, the maximum resolvable frequency is given by the Nyquist frequency \( f_{\text{Nyquist}} = N/2 \). If this condition is violated, higher frequencies will be misrepresented as lower ones, leading to distorted embeddings that no longer capture meaningful variations.  Thus, the choice of \(L\) should reflect both the parameter magnitude and the domain resolution, ensuring that the embedding remains spectrally accurate.

Embeddings can be extended to higher dimensions, by applying the one-dimensional embedding independently along each dimension and concatenating the results along the feature axis. This allows the model to capture independent frequency patterns along each direction while maintaining overall structural coherence. \\ 

\paragraph{Spectral Interpretation and Benefits for FNOs.} From a spectral perspective, embedding constant parameters as multi-frequency signals enables them to interact coherently with the model's convolutional or Fourier layers. Instead of entering as constant scalars, these parameters acquire structured spectral signatures that can modulate the learned representations across frequency bands. This not only improves the model's ability to capture nonlinear dependencies between parameters and spatial fields but also enhances gradient flow and provides a physically interpretable way to understand how parameters influence the solution space.

FNOs learn mappings between functions by operating directly in the frequency domain: they decompose the input into sine and cosine modes, apply learnable spectral transformations, and reconstruct the output in physical space. When constant parameters are embedded through sinusoidal functions, their spectral components naturally align with the Fourier modes used by the FNO. This synergy makes sinusoidal embeddings a principled and powerful mechanism for encoding constant parameters for FNOs, enabling the representation of rich, multi-scale dynamics with fewer layers and improved stability.

\begin{figure}[t]  \vspace{-1mm}
\centering   
\includegraphics[width=\textwidth]{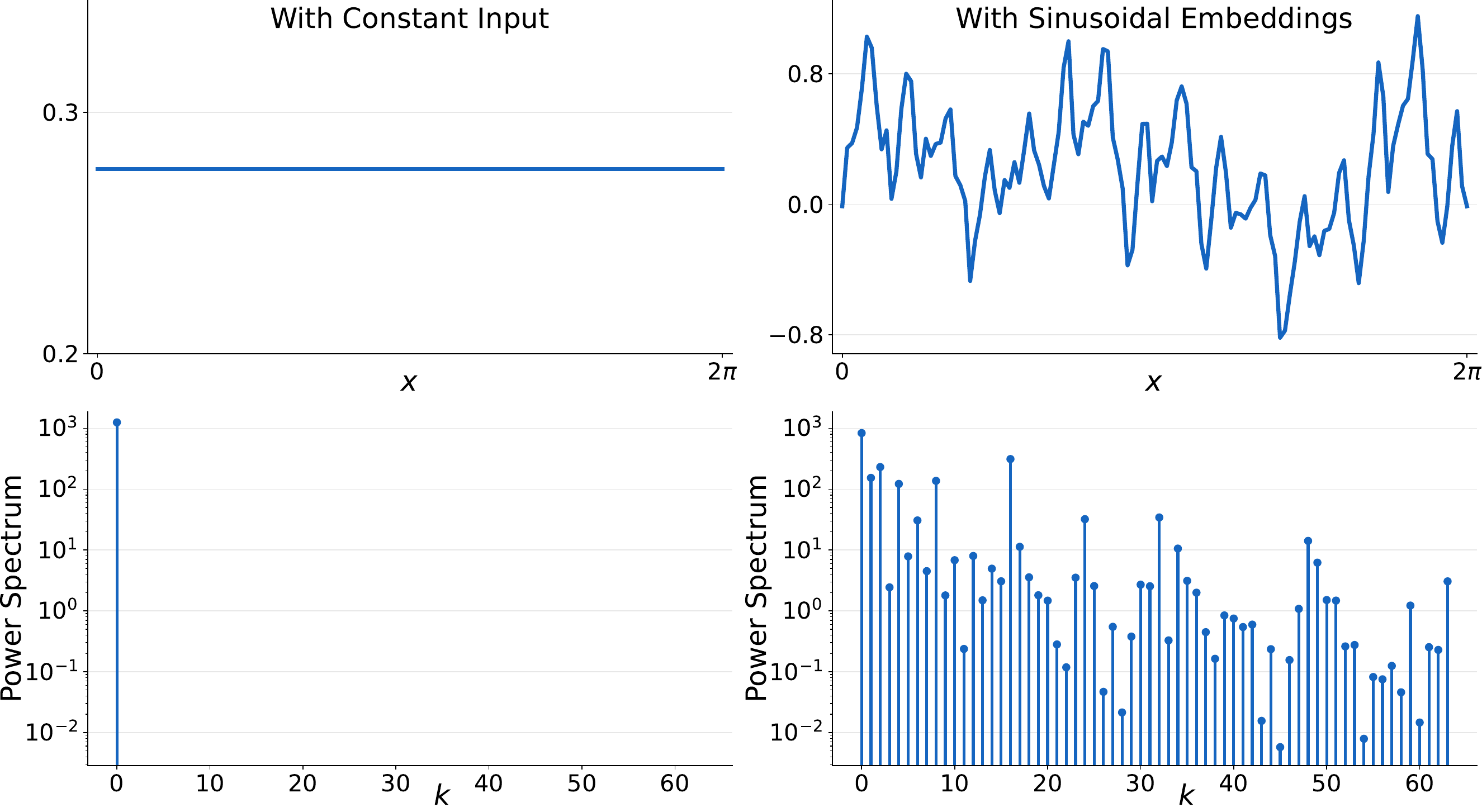}  \vspace{-6mm}
\caption{Effect of sinusoidal embedding on spectral richness. Encoding constant parameters as flat signals yields a spectrum with only the zeroth mode active after the lifting layer, while sinusoidal embedding produces rich spectral activations across all modes, providing the FNO with a more expressive input representation.} 
\label{fig: Embedding MLP Test} \vspace{2mm}  
\end{figure}

To illustrate this, we conducted an experiment where two constant parameters, \( p_1 = 2 \) and \( p_2 = -3 \), were encoded differently into signals of resolution 128 and passed through a randomly initialized 2-layer \codebox{ChannelMLP} with 64 hidden channels and a single output channel. This network emulates the lifting layer in the FNO. The resulting output signals and their power spectrum are displayed in \Cref{fig: Embedding MLP Test}.

In the first setting, \( p_1 \) and \( p_2 \) were represented as constant signals. After passing these two channels through the lifting layer, the output remains constant. Consequently, its spectral representation exhibits a single nonzero component at the zeroth mode, with all other modes inactive. This limited spectral content is what the first FNO layer, and in particular the initial spectral convolution, operates on. Since this representation is spectrally sparse, the learnable convolution has limited expressive power, and a larger number of Fourier layers is often required to capture more complex dynamics with richer spectral representations.

In the second setting, \( p_1 \) and \( p_2 \) were embedded as amplitude-modulated sine and cosine functions with frequencies \( 2^{i} \) for \( i = 0, \ldots, 5 \), resulting in 26 input channels. After passing these inputs through the lifting layer, the resulting signal exhibits a rich spectral distribution, with activations across all frequency modes. This provides the first spectral convolution in the FNO with a rich and diverse spectral representation, making it easier for the model to construct complex representations even within its initial Fourier layer. 

\vspace{3mm}

\paragraph{Implementation.} Two different frequency-modulating sinusoidal embedding variants are implemented in \texttt{NeuralOperator 2.0.0} in the \codebox{SinusoidalEmbedding} class of \codebox{neuralop/layers/embeddings.py}: the transformer-style embedding \texttt{``transformer"}~\citep{vaswani2017attention} which assigns frequencies that decay exponentially, and the NeRF-style embedding \texttt{``nerf"}~\citep{mildenhall2020nerf} which uses powers of two to generate exponentially increasing frequencies. As before, it is important to ensure that all the resulting frequencies in the sinusoidal embeddings do not exceed the Nyquist limit. A tutorial on how to use these sinusoidal embeddings and on their effects is available at \href{https://neuraloperator.github.io/dev/auto\_examples/layers/plot\_sinusoidal\_embeddings.html}{neuraloperator.github.io/dev/auto\_examples/layers/plot\_sinusoidal\_embeddings.html}.

\hfill 

\subsubsection{Multiple Input Functions and Output Functions} \label{sec: Multiple Input Functions and Output Functions}

\vspace{2.5mm}

A widespread yet unfounded misconception is that FNOs can only process a single input function. This misconception typically arises from superficial or biased comparisons to DeepONets~\cite{lu2019deeponet} which explicitly use two branches, occasionally leading to incorrect claims that only DeepONets can handle multiple inputs. This interpretation is entirely incorrect. FNOs can naturally accommodate multiple inputs of different types within a single unified architecture. Each input, whether a continuous field, discrete variable, or scalar quantity, can be embedded in a resolution-invariant manner so that all inputs share a common dimensionality and resolution. Once embedded, these inputs are simply concatenated as separate channels of the same functional representation. This design allows the FNO to capture cross-variable dependencies and integrate heterogeneous sources of information, such as physical parameters, boundary conditions, and auxiliary data, while preserving its inherent resolution invariance and functional generality. 

In addition, FNOs can seamlessly handle input functions defined on different spatiotemporal intervals and resolutions. In practice, each input function is discretized according to its own grid, which may differ in spacing, extent, or resolution. These inputs can be embedded into a shared latent representation using resolution-invariant transformations (e.g. by passing them through a spectral convolution with output resolutions different from input resolutions), allowing the model to align and process them consistently regardless of their original discretization. 

In a similar manner, FNOs can also produce multiple outputs with ease. The model can jointly predict several functions at once as different channels of the output. Additional resolution-invariant output layers can also be designed so that these output channels later diverge into distinct processing branches, enabling specialized downstream operations for each quantity. This flexibility supports both tightly coupled and loosely related outputs. Since the FNO operates in function space, each output channel can be evaluated at arbitrary resolutions and domains without retraining or modifying the architecture.

\hfill

\subsubsection{Handling Different Dimensionalities in Inputs and Outputs} \label{sec: Handling Different Dimensionalities in Inputs and Outputs}

In scientific and engineering systems, quantities of interest often exist across spaces of different dimensions. A problem may involve scalar constants, one-dimensional temporal signals, and multidimensional spatiotemporal fields. FNOs can accommodate this heterogeneity through resolution-invariant embedding and transformation mechanisms that map all inputs into a shared functional space. This ensures that the model interprets every variable as a function defined on the same domain and with compatible discretization.

Scalar and low-dimensional quantities can be lifted into higher-dimensional functional representations using embedding strategies. As an example, we discussed in \Cref{sec: sinusoidal embeddings} how scalar parameters can be embedded as functions using sinusoidal embeddings in a way that provides a rich spectral signature aligned with the Fourier basis used by the FNO. More generally, any low-dimensional input can be mapped into a higher-dimensional space using similar resolution-invariant transformations, ensuring that all input channels share compatible resolutions and can be fused seamlessly within the model.

Differences in dimensionality also arise between inputs and outputs. Many operator learning problems require mappings between functions of different dimensionalities, such as transforming a two-dimensional spatial field into a one-dimensional temporal response, or predicting multidimensional spatial fields from scalar or low-dimensional inputs. Resolution-invariant transformations that increase or decrease dimensionality can be leveraged for that purpose. New dimensions can be introduced by broadcasting an existing tensor along an additional axis, optionally combined with learned or predefined embeddings to introduce structured variation along the new dimension. Conversely, dimensionality can be reduced through principled resolution-invariant operations such as integration along selected axes. Naive summation is not suitable, since the result changes significantly with the resolution and typically diverges. Averaging, on the other hand, preserves resolution invariance: each sample contributes proportionally, and as the discretization becomes finer, the discrete average converges to a continuous integral of the function along that dimension. This approach ensures that the resulting reduced representation remains consistent under changes in resolution. Using this reasoning, higher-dimensional functions can be collapsed to lower-dimensional signals, scalar quantities, or categorical outputs for regression and classification tasks while maintaining stability across resolutions.

\hfill

\subsubsection{Multi-Resolution Datasets} \label{sec: Multi-Resolution Datasets}

\vspace{2mm}

FNOs are inherently capable of handling data defined at different spatial or temporal resolutions without requiring any architectural changes or interpolation layers. This property arises from their spectral domain formulation, which allows the model to operate directly in function space rather than on fixed discretizations. As a result, FNOs can process inputs and produce outputs at arbitrary resolutions, making them naturally suited for applications where data vary in granularity. The implementation of the \codebox{FNO} class in \texttt{NeuralOperator 2.0.0} reflects this principle, enabling the same model to learn across data samples with different resolutions without any modification to the underlying architecture. 

This is particularly valuable in science and engineering, where datasets are often available at multiple resolutions: coarse simulations efficiently capture global, large-scale dynamics, while fine-grained simulations resolve complex, small-scale features with higher fidelity. Leveraging these complementary sources of information can greatly enhance model efficiency, accuracy, and generalization. Multi-resolution training strategies can also be developed to improve efficiency and generalization, and will be discussed in \Cref{sec: ifno}. This can be particularly helpful for zero-shot super-resolution, since a model which experienced data at multiple resolutions is less likely to overfit any single resolution, and thus will typically generalize better to unseen resolutions. Super-resolution capabilities of FNOs will be discussed in more detail in \Cref{sec: super-resolution}.

\hfill

\subsubsection{Handling Complex-Valued Data} \label{sec: Handling Complex-Valued Data}

\vspace{2mm}

FNOs can be applied to complex-valued data. \texttt{NeuralOperator 2.0.0} introduces native support for complex-valued functions in its \codebox{FNO} class through the boolean argument \codebox{complex\_data}. When \codebox{complex\_data} is set to \texttt{True}, the model internally wraps the lifting and projection layers, as well as all individual operations in the FNO blocks, with complex-valued module variants implemented via the \codebox{ComplexValued} wrapper class in \codebox{neuralop/layers/complex.py}. 

This wrapper generalizes standard neural network layers to operate on complex data by creating two coupled versions of each layer that process the real and imaginary components separately and then recombine them in a consistent manner to preserve their complex structure. In the \codebox{FNO} implementation, this mechanism is applied not to the full model but individually to each component of the lifting, Fourier, and projection layers, ensuring that every transformation takes into account the complex-valued structure of the data. As a result, the FNO can accurately model mappings between complex-valued function spaces, making it a versatile operator-learning framework for both real and complex data without requiring users to manually separate or recombine components.

\hfill

\subsection{Handling Non-Periodic Domains} \label{sec: FNOs Non-periodic domains}

\vspace{3mm}

As discussed in~\Cref{sec: Non-periodicity}, spectral methods are inherently designed for periodic problems. However, it is a common misconception that this means FNOs cannot be used for non-periodic problems.

First, the FNO architecture is not composed solely of spectral convolution operations. The positional embeddings, the additional skip connections, and the linear transform acting on the chammel dimension (the \codebox{ChannelMLP}) can help ensure non-periodic boundary conditions. 

Additionally, as discussed in the FNO~\cite{li2020fourier} and PINO~\cite{li2024physics} papers, the physical domains can be extended into larger computational domains where periodic boundary conditions are enforced. Padding zeros to the input and computing the training loss on the original domain often suffices, as the FNO will automatically generate a smooth output on the extended domain. Domain padding is implemented in \texttt{NeuralOperator 2.0.0} in \codebox{neuralop/layers/padding.py}, and is also available through the \codebox{domain\_padding} argument of the \codebox{FNO} class. This argument accepts either a single numeric value specifying the padding percentage applied uniformly across all dimensions or a list of values \([p_1, p_2, \ldots, p_N]\) defining padding percentages for each dimension individually. Alternative padding strategies exist, such as mirror padding, where the signal is extended by reflecting it about one or both endpoints, thereby guaranteeing continuity of the extended signal. 

As discussed in \Cref{sec: Non-periodicity}, a more principled approach is to construct a smooth, periodic extension of the non-periodic signal. Fourier continuation methods~\citep{Fourier_continuation_2011, fontana2020Fourier, bruno2022two} extend the non-periodic function to a periodic one over an enlarged domain using smooth polynomial or spline-based interpolation. The resulting continuation preserves both the continuity and differentiability of the function across the artificial boundary, allowing Fourier-based techniques to achieve super-algebraic convergence even for originally non-periodic data. In this way, Fourier continuation provides a bridge between the mathematical elegance of spectral methods and the practical need to handle general, non-periodic problems reliably. While padding can be sufficient in many contexts, Fourier continuation provides a superior extension scheme when high precision is required, for instance in spectral derivative computations for PINOs, leading to the FC-PINO architecture~\citep{ganeshram2025fcpinohighprecisionphysicsinformed}. 

The Fourier continuation classes are available in the \texttt{NeuralOperator 2.0.0} library in\codebox{neuralop/layers/fourier\_continuation.py}, together with a tutorial detailing how to use them in the online documentation at \href{https://neuraloperator.github.io/dev/auto_examples/layers/plot_fourier_continuation.html}{neuraloperator.github.io/dev/auto\_examples/layers/plot\_fourier\_continuation.html}.

A smooth periodic extension of a given non-periodic signal can also be constructed by optimizing its spectral content on the extended domain, as discussed in \Cref{sec: Non-periodicity}. In this approach, the grid is extended by adding a small number of unknown points to the left and right of the original interval, and the values at these new grid points are determined by minimizing an appropriate spectral norm.

Additionally, specialized extension and padding strategies can be designed to explicitly enforce desired (non-periodic) boundary conditions. For example, interpolation-based extensions can be employed to smoothly continue the function beyond the original domain while constraining derivative values or other physical quantities at the boundaries. Such approaches ensure that the learned representation remains consistent with the underlying physical or mathematical constraints of the problem.

More generally, the FNO can be applied to arbitrary geometries by embedding them into a periodic cube or torus, for instance via the geometric encoder-decoder strategies discussed in \Cref{sec: Point Clouds}.

\hfill 

\subsection{Super-Resolution with FNOs}
\label{sec: super-resolution}

\vspace{2mm}

Many scientific simulations and experimental measurements are constrained by computational or physical limitations that restrict spatiotemporal resolution. The ability to reconstruct high-resolution fields from coarser data, called \mydef{super-resolution}, is both practically valuable and scientifically informative.

Traditional super-resolution approaches, such as interpolation or convolutional neural networks (CNNs), are typically tied to fixed discretizations and local receptive fields. CNN-based approaches often require retraining for each specific grid size or resolution ratio. FNOs, by contrast, operate on functions rather than fixed arrays. Their learned mapping is inherently defined in continuous space, which naturally enables evaluation at arbitrary resolutions without retraining or architectural modification.

This property allows FNOs to perform super-resolution by simple evaluation on finer grids. Suppose the model is trained on inputs and outputs sampled at a coarse resolution. At inference, one can resample the input field to a finer grid and apply the learned operator directly to obtain an output at the new resolution. This procedure is particularly effective when combined with multi-resolution training or incremental strategies such as iFNO (see \Cref{sec: ifno}), where the model has already learned to represent a range of spatial frequencies. Even if trained primarily on coarse data, the FNO's spectral filters can extrapolate to finer discretizations due to their frequency-domain formulation, enabling meaningful high-resolution predictions. \\

\paragraph{Implementation.} No architectural changes are required for super-resolution. The same FNO model can be evaluated on arbitrary grid sizes at inference by adjusting only the discretization of the input field and the inverse Fourier transform. This flexibility stems from the operator's formulation in continuous spectral space, making it agnostic to the discretization used at training time. Thus, super-resolution emerges as an intrinsic capability of the model, not as an additional module or loss component. When super-resolution is a downstream task of interest, model hyperparameters must be tuned with additional care to limit aliasing and artifacts at higher resolutions, as discussed in \Cref{sec: n_modes}. \\

\paragraph{Zero-Shot Super-Resolution.} Regarding terminology, in the context of operator learning, \mydef{zero-shot super-resolution} refers to evaluating a model trained exclusively on coarse-resolution data directly at finer resolutions without any retraining or fine-tuning. By contrast, \mydef{few-shot super-resolution} involves a brief adaptation stage using only a small number of high-resolution examples to refine the model's predictions on fine grids, leveraging prior knowledge acquired from extensive coarse-resolution training. \\

\paragraph{Error Analysis.} A high-level error analysis for neural operators evaluated on discretizations different from those used during training is provided by \citet{Berner2025Principles}. This encompasses zero-shot super-resolution as a special case, where a model trained on coarse discretizations is applied directly to finer grids without retraining. The analysis shows that the total prediction error of a neural operator trained on discretization~$X$ and evaluated on an unseen discretization $Y$ can be decomposed into three main components: the smallest achievable approximation error of the true (continuous) operator within the neural operator class, and the discretization errors associated with $X$ and $Y$. Under suitable regularity assumptions, the discretization errors vanish as the resolution increases, due to convergence properties of neural operators~\citep{Berner2025Principles,kovachki2021universal}. Consequently, when the model is sufficiently expressive and trained to near-optimality on fine enough data, the prediction error on higher-resolution discretizations can be made arbitrarily small.

In practice, we recommend three complementary steps: choose suitable numerical settings, use multi-resolution training to strengthen generalization, and incorporate physics-informed losses when additional fine-scale accuracy is needed. These steps address different problems.

A guiding principle for training neural operators is:
\begin{quote}
    \mydef{\emph{A neural operator should be trained as a neural operator when evaluation at new resolutions is the goal, not as a neural network with a fixed discretization.}}
\end{quote}

\hfill

\subsubsection{Numerical Implementation and Hyperparameter Selection}
\label{sec: sr_implementation}

\vspace{2mm}

Hyperparameters should be chosen for the intended cross-resolution use of the model, not only for accuracy on the training grid. In particular, more Fourier modes are not always better. Nonlinearities generate additional frequencies, which can exceed what a coarse computational grid can resolve and alias into lower frequencies. A model may then learn to use these aliased signals and perform poorly on a finer grid where the aliasing changes. Reducing the retained bandwidth or evaluating nonlinearities on a finer intermediate grid can help. \Cref{sec: n_modes} explains these mechanisms and the corresponding recommendations.

Preprocessing and boundary treatment also matter. Downsampling can introduce aliasing before the model sees the data, while an inappropriate periodic extension can introduce boundary oscillations. Domain padding or Fourier continuation can help with non-periodic fields. These choices should remain consistent with the physical domain as the resolution changes, as discussed in Sections~\ref{sec: downsampling} and~\ref{sec: FNOs Non-periodic domains}. Standard practices for optimization and validation remain essential, as described in \Cref{sec: General Machine Learning Practices}.

With suitable numerical settings, an FNO trained on coarse data can produce useful finer-grid predictions without systematic spurious patterns. This has been demonstrated for instance in the original FNO paper~\citep{li2020fourier} and in climate-downscaling applications~\citep{yang2023fourier,jiang2023efficient}. Prediction errors may nevertheless increase, and some small-scale features may remain inaccurate.

\hfill 

\subsubsection{Multi-Resolution Training}
\label{sec: sr_training}

\vspace{2mm}

Recall the earlier guiding principle: \emph{a neural operator should be trained as a neural operator when evaluation at new resolutions is the goal, not as a neural network with a fixed discretization.} 

Training should encourage learning a mapping between functions rather than only learning at a single fixed resolution. If all supervision and model selection are confined to a single poorly resolved grid, the model can learn discretization-specific behavior and become sensitive to resolution changes, much like a conventional fixed-grid neural network.

Multi-resolution training is a natural way to exploit the operator-learning paradigm. The same parameters describe one operator, while different grids provide different discretizations of its input and output functions. Training across these grids discourages dependence on any single discretization. Coarse examples provide inexpensive information about large-scale behavior, while finer examples help constrain smaller structures and improve spectral accuracy. The losses should approximate the same function-space discrepancy at every resolution, as discussed in \Cref{sec: data losses}.

Sections~\ref{sec: Multi-Resolution Datasets} and~\ref{sec: ifno} discuss multi-resolution datasets and training strategies, including the incremental FNO~\citep{George2022Incremental}. Other examples include multi-resolution active learning~\citep{Li2024MultiResolution} and multi-level Monte Carlo training~\citep{Rowbottom2025Multilevel}. Multi-resolution training is also used within broader applications. For example, FunDPS uses a coarse-to-fine training schedule for PDE reconstruction and inverse problems~\citep{yao2025guided}.

The experiments of \citet{Berner2025Principles} illustrate why architecture and training should be considered together. They show that mixed-resolution training improves unseen-resolution performance for a neural operator with a kernel-interpolated U-Net backbone, whereas the conventional U-Net generalizes poorly. Training across resolutions therefore complements the operator architecture rather than replacing it. Single-resolution training can already support successful super-resolution, but multi-resolution training can make that capability more robust. 

\hfill

\subsubsection{Physics-Informed Losses for Fine-Scale Accuracy}
\label{sec: sr_physics}

\vspace{2mm}

Coarse observations may not provide enough information to determine small-scale behavior accurately. When the governing equations are known, physics-informed losses provide additional guidance by penalizing violations of the PDE and its boundary conditions. These losses can be evaluated on finer grids without requiring corresponding high-resolution solution labels.

This is the strategy developed in PINO~\citep{li2024physics}: combine coarse-resolution solution data with physics constraints evaluated at finer resolutions. Data teach the model the solution patterns represented in the observations, while the equations help constrain behavior that those observations do not adequately resolve. Improvements in spectral fidelity are illustrated in the PINO paper~\citep{li2024physics} and by \citet{azizzadenesheli2024neural}. Their spectrum-recovery example also includes refinement of individual test predictions by minimizing the PDE residual after training.

\Cref{sec: PINO} explains how to combine data and physics losses, compute the required derivatives, and optionally refine individual predictions. For super-resolution, physics can provide information beyond the coarse labels. It can therefore help close an accuracy gap that changing the evaluation grid alone cannot remove.

\hfill

\subsubsection{Misinterpretations in Recent Negative Results}
\label{sec: sr_negative_results}

\vspace{3mm}

\paragraph{Sakarvadia et al.}\!\!\cite{Sakarvadia2026FalsePromise} claim that machine-learned operators are ``not able to perform accurate inference at resolutions different from those on which they were trained.'' Their experiments do not establish this strong and general claim, which conflicts with earlier successful demonstrations~\citep{jiang2023efficient,li2020fourier,yang2023fourier}. In particular, their interpretation does not adequately separate implementation errors, numerical configuration choices, and remaining prediction errors.

The pronounced stripes and line artifacts reproduced in their results arise from incorrect handling of Hermitian symmetry in the inverse real Fourier transform. This implementation error appeared on certain GPU/software configurations, particularly at power-of-two grid sizes. Contrary to the authors' claim, these artifacts arise from an implementation issue rather than from the FNO architecture itself. The implementation issue has been corrected in newer versions of \texttt{neuraloperator}. It is now possible to set \texttt{enforce\_hermitian\_symmetry} to \texttt{True} when defining the FNO, which resolves that issue. Setting it to \texttt{False} restores the legacy implementation for diagnostic comparisons.

\begin{figure}[t]  \vspace{-3mm}
\centering   
\includegraphics[width=\textwidth]{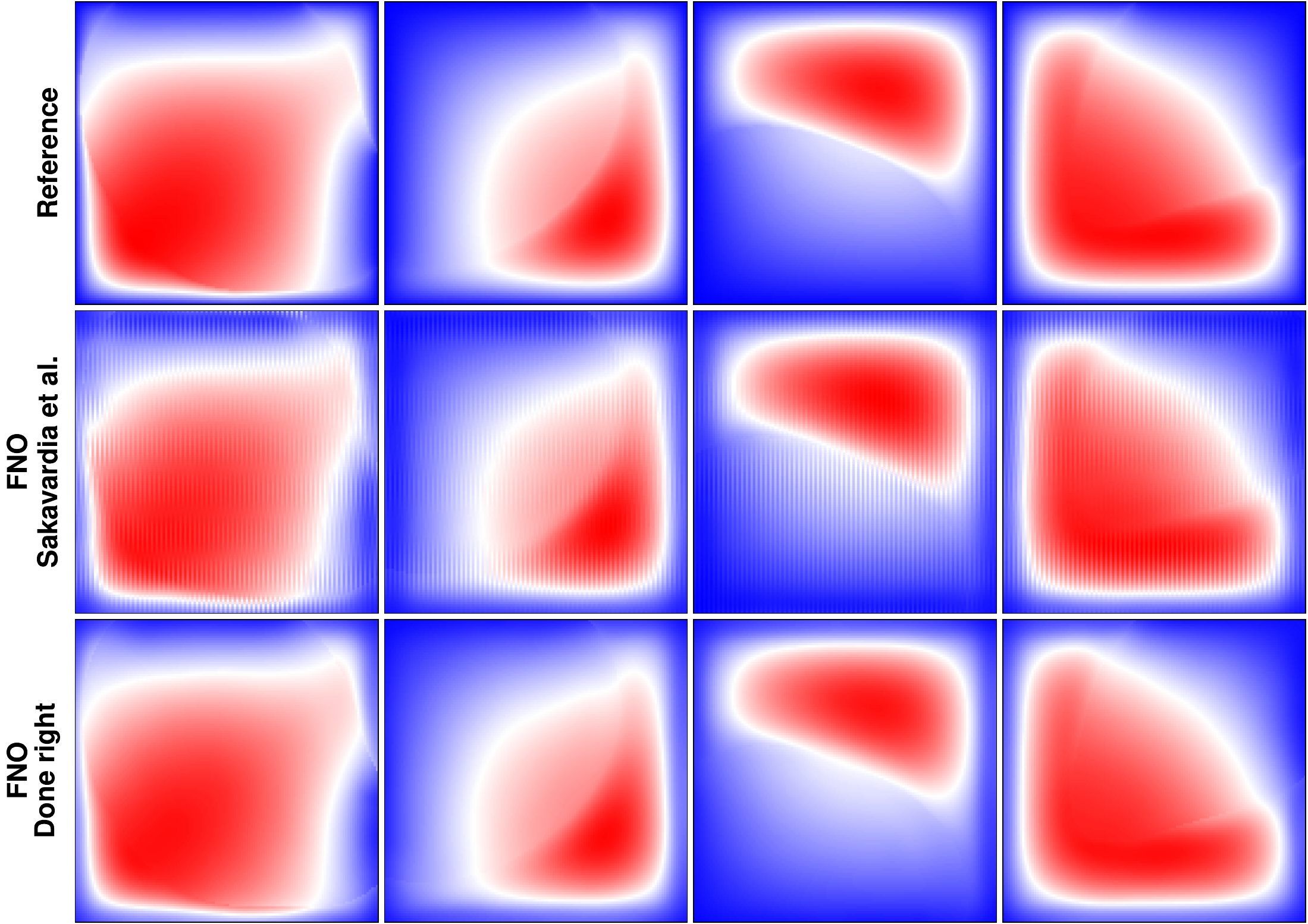}  \vspace{-6.5mm}
\caption{Zero-shot super-resolution for the Darcy flow problem of~\citet{Sakarvadia2026FalsePromise}. Each column shows a different test example. The rows display reference solutions (top), FNO predictions using the setup of \citet{Sakarvadia2026FalsePromise} (middle), and FNO predictions after correcting the implementation issue and tuning the hyperparameters correctly (bottom). The models are trained at $16\times16$ and evaluated at $128\times128$ without fine-resolution supervision or fine-tuning. The pronounced line artifacts in the middle row arise from an inverse-transform implementation issue. Correcting this issue and choosing suitable Fourier bandwidth and domain padding produces finer-grid predictions without artifacts, as can be seen on the bottom row.} 
\label{fig: aliasing_paper_fix}  \vspace{-2mm}
\end{figure}

The implementation defect was compounded by unsuitable numerical settings. \citet{Sakarvadia2026FalsePromise} also omitted domain padding and used a number of Fourier modes that was slightly too large for the coarse training resolution. Padding can improve the treatment of non-periodic boundaries, while reducing the retained bandwidth helps limit aliasing from frequencies generated by nonlinearities. These considerations remain relevant after the GPU implementation has been corrected. The experiments therefore required both a software correction and more careful hyperparameter selection. After fixing the implementation and better tuning the FNO hyperparameters, we can obtain good super-resolution results without artifacts, as shown in \Cref{fig: aliasing_paper_fix}. While high-frequency spectral discrepancies remain, substantially cleaner predictions are possible without additional fine-resolution supervision.

The authors were made aware of the implementation issue, reran their code with the corrected implementation and acknowledged that it produced better predictions without artifacts. Unfortunately, despite recommendations to revise their results and claims, they decided to publish them unchanged, presenting artifacts they knew arose from an implementation issue as evidence against zero-shot super-resolution. This raises serious concerns about the reliability of the paper's results and conclusions more broadly. They also present multi-resolution training as its remedy, although this strategy was already established in prior works~\citep{George2022Incremental,Li2024MultiResolution,Rowbottom2025Multilevel}. The final version cites these precedents, but still contrasts an alleged failure of neural operators with an established training strategy that builds on their capabilities. Multi-resolution training is a natural use of the operator-learning paradigm, not a new remedy that demonstrates its failure. 

\paragraph{Colagrande et al.}\!\!\!\!\cite{Colagrande2026Limits} conduct cross-resolution experiments with FNOs. The authors train an FNO on an $85\times85$ grid and evaluate it on a $211\times211$ grid, retaining the full Fourier bandwidth of the training grid and thus leaving no spectral margin for frequencies generated by nonlinearities. The study also compares no padding with a fixed padding width of ten grid points, which corresponds to different physical extension lengths at the two resolutions. Together with the lack of experiments on multi-resolution training, these choices leave important sources of resolution dependence uncontrolled. Moreover, layerwise spectra alone do not establish that newly generated frequencies are aliasing errors. The experiments therefore demonstrate the sensitivity of their chosen setup rather than establish a general limitation of properly configured FNOs, offering very limited insight into the super-resolution capabilities of neural operators.

\hfill 

\subsection{Examples of Successful Applications of Neural Operators} \label{sec: Applications}

\vspace{3mm}

FNOs are particularly well-suited for learning operators in science and engineering. We highlight in this section the versatility of the FNO and its extensions by mentioning how they have already been successfully applied in very diverse fields.

FNOs have fostered significant advancements in computational mechanics, including modeling porous media, fluid mechanics, and solid mechanics~\citep{choubineh2023fourier,you2022learning} for instance by learning stress and strain fields~\citep{khorrami2025physics,Rashid2022}. \textsc{FNO}s constitute the first machine learning approach to successfully model turbulent flows with zero-shot super-resolution capabilities~\citep{li2020fourier}, and have been used in large eddy simulations of three-dimensional turbulence~\citep{zhijie2022fourier}. 

\textsc{FNO}-based architectures can match the accuracy of physics-based numerical weather prediction systems while being orders-of-magnitude faster~\citep{thorsten2023fourcastnet,mahesh2024huge}, in particular with the Spherical FNO~\cite{bonev2023spherical} which facilitates stable simulations of atmospheric dynamics on Earth. The super-resolution capabilities of FNOs have also been leveraged for downscaling of climate data, i.e., predicting climate variables at high resolutions from low-resolution simulations~\citep{jiang2023efficient,yang2023fourier}. FNOs have also been used for ocean modeling~\citep{sun2024streamliningoceandynamicsmodeling}. 

In the geosciences, \textsc{FNO}s have been used for seismic wave propagation and inversion~\citep{KonukShragge2025,lehmann2023fourier,yang2021seismic,Yang2023}, in hydraulic tomography~\Citep{Guo2024_RGA_FNO_HT}, and to model multiphase flow, which is critical for applications such as contaminant transport, water-flooding modeling, carbon capture and storage, hydrogen storage, and nuclear waste storage~\citep{Ma2024Enhancing,gege2022u,Wen2023CO2}. In the biomedical domain, neural operators have been used to enhance medical imaging~\citep{Dai2023,jatyani2024unified,jatyani2025coarsetofine,Wang2025Ultrasound,wang2026coarse} and also to improve the design of medical devices, such as catheters with reduced risk of urinary tract infection~\cite{Zhou2024}.

\begin{figure}[t]    
\centering   \
\includegraphics[width=0.99\textwidth]{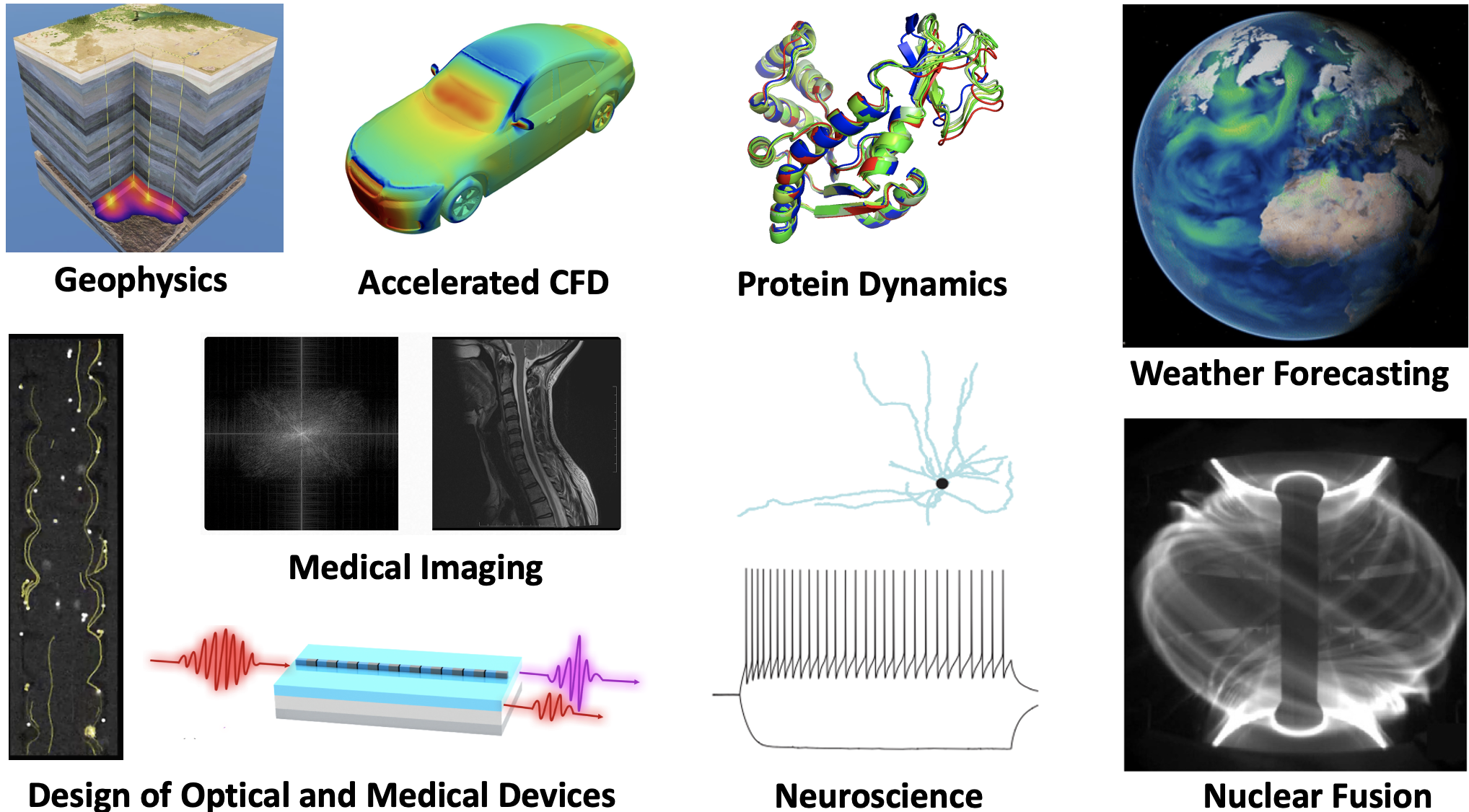}  \vspace{-1.5mm}
\caption{Examples of successful applications~\citep{duruisseaux2025photonics,ghafourpour2025noble,Gopakumar2024,jatyani2024unified,thorsten2023fourcastnet,li2024geometry,wen2022,xu2024equivariant,Zhou2024} of FNOs and variants.} \label{fig: Applications} 
\vspace{3mm} \end{figure}

Other application areas of FNOs include nuclear fusion (to accelerate magnetohydrodynamics simulations for plasma evolution~\citep{Carey2025,Gopakumar2024,Pamela2025}), computer vision (for image classification~\citep{kabri2023resolution}, image segmentation~\citep{wong2023fnoseg3d}, and as token mixers in vision transformers~\citep{guibas2021adaptive}), photonics simulations~\citep{duruisseaux2025photonics,gu2022neurolight,zhu2024pace}, neuroscience~\citep{centofanti2024learning,ghafourpour2025noble,ghafourpour2026capturing,pellegrini2025learning}, and simulations of quantum systems~\cite{pipi2025inversefno,pipi2026,rychetsky2026,shah2024}.

FNOs and their foundational components have also been employed to extend generative modeling and sampling approaches to the operator setting and function spaces. For example,
Generative Adversarial Neural Operators (GANOs)~\citep{rahman2022generative} generalize generative adversarial networks to function spaces. Diffusion FNOs (DiffFNOs)~\citep{liu2025difffnodiffusionfourierneural} directly integrate diffusion processes with FNO architectures to enable efficient, arbitrary-scale image generation. Score-based diffusion models were also formulated in function space using denoising diffusion operators (DDOs)~\citep{lim2025score}. Similarly, Operator Flow Matching (OFM)~\citep{shi2025stochasticprocesslearningoperator} extends flow matching to learn a stochastic process on function spaces for functional regression. Recently, Function-space Diffusion Posterior Sampling (FunDPS)~\citep{yao2025guided} introduced a pioneering framework for conditional sampling in PDE-based inverse problems, leveraging neural operators for discretization independence and plug-and-play guidance for flexibility across different measurement models.

\hfill

\subsection{Inverse Design with Neural Operators} \label{sec: Inverse Design}

\vspace{2mm}
 
Once trained, neural operators can produce high-resolution predictions at a fraction of the computational cost of traditional numerical solvers, often achieving speedups of several orders of magnitude, from 100$\times$ to 1,000,000$\times$. This efficiency makes them ideal for inverse design problems, where a large number of model evaluations are required to search for optimal configurations. In addition, (most) neural operators (including FNOs) are differentiable, which further enhances their value for design tasks, since gradients of any objective with respect to the input parameters can be obtained directly through backpropagation. These two properties, speed and differentiability, make neural operators a natural foundation for combining data-driven simulation with optimization.

In the context of inverse design, a trained neural operator acts as a fast surrogate for the forward model that maps design parameters to physical responses. Once this mapping is available, one can define an objective function that evaluates how well a design achieves desired target responses or optimizes specific quantities of interest relevant to system performance. The inverse problem is formulated as the minimization of this objective function with respect to the design parameters. 

A first approach is to sample the design space extensively by evaluating a dense grid of design parameters, running the corresponding simulations with the surrogate model, and ranking the outcomes using an objective function that encodes the design goal. When each evaluation requires a full high fidelity numerical solve, a large exploration becomes impractical. The speedups provided by the FNO (or another neural operator) typically allow millions of predictions to be generated within minutes on a single GPU, in regimes where traditional numerical simulations would require months or years of computation. This ability makes it feasible to explore, tune, and refine design parameters within a single day and provides a more global view of the overall design landscape than local iterative methods such as gradient-based optimization, which depend on good initialization, carefully chosen objective functions, and can converge to local optima.

A complementary strategy exploits the fact that the neural operator is differentiable so that the entire design pipeline becomes amenable to gradient based optimization. Gradients are propagated through the operator to update the design variables efficiently, allowing the system to move iteratively toward configurations that achieve the target performance. This approach removes the need for repeated calls to an expensive high fidelity solver and enables direct optimization in high dimensional design spaces. In many settings gradient based optimization requires far fewer evaluations than extensive sampling and becomes increasingly attractive as dimensionality grows and exhaustive search becomes prohibitively costly.

\begin{figure}[t]  
\centering   
\includegraphics[width=\textwidth]{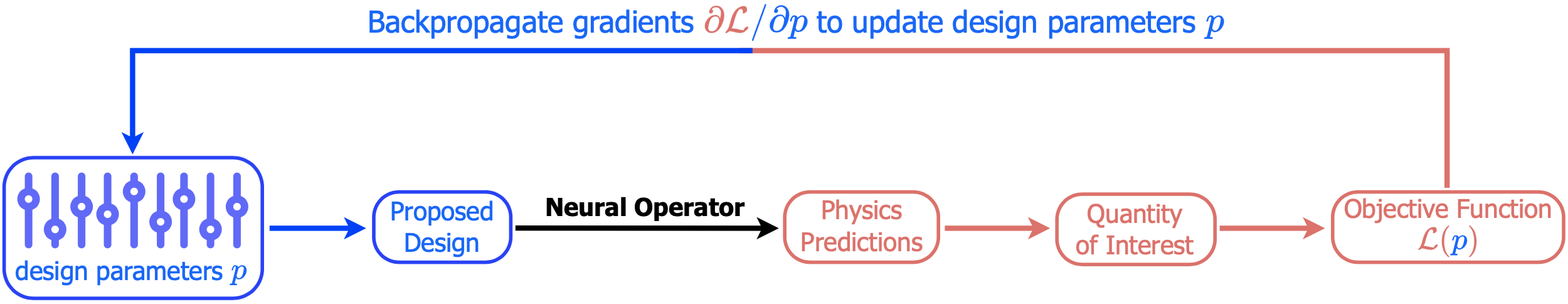}  \vspace{-2.5mm}
\caption{The gradient-based inverse optimization design pipeline, accelerated using a differentiable neural operator for forward modeling. A collection of design parameters $p$ specifies a tentative design. After proper encoding, the design is passed through the differentiable neural operator which predicts certain physics variables from which quantities of interest can be derived. A loss function $\mathcal{L}(p)$ is then evaluated, and gradients are backpropagated through the entire pipeline so the design parameters can be updated through gradient-based optimization. \label{fig: Inverse Design} } \vspace{5mm}
\end{figure}

The effectiveness of neural-operator-based inverse design depends critically on the fidelity and robustness of the learned forward model. For the gradient-based optimization to succeed, the neural operator must not only approximate the underlying physics accurately but also provide reliable gradients that capture true physical sensitivities. This requires the operator to generalize well across the entire design space, including configurations that may differ substantially from those seen during training. To achieve this, the training dataset should encompass a sufficiently broad and diverse range of input conditions and physical regimes. Incorporating variations in geometry, initial conditions, boundary conditions, or other external parameters helps prevent overfitting to narrow regions of the design space and ensures that the learned mapping remains smooth and physically consistent.

\hfill

\paragraph{Design Space.} The central aspect of inverse design is the definition of the design space. The design space encompasses all admissible inputs that describe the system under consideration. Its formulation must balance physical realism with diversity and computational tractability. If the space is too unconstrained, optimization may yield designs that are physically meaningless or not manufacturable. Conversely, an overly restrictive space may prevent the discovery of novel, high-performing designs. A practical approach is to use compact parameterizations that preserve essential features of the design while reducing dimensionality. Low-dimensional parameterizations of designs can also simplify gradient computation and make inverse optimization more stable.

\vspace{1mm}

Enforcing known and desired constraints plays a key role in ensuring the feasibility and physical validity of the final designs. These can be incorporated directly into the definition of the design space, for example by constructing the parameterization so that it only produces valid designs. Constraints can also be introduced through additional terms in the inverse optimization objective that penalize violations of known physical or manufacturing limits. In cases where strict feasibility is required, projection methods can be used to map intermediate solutions back into the admissible region after each optimization step. The choice of constraint handling strategy often depends on the sensitivity of the system to infeasible designs and the degree of flexibility allowed in the search process. \\

\paragraph{Distribution Shifts.}  A practical challenge can arise from distribution shift during inverse optimization. As the design variables are updated iteratively to improve the objective, the optimization trajectory may move into regions of the design space that were underrepresented or entirely absent in the training dataset. In such cases, the neural operator begins to extrapolate beyond its training distribution, and its predictions of system behavior can become unreliable. Even small inaccuracies in the forward model can lead to large deviations in the computed gradients of derived quantities of interest, causing the inverse optimization to diverge or converge toward physically meaningless solutions. This issue is particularly critical in problems where the objective depends on delicate nonlinear interactions, since errors can compound rapidly over successive iterations.

\vspace{1mm}

To ensure robust performance, the training data should ideally cover the entire relevant design space, capturing the diversity of conditions that the optimizer is likely to encounter. When full coverage is infeasible, an alternative strategy is to restrict the optimization to a subset of the parameter space that lies within or near the training distribution, thereby avoiding extrapolation into regions where the model's predictions might be unreliable. 

\vspace{2mm}

Throughout the optimization process, it is important to periodically assess the accuracy of the neural operator by comparing its outputs against those of a high-fidelity numerical solver. This validation step helps confirm that the discrepancy between the surrogate model and the ground-truth solver remains within acceptable limits. Monitoring how far the current design candidate lies from the training data manifold using appropriate distance metrics provides a practical safeguard against distribution shift. When the optimization process consistently moves into regions of interest that are insufficiently represented during training, the operator can be augmented with new data points from these regions and fine-tuned to restore predictive accuracy. This adaptive retraining approach gradually aligns the surrogate's domain of validity with the evolving design space explored by the optimizer. Uncertainty quantification methods can further provide online diagnostics of the model's reliability. 

\vspace{1mm}

In all cases, the final designs obtained from neural-operator-based optimization should be validated using either the full numerical solver or experimental testing. Even when uncertainty monitoring and retraining are employed, small modeling inaccuracies can accumulate during optimization and lead to suboptimal or unphysical results. Evaluating the final candidates with the reference solver verifies that the surrogate model has not introduced significant bias and that the predicted quantities of interest are consistent with true system behavior. This final verification step is essential for ensuring that neural-operator-driven design pipelines produce solutions that are both highly desirable and physically reliable.

\hfill

\paragraph{Examples of Successful Applications.}

In geophysics and subsurface modeling, neural operators have replaced iterative PDE solvers in hydraulic tomography and seismic inversion workflows, accelerating parameter reconstruction while maintaining high accuracy~\citep{Guo2024_RGA_FNO_HT,yang2021seismic}. They have also been applied to accelerate metamaterials design~\citep{Liu2025SDF,Schmid2025DAGA}. In biomedical engineering, neural operators have learned mappings from device geometry to fluid or microbial responses, enabling the design of devices such as anti-infection catheters that were experimentally validated and demonstrated significant performance gains~\citep{Zhou2024}. Applications in aerospace engineering include shape optimization tasks~\citep{li2023fourier,lin2025mGNO}. In electromagnetics and photonics, neural operators have supported free-form and optical device design~\citep{Augenstein2023NeuralOperatorSurrogate,duruisseaux2025photonics,gu2022neurolight,zhu2024pace}, and their integration with adjoint frameworks has enabled wavefront shaping in tunable metasurfaces~\citep{Kang2024AdjointFNO}. FNOs have also been used for control to steer quantum systems toward desired states~\citep{pipi2025inversefno,pipi2026}, and for the design of semiconducting quantum-dot devices~\citep{rychetsky2026}.

\vspace{1mm}

Collectively, these applications demonstrate that neural operators offer differentiable, high-fidelity surrogates that reduce design cycles and expand accessible design spaces.

\clearpage 
\section{Hyperparameter Tuning} \label{sec: Hyperparameter Tuning}

\subsection{General Machine Learning Practices}  \label{sec: General Machine Learning Practices}

The fundamental principles and best practices of machine learning apply to neural operators just as they do to other deep learning architectures. Effective training relies on careful tuning of hyperparameters, appropriate optimization strategies, consistent data preprocessing, regularization, and thorough model validation. These foundational practices ensure that neural operators achieve their full potential rather than being limited by poor experimental setup. We summarize briefly some essential practices that should be adopted when developing and evaluating neural operators, and refer the reader to \citep{Goodfellow-et-al-2016} for a comprehensive overview of general deep learning methodology.

\paragraph{Hyperparameter Tuning.} It is imperative to tune the model and training hyperparameters. The default model and training configurations, or those specified in the original papers, are rarely optimal for a new dataset or problem. The most influential model and training hyperparameters should be explored systematically within reasonable ranges, for instance using tailored grid searches. Random search or adaptive sampling methods can also help cover high-dimensional spaces of hyperparameter values efficiently. In general, model performance should not be reported based on default configurations but on the best performance obtained after a reasonable and explicitly specified amount of tuning.

\paragraph{Model Assessment.}  Evaluating a model often requires more than reporting a single error metric. A robust assessment involves quantitative, qualitative, and statistical analyses of performance across diverse conditions. Furthermore, evaluation should extend beyond accuracy to include computational cost both in terms of computational time and memory usage, and stability, ensuring that reported results reflect genuine improvements rather than isolated or unstable outcomes.

\paragraph{Learning Rate Schedule.} The optimizer and its learning rate schedule have a strong impact on both convergence and generalization, as they govern the stability and speed of optimization. A well-designed scheduler is essential for achieving stable and efficient training and can be either predefined or adaptive. Predefined schedulers such as \texttt{StepLR} or \texttt{ExponentialLR} follow a fixed rule for decreasing the learning rate, while cosine annealing introduces smooth periodic decays that can help the model escape shallow minima. Adaptive schedulers, such as \texttt{ReduceLROnPlateau}, adjust the learning rate dynamically based on training progress when a monitored metric stops improving. The appropriate choice depends on the optimization landscape, model sensitivity, and the desired balance between stability and exploration.

\paragraph{Batch size.} Batch size balances gradient smoothness and generalization. Larger batches offer stable optimization and faster throughput but may oversmooth gradients and harm generalization. Smaller batches add helpful stochasticity yet can slow or destabilize training. It can be helpful to experiment with different batch sizes, and in the process, adjusting the learning rate schedule and regularization might be needed.

\paragraph{Data Normalization and Preprocessing.}  Proper normalization is essential for stable optimization. Input and output quantities should be rescaled to comparable magnitudes using statistics computed on the training set. Standardization by mean and variance is common, but domain-aware normalization based on physical scales or boundary conditions can further improve performance. Within the model, normalization layers may improve stability and training speed. 

\paragraph{Overfitting and Underfitting.}  Persistent high errors on both training and validation sets often indicate underfitting, which can stem from insufficient model expressivity, input functions that are not informative enough, inadequate training time, or overly strong regularization. A widening gap between training and validation errors suggests overfitting and calls for additional data, smaller architectures, or stronger regularization. Beyond scalar metrics, visual comparisons between predicted and reference fields can provide important qualitative insight into model behavior and highlight specific failure modes.

\paragraph{Regularization.} Regularization helps prevent overfitting and encourages smoother learned operators. Weight decay and dropout remain reliable tools for managing model complexity, while early stopping based on validation loss adds protection when data are limited. More advanced methods, including spectral normalization or consistency penalties, can further constrain learning in high-dimensional spaces. Each regularization choice should be tuned and justified rather than applied automatically.

\paragraph{Profiling and Numerical Stability.}  Profiling tools should be used to identify bottlenecks in data loading, memory use, and GPU utilization. Debugging should include checking gradient magnitudes, loss scaling, and consistency of output ranges. Gradient clipping and mixed-precision training can improve stability without reducing accuracy. Before running large experiments, it is useful to verify convergence on a small subset of data to ensure the pipeline behaves as expected.

\paragraph{Experiment Design and Tracking.} Reproducibility requires structured experiment management. Every experiment should record hyperparameters, data splits, random seeds, and software versions. Training curves and intermediate checkpoints should be logged using experiment tracking tools such as \texttt{Weights and Biases (WandB)}. These tools also provide features to visualize the effect of different hyperparameters on performance and to help compare and identify optimal configurations.

\hfill 

\subsection{Number of Fourier modes: \codebox{n\_modes}}  \label{sec: n_modes}

\vspace{2mm}

The number of Fourier modes, \codebox{n\_modes}, is one of the most influential hyperparameters in the FNO. It determines the spectral resolution of the spectral convolutions in the model and directly shapes the range of spatial or temporal scales that each layer can represent. Despite its importance, it is often tuned incorrectly. When \codebox{n\_modes} is set too small, the model focuses only on low-frequency components, limiting its expressivity and potentially preventing the model from learning the full dynamics of the system due to under-parameterization. Conversely, choosing an excessively large number of modes can lead the model to learn spurious high-frequency behavior that does not exist in the data, resulting in artifacts or instability. Selecting \codebox{n\_modes} thus requires balancing these two effects: it must be large enough to capture the essential dynamics while remaining within the Nyquist limit to avoid learning nonphysical or aliased features.

\begin{figure}[h]  
\centering   
\includegraphics[width=\textwidth]{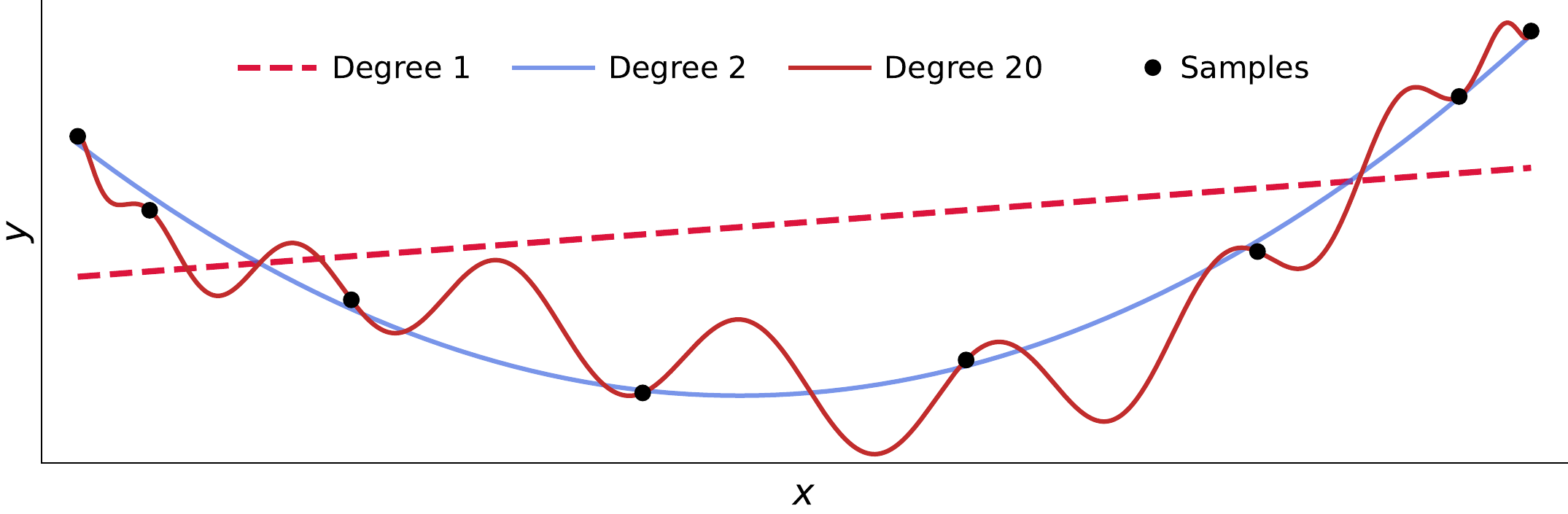}  \vspace{-8mm}
\caption{Polynomial fitting. A degree-$1$ polynomial underfits the sampled parabolic trend. A degree-$20$ polynomial interpolates the samples, but develops spurious oscillations between them. \label{fig: parabola} } \vspace{-12mm} 
\end{figure}

\paragraph{Analogy with Polynomial Fitting.}
A useful intuition comes from fitting a polynomial to a finite set of sampled points which lie approximately on a parabola, as illustrated in \Cref{fig: parabola}. Fitting a polynomial of degree 1 is too restrictive: it cannot reproduce the curvature and underfits the data. At the other extreme, a very high-degree polynomial can pass almost exactly through every sampled point, achieving a very small error on the observed data, while oscillating strongly between the samples. If this polynomial is then evaluated at new locations, its predictions can be very bad despite its excellent fit to the original points. The choice of \codebox{n\_modes} in an FNO involves a similar tradeoff. With too few Fourier modes, the model lacks the spectral resolution needed to represent important fine-scale structure. With too many modes, the model can make use of frequencies close to the largest ones that the grid can resolve. The nonlinearities in subsequent layers can then generate even higher frequencies, beyond the Nyquist limit of the training grid, which alias into lower frequencies, so the model may perform well on the training discretization while learning a representation that does not behave consistently as a continuous function. When evaluated on a finer grid, these higher frequencies become resolvable and can appear as spurious oscillations or other artifacts. The analogy therefore highlights an important principle: the goal is not simply to maximize model capacity, but to choose a level of spectral complexity that is supported by the resolution of the available data.

\clearpage

It should be noted that truncating the highest frequency modes does not necessarily prevent the model from capturing high-frequency details. Through the interaction of multiple spectral layers, nonlinearities, and residual connections, the FNO can reconstruct higher-frequency behavior from compositions of lower-frequency transformations, as discussed in \Cref{sec: FNO general def,sec: ChannelMLP,sec: skip connections}. In practice, restricting \codebox{n\_modes} to a physically meaningful range often yields smoother, more stable, and better-generalizing models without sacrificing the fidelity of small-scale dynamics.

The hyperparameter \codebox{n\_modes} controls how many Fourier coefficients are retained and acted upon within each \codebox{SpectralConv} layer. As described in \Cref{sec: SpectralConv}, each convolutional layer in the FNO transforms its input into the frequency domain, applies a complex-valued linear transformation to a subset of modes, and then reconstructs the output in physical space. The number of modes directly determines the spectral resolution of the operator, defining the range of spatial scales that can be explicitly represented and learned. In practice, \codebox{n\_modes} acts as a bandwidth parameter that regulates how much of the input signal's frequency content the layer can manipulate through learnable transformations.

Within this spectral mapping, the retained Fourier block is multiplied by complex-valued matrices that mix input and output channels at each frequency. Increasing \codebox{n\_modes} expands the number of active frequency blocks and proportionally raises the number of learnable parameters associated with spectral mixing, thereby enhancing the operator's ability to model fine-scale variations. However, this comes at a higher computational and memory cost, and may introduce instability if the model attempts to learn frequencies that are poorly resolved by the discretization. Importantly, \codebox{n\_modes} does not change the global receptive field, which remains a consequence of the Fourier transform itself, but rather specifies which frequency bands receive trainable attention. Smaller values constrain the model to coarse global structures, while larger values enable it to capture multiscale behavior and fine spatial detail within the limits of numerical resolution.\\

\paragraph{Conducting a Preliminary Spectral Analysis.}  A principled way to determine a good starting value for \codebox{n\_modes} is to examine the power spectra of the training data. By analyzing the power spectrum of representative samples, one can identify the smallest subset of low frequencies that collectively carry most of the power of the signal or target field. Once again, we emphasize that \codebox{n\_modes} must respect the Nyquist limit imposed by the grid resolution. \\

\paragraph{Anisotropy.} Many physical systems display anisotropic behavior, with characteristic correlation lengths that differ across spatial directions. In such cases, the choice of \codebox{n\_modes} should reflect the directional variability of the underlying dynamics. For example, in fluid flows that are elongated along one axis but exhibit rapid variations along another, assigning a larger number of modes to the direction of finer variation allows the operator to capture anisotropic structures more effectively while avoiding redundant capacity along smoother directions. When time is also treated spectrally, a separate temporal mode count can be introduced to control the balance between temporal expressivity and numerical stability, ensuring that the model captures relevant temporal frequencies without amplifying noise or stiffness. The \codebox{FNO} class implementation in \texttt{NeuralOperator 2.0.0} supports this flexibility by allowing \codebox{n\_modes} to be specified as a tuple, with each element indicating the number of modes retained along the corresponding spatial (or temporal) dimension, while making sure that it remains within the Nyquist limit for each dimension.  \\

\paragraph{Spectral Resolution and the Nyquist Constraint.} The selection of \codebox{n\_modes} must be mindful of the Nyquist limit imposed by the resolution, discussed in \Cref{sec: Nyquist}. The Nyquist--Shannon sampling theorem implies that if the discrete grid along dimension $m$ has $N_m$ points, then the maximum representable (non-aliased) frequency is $N_m/2$. Frequencies beyond this limit fold back into the lower band, corrupting the spectral representation. While \codebox{n\_modes} could be selected to be $N_m$ (corresponding to $N_m$//2 negative modes, the 0 mode, and $N_m$//2 or ($N_m$//2 - 1) positive modes) without surpassing the Nyquist limit within the \codebox{SpectralConv} layers, applying nonlinearities typically creates higher frequency content.

\begin{figure}[t]   \vspace{5mm}
\centering   
\includegraphics[width=\textwidth]{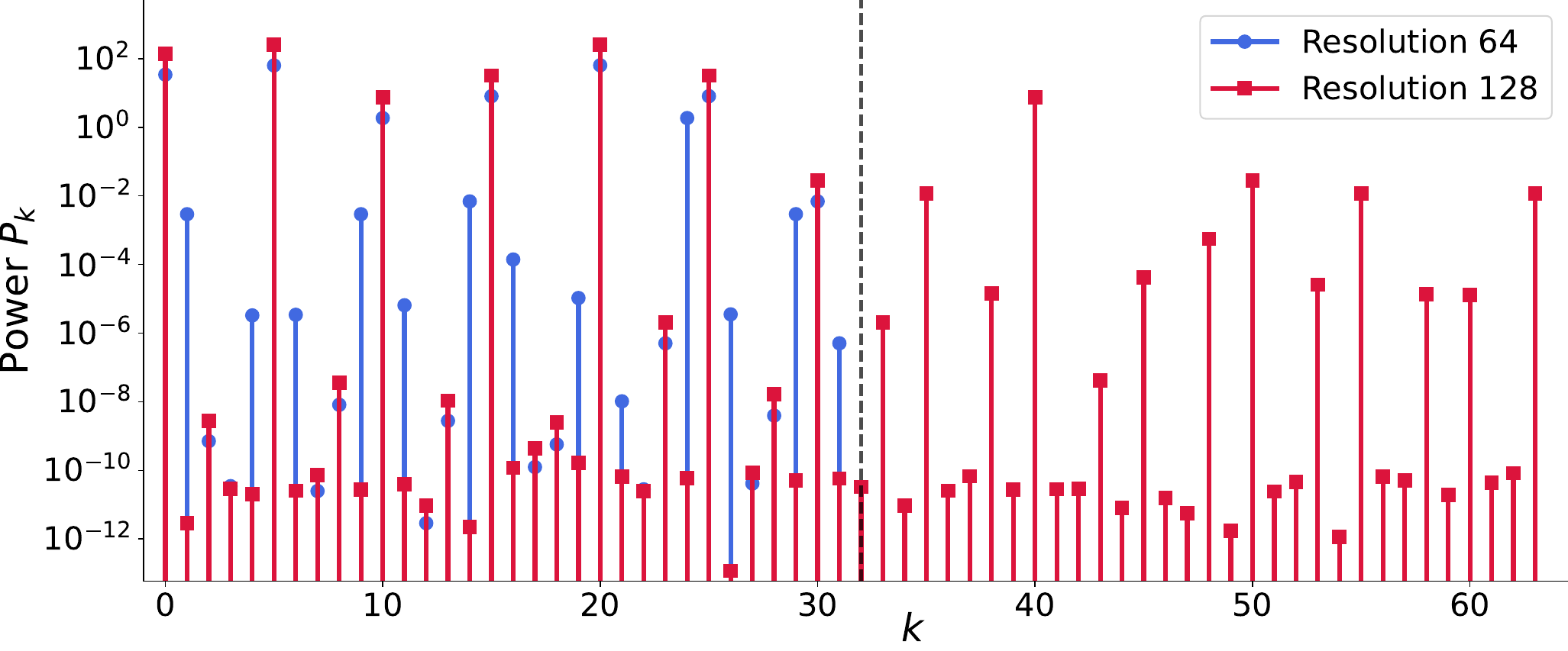}  \vspace{-5mm}
\caption{Power spectra of the signal $f(x) = 0.5\sin(5x) + 0.5\cos(20x)$ after being passed through a \texttt{GELU} nonlinearity, when sampled on grids of resolution $64$ (blue) and $128$ (red). The vertical line marks the Nyquist frequency $k = 32$ for the coarse grid. For resolution $128$ the nonlinear activation generates additional energy  in higher modes while for resolution $64$ this high frequency content is aliased back into lower modes which changes the apparent spectrum. \label{fig: nonlinearity test} } \vspace{8mm}
\end{figure}

\hfill 

As an example, consider the effect of applying the nonlinearity $\sigma(x) = x^2$. Take a discrete signal $\{x_n\}_{n=0}^{N-1}$ and define the squared sequence $\{y_n\}_{n=0}^{N-1}$  via $y_n = x_n^2$. Define their Fourier coefficients $\{X_k\}_{k=0}^{N-1}$ and $\{Y_k\}_{k=0}^{N-1}$, respectively. Squaring the formula for the inverse DFT representation~\eqref{eq: DFT IDFT} of $x_n$ gives
\begin{equation}
x_n^2
= \frac{1}{N} \sum_{p=0}^{N-1} \sum_{q=0}^{N-1}
X_p \, X_q \,
e^{2\pi i (p+q)n / N}.
\end{equation}
Substituting this into the formula for $Y_k$ and using the orthogonality relation
\begin{equation}
\sum_{n=0}^{N-1} e^{2\pi i r n / N}
=
\begin{cases}
N & \text{if $r$ is a multiple of $N$}, \\[1mm]
0 & \text{otherwise},
\end{cases}
\end{equation}
yields
\begin{equation}
Y_k
 =   \frac{1}{\sqrt{N}} \sum_{n=0}^{N-1} x_n^2 \, e^{-2\pi i kn / N}  =  \frac{1}{\sqrt{N} N}
\sum_{p=0}^{N-1} \sum_{q=0}^{N-1}
X_p \, X_q
\sum_{n=0}^{N-1} e^{2\pi i (p+q-k)n / N} = \!\!\!\!\!\!\! \mathop{\sum_{p=0}^{N-1} \sum_{q=0}^{N-1}}_{p+q-k \text{ is a multiple of } N} \!\!\!\!\!\!\!X_p \, X_q.
\end{equation}
Therefore, if the signal $x_n$ only activates the $K$ lower frequency modes ($X_k = 0$ if $|k| > K$), the squared signal $x_n^2$ activates an extended set of frequencies, up to $2K$. \\ 

As a result, even if the number of modes \codebox{n\_modes} is chosen to be $N$ so that the first \codebox{SpectralConv} acts on a completely resolved signal, the subsequent nonlinearity produces a signal whose wider bandwidth cannot be represented with only $N$ points, so when the FFT is applied again in the next \codebox{SpectralConv} the unresolved frequencies fold back into the discrete spectrum and lead to aliasing. This aliasing phenomenon is illustrated in \Cref{fig: nonlinearity test} for the \texttt{GELU} nonlinearity. 

As a consequence, if the FNO is trained at resolution $N$ with a number of modes \codebox{n\_modes} equal to $N$, the spectral convolutions are trained to operate on aliased signals produced by the nonlinearities. When the model is later evaluated at a higher resolution, the nonlinearity is now fully resolved in higher frequency modes, so the aliasing disappears and the FNO applies incorrect transformations to the lower frequency modes which are no longer contaminated by aliasing.

Therefore, when zero-shot super-resolution is a downstream task of interest, we highly recommend reducing the number of modes \codebox{n\_modes} to account for the effect of the nonlinearities, at least below $N/2$ when the training resolution is $N$.

Alternatively, we can insert upsampling layers before the nonlinearities so that the effect of the nonlinearities can be resolved properly with the extended resolution, and then use spectral downsampling (or equivalently low-pass spectral filtering) after the nonlinearities to recover the desired resolution. This procedure removes the higher frequency components generated by the nonlinearities instead of allowing them to alias and fold back into lower frequency bins, so the spectral convolution layers are trained only on the true low frequency content associated with the target resolution. As a result, these convolutions remain valid when the model is later evaluated on higher resolution grids. Conceptually related upsample-filter-downsample schemes have been employed in prior architectures that combine learned upsampling and downsampling, most notably in the alias free design of StyleGANs~\citep{Karras2019stylegan,Karras2020stylegan2} where it is used to suppress aliasing and control frequency content across resolutions.

One way to choose the upsampling factor is to estimate how strongly the nonlinearity broadens the spectrum of a typical input signal. This can be done by examining the polynomial expansion of the nonlinearity and assessing both the frequency support and the relative magnitude of the higher order terms  (as was done in the example above with \(\sigma(x) = x^2\)). Terms whose contribution remains above a chosen amplitude threshold indicate how far the spectrum can extend beyond the original band limit and therefore how large the intermediate Nyquist frequency must be. In practice, a suitable upsampling factor can also be determined empirically. One can apply the nonlinearity at different intermediate resolutions to representative inputs, compute the resulting Fourier spectra, and measure how much high-frequency energy aliases back into the low-frequency band of the target grid. The smallest upsampling factor for which this aliased energy stays below a prescribed tolerance then provides a principled choice for the resolution required to accurately resolve the effect of the nonlinearity.

\hfill 

\subsection{Number of Fourier Layers: \codebox{n\_layers}} \label{sec: number of layers}

\codebox{n\_layers} specifies the number of \codebox{FNOBlock} Fourier layers that compose the FNO. It serves as a primary determinant of the model's nonlinear expressivity, balancing representational power with computational efficiency and optimization stability. Increasing the number of layers deepens the model and introduces additional stages of nonlinear processing, enhancing the model's ability to approximate complex mappings between function spaces. From an operator-theoretic perspective, each block can be interpreted as one iteration of a nonlinear transformation acting on the input function, so stacking multiple layers allows the model to successively refine the solution representation.

A larger value of \codebox{n\_layers} increases the expressivity of the model by incorporating more operations and in particular more nonlinear components. Shallow models with one or two layers typically suffice for smooth or weakly nonlinear operators, whereas deeper configurations capture intricate, multi-scale, and highly nonlinear dependencies. However, as in other deep architectures, adding too many layers can lead to diminishing returns, as the improvement in accuracy saturates while the computational cost grows linearly with depth.

 Deeper FNOs can exhibit optimization challenges similar to those encountered in conventional deep networks, such as vanishing or exploding gradients. These difficulties are mitigated in practice by the residual skip connections included in every \codebox{FNOBlock}, which preserve stable gradient flow across layers and allow each block to learn residual refinements rather than complete transformations. This residual formulation promotes smoother training dynamics, faster convergence, and improved numerical stability, even in models with a large number of layers. We recommend starting with a value of \codebox{n\_layers} between 3 and 6, and then increasing it only if stronger nonlinearity seems to be required. 

\clearpage

\subsection{Hidden Channels: \codebox{hidden\_channels}} \label{sec: hidden channels}

\vspace{2mm}

The parameter \codebox{hidden\_channels} is one of the central FNO hyperparameters, controlling the dimensionality of the latent feature space in which the operator is learned. It defines the number of channels used internally within each Fourier layer (\codebox{FNOBlock}), and more precisely within every component of the \codebox{FNOBlock}, i.e. the \codebox{SpectralConv}, the \codebox{ChannelMLP}, and the lifting and projection layers. In essence, \codebox{hidden\_channels} determines the expressive capacity of the model: increasing it allows the model to represent more complex relationships between input variables and capture richer, multi-scale interactions, whereas smaller values yield a more compact model with reduced computational and memory cost.

Within each Fourier layer, \codebox{hidden\_channels} sets the number of feature channels over which both the spectral convolutions and the channel-wise MLPs operate. The spectral convolutions operate in the Fourier domain, where they learn complex-valued weight matrices that mix information across channels within each retained frequency mode, while the \codebox{ChannelMLP} acts in the physical domain, applying nonlinear transformations across the same channel dimension independently at every spatiotemporal location.

A useful analogy can be drawn with convolutional neural networks (CNNs) where an RGB image initially has three RGB colors input channels, which are then transformed into a hierarchy of feature maps with progressively higher numbers of channels, such as 64, 128, 256. This expansion allows the network to disentangle and represent increasingly abstract and complex features before eventually reducing them back to the three RGB channels for output. Similarly, in the \codebox{FNO}, the parameter \codebox{hidden\_channels} specifies the number of intermediate feature channels that the operator uses to process information. A higher number of hidden channels equips the model with a richer basis for representing physical quantities and nonlinear couplings, enabling it to capture both global smooth structures and fine-scale variations within the data.

Overall, choosing the value of \codebox{hidden\_channels} involves balancing expressivity and efficiency. Larger values improve the model's capacity to learn complex operator mappings and represent multi-frequency behavior, but they also rapidly increase the computational cost with the number of channels and can lead to over-parametrized models. In practice, we recommend starting with lower values of \codebox{hidden\_channels} around 16-32, especially when working in higher dimensions, and then increasing it progressively if higher expressivity is required. 

\hfill

\subsection{Other Hyperparameters} \label{sec: Tuning other hyperparameters}

\vspace{1mm}

We list here a few additional functionalities made available through arguments of the \codebox{FNO} class in \texttt{NeuralOperator 2.0.0}, and discuss how to tune them.

\vspace{2.5mm}

\paragraph{Complex-Valued Support.} \texttt{NeuralOperator 2.0.0} incorporates native support for complex-valued functions in its \codebox{FNO} class through the boolean argument \codebox{complex\_data}. When \codebox{complex\_data} is set to \texttt{True}, the model internally wraps the lifting and projection layers, as well as all intermediate FNO blocks, with complex-valued module variants implemented via a \codebox{ComplexValued} wrapper class. This design ensures that every transformation is performed directly in the space of complex numbers. Consequently, the FNO can accurately represent mappings between complex-valued function spaces. This architectural feature makes the FNO a versatile operator-learning framework applicable to both real-valued and complex-valued problems without requiring manual handling of real and imaginary components in user code.

\vspace{2.5mm}

\paragraph{Normalization.} The \codebox{norm} parameter in the \codebox{FNO} class of \texttt{NeuralOperator 2.0.0} provides a flexible mechanism to incorporate a variety of normalization strategies. Specifically, the parameter accepts string options such as \texttt{``ada\_in''}, \texttt{``group\_norm''}, \texttt{``instance\_norm''}, or \texttt{None}, and is propagated to all FNO blocks and channel mixing layers. If a normalization type is specified, the corresponding normalization module is inserted after spectral convolution operations and before nonlinear activations in the FNO blocks. This can facilitate both the stabilization of training dynamics and the enhancement of generalization capabilities in operator learning tasks. A tutorial displaying the effect of these different normalization layers is available at \href{https://neuraloperator.github.io/dev/auto\_examples/layers/plot\_normalization\_layers.html}{neuraloperator.github.io/dev/auto\_examples/layers/plot\_normalization\_layers.html}.

\vspace{2mm}

\paragraph{Nonlinearity.} \texttt{NeuralOperator 2.0.0} provides flexible control over the nonlinear activation functions used within its \codebox{FNO} architecture through the \codebox{non\_linearity} argument. By default, the model employs the Gaussian Error Linear Unit (\texttt{GELU}) activation, which has been shown to perform well in smooth operator learning tasks. However, users can supply any valid \texttt{torch.nn.Module}, such as \texttt{ReLU}, \texttt{LeakyReLU}, or a custom nonlinearity, when initializing the model. This configurability allows practitioners to tailor the activation behavior to the characteristics of specific datasets or physical systems. The choice of nonlinearity can significantly influence the model's expressivity, gradient flow, and optimization stability, thereby affecting convergence speed and generalization. 

\hfill

\paragraph{Precision.} \texttt{NeuralOperator 2.0.0} introduces explicit control over numerical precision within its \codebox{FNOBlock} modules through the \codebox{fno\_block\_precision} argument. This parameter accepts values such as \texttt{"full"}, \texttt{"half"}, or \texttt{"mixed"}, determining whether computations are performed in full precision (\texttt{float32}), reduced precision (\texttt{float16}), or a combination of both. 

\vspace{1mm}

Using lower precision can substantially accelerate training and inference on modern hardware that supports mixed-precision arithmetic while also reducing memory consumption. However, reduced precision may lead to numerical instability in deep or highly expressive operator models. To address this, the FNO includes an optional stabilization mechanism controlled by the \codebox{stabilizer} argument, which currently supports a \texttt{"tanh"} normalization function. When enabled, this transformation constrains activation magnitudes in the spectral domain, preventing divergence or overflow. 

\hfill 

\paragraph{Separable Spectral Convolutions.} The \codebox{separable} option in the \codebox{FNO} class of \texttt{NeuralOperator 2.0.0} determines whether spectral convolution layers mix channels in Fourier space. 

\vspace{1mm}

By default, \codebox{separable} is \texttt{False}, and each spectral convolution layer leverages a complex weight tensor of shape \texttt{(in\_channels, out\_channels, *modes)}. This implements a frequency-specific linear map from input to output channels, allowing channel mixing that can vary across Fourier modes. 

\vspace{1mm}

In contrast, when the \codebox{separable} option is set to \texttt{True}, the number of \texttt{in\_channels} must be the same as the number of \texttt{out\_channels} and each spectral convolution layer leverages a complex weight tensor of shape \texttt{(channels, *modes)} which does not mix information across channels. Instead of learning a full channel-to-channel mixing matrix at each Fourier mode, it learns a single complex gain for each channel and retained frequency. Each spectral convolution layer takes the Fourier transform of each channel, scales every retained Fourier coefficient in that channel by its own learned complex number, and then inverse Fourier transforms back to physical space. In this case, each channel is modulated independently in the frequency domain, with no cross-channel coupling inside the spectral convolution.

\vspace{2mm}

Full channel mixing is typically much more expressive and preferred because it enables frequency-dependent fusion of features across channels, which can be critical when channels represent interacting fields or when useful representations require combining channels in the spectral domain. However, this added flexibility comes with higher parameter count and computational cost. The separable spectral convolution replaces these mode-wise matrices with a single complex coefficient per channel per retained mode, yielding substantial savings in memory and compute, but it eliminates cross-channel interactions within the spectral convolution itself. Separability can be most appropriate in applications where channels must remain largely independent, or when channel mixing is intentionally handled elsewhere.

\clearpage

\subsection{Hyperparameter Explorations using Small ``Overfitting'' Studies} \label{sec: Overfitting studies}

\vspace{4mm}

Conducting small ``overfitting'' studies, where the model is trained to reproduce a small subset of the dataset (e.g. 20-100 samples), can rapidly provide insightful information about the effect of hyperparameters before undertaking full-scale training. The goal is not to achieve generalization to unseen examples, but to determine whether the current configuration is appropriately designed for the intended operator learning task. Note that the reduced dataset yields far fewer gradient updates per epoch than the full dataset, so these studies typically require substantially more epochs (possibly on the order of tens of thousands) along with corresponding retuning of the learning rate scheduler.

\vspace{2mm}

These studies can be used to determine whether a given model has sufficient expressive power to reproduce representative examples from the training distribution. If the model fails to fit such a small dataset, it is unlikely to perform well on the full dataset and generalize to unseen examples. Conversely, if it succeeds easily, one can then analyze how architectural choices influence convergence, the smoothness of the predictions, and the physical plausibility of the outputs. For instance, with FNOs, varying the number of modes, the number of hidden channels, the number of layers, and the choices of nonlinearity and skip connections can reveal what helps the model capture the relevant spatial and temporal structures in the data. 

\vspace{2mm}

These studies can also provide a low-cost setting for testing how data preprocessing strategies (e.g. normalization, embeddings, subsampling, truncation, padding, masking, domain rescaling) affect training behavior. They can also help select better choices of loss functions which are tailored to capturing the features of highest interest. Training until the error is driven as low as possible across a range of loss functions and normalization strategies provides a fast, interpretable probe of what good performance looks like under each metric, including which spatiotemporal features are captured or not. Visual inspection of predicted versus reference solutions is particularly valuable here. The resulting near-minimum training errors also provide an empirical upper bound on what the best case scenario would look like on the full dataset for each configuration, offering a rapid way to identify promising losses and preprocessing choices before scaling to the full dataset.

\vspace{2mm}

Finally, progressively increasing the size of the small training set while observing how hyperparameters must be adjusted can be insightful. By examining how parameters such as learning rate, batch size, regularization strength, or model width need to change with the dataset size, these studies help identify smaller potential ranges of training configurations that are likely to perform well on the full dataset. This process also offers an early estimate of the overall training cost. Furthermore, as the small dataset becomes sufficiently large, such experiments can reveal whether the information contained in the input functions truly has a high predictive power over the outputs, or whether the model has merely been memorizing individual samples, a behavior that becomes increasingly difficult as the dataset size grows.

\clearpage

\section{Advanced Training Strategies} \label{sec: Advanced Training Strategies}

\subsection{Data Losses for Training} \label{sec: data losses}

\vspace{1mm}

When training neural operators, data losses provide a fundamental measure of how closely the predicted output function $\tilde{g}$ matches the ground truth function $g$. Since neural operators act on function spaces rather than finite-dimensional vectors, it is crucial to define losses that remain consistent under different discretizations of the domain. This ensures that the optimization objective reflects genuine functional discrepancies and not artifacts of the chosen resolution or sampling scheme. 

\paragraph{Absolute Losses.} The most widely used data loss is the (squared) \mydef{absolute $L^2$ loss}, which measures the energy of the pointwise error between $\tilde{g}$ and $g$:
\begin{equation}
\label{eq:absL2-cont}
  \mathcal{L}_{L^2}^{\mathrm{abs}}(\tilde{g},g) = \| \tilde{g} - g \|_{L^2}^2
\ = \  \int_{D_g} | \tilde{g}(y) - g(y) |^2 \, \mathrm{d}y .
\end{equation}
The absolute $L^2$ loss can be generalized to the family of \mydef{absolute $L^p$ losses} for $1 \le p < \infty$:
\begin{equation}
\label{eq:absLp-cont}
\mathcal{L}_{L^p}^{\mathrm{abs}}(\tilde{g},g)
= \| \tilde{g}- g \|_{L^p}^p
= \int_{D_g} | \tilde{g}(y) - g(y) |^p \, \mathrm{d}y.
\end{equation} 

The parameter $p$ modulates sensitivity to outliers: $p < 2$ yields robustness to isolated large deviations, while $p > 2$ emphasizes regions of large error. In smooth regression problems, $p = 2$ is typically preferred for its differentiability and numerical stability, whereas $p = 1$ or mixed $L^1$--$L^2$ losses can be useful when the data contain discontinuities or sparse structures.

For applications where the derivatives of $g$ carry important information, Sobolev losses provide a more informative learning signal. The most common example is the \mydef{absolute $H^1$ loss}, which penalizes differences in both function values and their gradients in the $L^2$ sense:
\begin{equation}
\label{eq:H1-cont}
\mathcal{L}_{H^1}^{\mathrm{abs}}(\tilde{g},g)
= \| \tilde{g} - g \|_{H^1}^2
= \int_{D_g} \Big( | \tilde{g}(y) - g(y) |^2 + | \nabla \tilde{g}(y) - \nabla g(y) |^2 \Big) \mathrm{d}y.
\end{equation}
The second term encourages the model to match the local variations and smoothness of the target function, which is particularly important when the data exhibit high-frequency or oscillatory behavior. The derivative terms can be evaluated using finite differences, spectral differentiation, or automatic differentiation depending on the discretization and model implementation. However, the effectiveness of Sobolev losses depends on the quality of derivative estimates. If ground truth derivatives are noisy or poorly approximated, they may introduce instability into the training process. 

Beyond the $L^p$ and Sobolev losses, several specialized losses can be leveraged to capture different aspects of operator learning. For instance, \mydef{weighted $L^p$ losses} introduce spatial weights $w(y)$ to emphasize certain regions of higher interest, via
    \begin{equation}
    \mathcal{L}_{L^p_{w}}^{\mathrm{abs}}(\tilde{g},g) = \int_{D_g} w(y) | \tilde{g}(y) - g(y) |^p \, \mathrm{d}y.
    \end{equation}
    and \mydef{spectral losses} measure differences in the spectral domain, useful for enforcing frequency fidelity, for instance as
    \begin{equation}
    \mathcal{L}_{\mathrm{Fourier}}(\tilde{g},g) = \sum_{k} | \mathcal{F}(\tilde{g})(k) - \mathcal{F}(g)(k) |^2.
    \end{equation}

\paragraph{Relative Data Losses.} While absolute losses penalize absolute discrepancies, they may be sensitive to the scale of $g$. Relative versions of these losses can be defined by normalizing with the corresponding norm of the ground truth function, ensuring scale invariance. For instance, the \mydef{relative $L^p$ loss} is obtained via
\begin{equation}
\mathcal{L}_{L^p}^{\mathrm{rel}}(\tilde{g},g)
= \frac{ \| \tilde{g} - g \|_{L^p}^p }{ \| g \|_{L^p}^p + \varepsilon } 
\end{equation}
where $\varepsilon$ is a small constant added for numerical stability. Relative losses are particularly useful when the target function varies significantly in magnitude across samples or when absolute scaling is uninformative.

\paragraph{Discretization.} In practice, the functions are discretized so the integrals in the function space losses introduced above need to be approximated using evaluations of the integrands on discretization points $\{y_j\}_{j=1}^m$. A key requirement for meaningful loss evaluation is a principled discretization of these integrals. Because neural operators are defined on function spaces, the discrete losses should approximate and converge to their continuous counterparts as the resolution increases. Using nonuniform or inconsistent quadrature weights can bias the optimization and hinder convergence to the correct functional solution. We can associate appropriate  quadrature weights $\{\Delta_j\}_{j=1}^m$ to the discretization points $\{y_j\}_{j=1}^m$, ensuring that $\sum_j \Delta_j$ approximates the measure of $D_g$ and that the quadrature rule is consistent across resolutions. This helps preserve the resolution-invariance of the loss function. 

For instance, the absolute $L^2$ loss can be approximated via
\begin{equation}
 \mathcal{L}_{L^2}^{\mathrm{abs}}(\tilde{g},g) 
\ = \  \int_{D_g} | \tilde{g}(y) - g(y) |^2 \, \mathrm{d}y \ \approx \  \sum_{j=1}^m | \tilde{g}(y_j) - g(y_j) |^2 \, \Delta_j.
\end{equation}
When using FNOs on uniform grids with $\Delta_j = \Delta$, the expression is proportional to the Mean Squared Error (MSE), which is commonly employed in neural network regression tasks. \\

\paragraph{Implementation.} The $L^p$ and $H^1$ losses are implemented in \texttt{NeuralOperator 2.0.0} as the classes \codebox{LpLoss} and \codebox{H1Loss} in \codebox{neuralop/losses/data\_losses.py}. After creating a \texttt{loss\_fn} loss function by instantiating from these classes, the corresponding absolute and relative losses can be evaluated using \texttt{loss\_fn.abs(y\_tilde, y)} and \texttt{loss\_fn.rel(y\_tilde, y)}. The \codebox{LpLoss} and \codebox{H1Loss} can also take as an argument the \codebox{measure} of the domain (either as a single scalar common to every dim, or as a list of scalars for the different dimensions), and \codebox{reduction} specifying whether to \texttt{``sum"} or take the \texttt{``mean"} across the batch and channel dimensions. The forward of the \codebox{abs} and \codebox{rel} methods of these classes can also take an argument \codebox{quadrature} specifying the quadrature weights for the integral.

\hfill 

\subsection{Autoregressive Rollouts} \label{sec: Autoregressive}

\vspace{2mm}

\subsubsection{Single-Step Autoregressive Rollouts} \label{sec: Single-Step Autoregressive}

\vspace{2mm}

\paragraph{Formulation.} In spatiotemporal forecasting, FNOs have emerged as a powerful paradigm for learning mappings between function spaces that describe evolving physical or dynamical systems. FNOs can be trained to predict system evolution over a fixed temporal increment $\Delta t$ by learning an operator $\mathcal{G}_{\theta}$ approximating the mapping from the solution $u$ at time $t$ to the solution $u$ at time $t+\Delta t$, i.e.
\begin{equation}
u(t+\Delta t, \cdot) \ \approx  \ \mathcal{G}_{\theta}\bigl(u(t, \cdot)\bigr) \coloneqq \hat{u}(t+\Delta t, \cdot)
\end{equation}
During inference, this operator is applied recursively to its own predictions to extend the forecast over longer horizons, a process known as an \mydef{autoregressive rollout}. In this mode, starting from time $t=0$ and  $\tilde u(0, \cdot) = u(0, \cdot) $, the predicted future states are generated iteratively as
\begin{equation}
\hat{u}((n+1)\Delta t, \cdot) = \mathcal{G}_{\theta}\bigl(\hat{u}(n\Delta t, \cdot)\bigr), \quad n=0,1,2,\ldots
\end{equation}
This approach enables the model to capture long-range dynamics from short-term supervised training data.

In some settings, it can be advantageous to use the residual or step 
\begin{equation} \Delta u(t,\cdot) = u(t+\Delta t, \cdot) - u(t, \cdot) \end{equation} 
as the learning target, and then roll out the dynamics autoregressively via 
\begin{equation}
\hat{u}((n+1)\Delta t, \cdot) = \hat{u}(n\Delta t, \cdot) + \mathcal{G}_{\theta}\bigl(\hat{u}(n\Delta t, \cdot)\bigr), \quad n=0,1,2,\ldots
\end{equation}

\paragraph{Advantages.}  The autoregressive formulation reduces the dimensionality of the forecasting problem by removing the temporal dimension. Instead of learning a complex mapping from an initial spatial state $u(t_0,\cdot)$ to the spatiotemporal solution $u(t,\cdot)$  on the entire time  interval $t\in[0,T]$,  it allows the model to learn only the mapping from one spatial state to the next. In addition, the model only needs to learn how the current spatial state $u(t,\cdot)$ evolves after a small time $\Delta t$, as opposed to learning how an initial spatial state $u(t_0,\cdot)$ affects the evolution over a much longer time horizon $T$. The operator learned in this setting is closer to the identity mapping because the system typically evolves only slightly between two adjacent time steps. Learning such a near-identity transformation is often more stable and easier to optimize than learning a long-horizon mapping that must model large cumulative changes. This simplification lowers the functional complexity of the learning task and allows the model to focus on short-term dynamics that are often smoother and easier to capture. 

Another benefit is the large amount of training data that can be generated. Every trajectory from the dataset provides many consecutive one-step training pairs $(u(t), u(t+\Delta t))$ rather than a single long-horizon input and output sample. This makes the autoregressive method highly data efficient since one simulation can produce a large number of training samples that cover a wide variety of temporal transitions.

The autoregressive structure also aligns well with the nature of physical dynamical systems, which evolve continuously through small incremental updates. By training on these short temporal increments, the model learns accurate local dynamics that can be repeatedly composed to generate complex long-term behavior. This formulation also improves generalization since the training objective is based on short-term consistency rather than long-horizon outcomes that depend heavily on specific initial conditions. Note as well that the models do not need to satisfy the typical numerical stability conditions of numerical integrators, since those typically arise from the requirement that the steps be small enough to approximate temporal derivatives accurately enough. As a result, they can be used to learn the evolution over larger time steps $\Delta t$ than their numerical solver counterpart, thereby further accelerating the simulation process. \\

\paragraph{Challenges.} Despite its simple and intuitive formulation, autoregressive rollout introduces challenges that can compromise stability and accuracy over long time horizons. A central difficulty is the mismatch between training and inference distributions. During training, models are typically conditioned on ground truth inputs drawn from the data distribution. At inference, the operator must recursively operate on its own predictions, which induces a shift in the input distribution and leads to compounding errors across steps. This mechanism drives error growth in high dimensional and chaotic dynamical systems where small local discrepancies can cascade into large global instabilities. A further challenge arises from cumulative representational errors that distort spatial or temporal modes of the predicted fields. Neural operators may bias toward dominant low-frequency components and fail to maintain small amplitude or fine scale structure during extended rollouts. Although these discrepancies are minor at short horizons, they accumulate in time and degrade fidelity of the predicted dynamics. Similar behavior is familiar in numerical analysis, where aliasing and truncation of high-frequency modes cause loss of physical realism and eventual instability~\citep{brandstetter2023messagepassingneuralpde}.

Several strategies mitigate these shortcomings. Training procedures that emulate the test time regime have been proposed, including unrolled objectives and pushforward style training where gradients propagate only through later steps of the rollout~\citep{brandstetter2023messagepassingneuralpde}. Introducing stochasticity during training through denoising or noise perturbed objectives can improve robustness to the imperfect inputs encountered at inference and stabilize long trajectories~\citep{hao2024dpot}. Another line of work enforces dissipative or contractive behavior in order to prevent spurious growth of energy and to restore trajectories toward the attractor in chaotic regimes, which yields gains in long term stability~\citep{li2022learning}. A complementary approach integrates the autoregressive mechanism directly into training by optimizing losses over short rollouts. Since backpropagation through many steps is memory intensive, practical implementations rely on gradient checkpointing or truncated objectives that limit gradient flow to the final steps, thereby aligning training with inference while keeping resource usage manageable~\citep{brandstetter2023messagepassingneuralpde,lam2023learning}. Finally, enforcing geometric, conservation laws and other physics constraints can also substantially extend stable rollout horizons, as demonstrated for dynamics on the sphere where spherical harmonics avoid artifacts of planar Fourier transforms and enable year-scale stable forecasts under autoregression~\citep{bonev2023spherical}.

\citep{li2022learning} also recommend that autoregressive rollouts for chaotic systems should be evaluated not solely on trajectory accuracy but also on their ability to reproduce correct statistical behavior over long horizons. In chaotic regimes, the divergence of individual trajectories is unavoidable beyond a short predictability window, yet the invariant measure or ensemble statistics can remain stable if the learned operator captures the correct underlying dynamics. Even with small step errors, long-term rollouts may faithfully recover attractor geometry and power spectra, thereby yielding physically meaningful forecasts at the statistical level. This reinforces that model stability in the autoregressive setting is not equivalent to pointwise error control but rather to maintaining consistency of the learned flow with the true dynamical measure.

\vspace{4mm}

\subsubsection{The Recurrent Neural Operator (RNO)} \label{sec: RNO}

\vspace{2mm}

The \mydef{Recurrent Neural Operator (RNO)}~\citep{liu2023tipping} extends FNOs to systems with long-term temporal dependencies by introducing recurrence and memory directly on function spaces. Conceptually, the RNO builds on the idea of sequence-to-sequence modeling that is widely used in time series analysis, but generalizes it to the function space setting. The RNO allows to perform autoregressive rollouts not merely by mapping one state to the next, but by learning mappings from multiple preceding time intervals to one or several future time intervals. Alternatively,  it can also be used to predict the solution at future time step(s) from the solution at multiple preceding time steps. This allows it to leverage temporal context and capture the evolution of complex, history-dependent dynamics across time.

Its construction follows the logic of recurrent neural networks (RNNs)~\citep{elman1990finding}, particularly the gated recurrent unit (GRU)~\citep{cho2014learning}, but generalizes the computation from finite-dimensional vectors to infinite-dimensional functions. At each time step $\tau$, the RNO cell receives as input a function $u_{\tau}(x,t)$, representing the system state over a short temporal window, together with a hidden state function $h_{\tau}(x,t)$ that stores the latent memory of past evolution. The update of this hidden state is governed by a system of gates analogous to those in a GRU, but implemented through FNO layers that act as learned integral operators rather than simple matrix multiplications~\citep{liu2023tipping}. 

\begin{figure}[t]  \vspace{-2mm}
\centering   
\includegraphics[width=\textwidth]{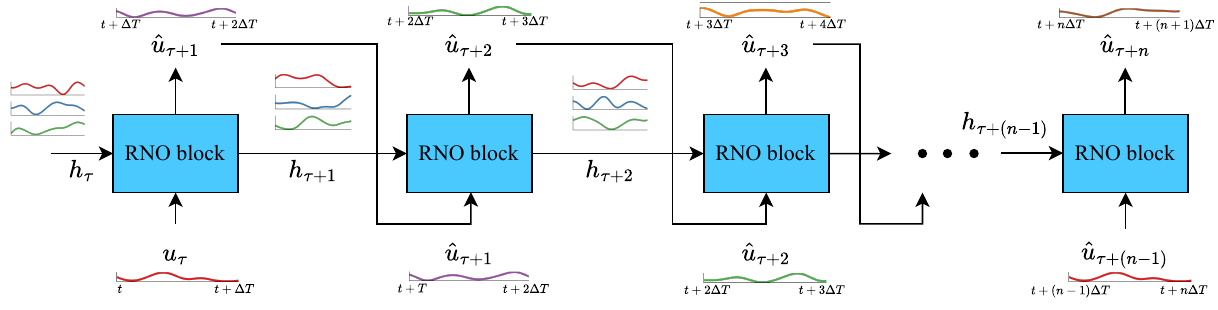}  \vspace{-6mm}
\caption{The Recurrent Neural Operator (RNO) architecture (figure extracted from~\citep{liu2023tipping}).   
\label{fig: RNO} } \vspace{6mm}
\end{figure}

The RNO maintains the discretization-invariant nature of the FNO while introducing a principled mechanism for temporal recurrence, enabling it to perform stable long-horizon autoregressive rollouts that incorporate information from multiple past states and predict function-valued trajectories far into the future. By introducing a latent functional state, the RNO can model non-stationary dynamics and systems with long memory that cannot be adequately represented by Markovian operators. This functional recurrence enables the model to capture transitions, delayed effects, and gradual regime shifts over extended time horizons without requiring post-event data or explicit numerical integration. In essence, the RNO bridges neural operator learning and recurrent modeling, providing a unified framework for learning dynamical systems that evolve over both continuous spatial domains and long temporal scales.

The RNO architecture will be made available in a future release of \texttt{NeuralOperator} through the \codebox{RNO} class in \codebox{neuralop/models/rno.py}.

\hfill 

\subsection{Multi-Resolution Training and the Incremental FNO (iFNO)}
\label{sec: ifno}

In many scientific and engineering applications, data are available at multiple spatiotemporal resolutions. For instance, coarse simulations can be used for rapid exploration, while fine-grained simulations capture small-scale phenomena with higher fidelity. Leveraging this diversity across resolutions can substantially improve both data efficiency and model robustness.

\vspace{2mm}

\subsubsection{Multi-Resolution Training}  

\vspace{2mm}

\mydef{Multi-resolution training} provides a principled framework to exploit multi-resolution datasets, by training on coarse-resolution data first and progressively introducing finer-resolution examples. Coarse data capture smooth, large-scale behavior and are computationally inexpensive, while higher-resolution data contain increasingly complex, high-frequency structures. This staged progression allows the model to develop representations hierarchically, from global to local, while maintaining numerical stability and training efficiency.

Even if data samples are only available at high resolution, they can be downsampled (see \Cref{sec: downsampling} regarding downsampling strategies) to generate additional coarse-resolution training examples. These synthetic low-resolution samples augment the available data spectrum and enrich the learning signal. Such multi-scale augmentation helps the model build consistent representations across resolutions and encourages scale-invariant generalization. This approach is particularly useful in scientific domains where high-fidelity simulations are expensive to produce but can be reused effectively through structured resampling.

Multi-resolution training is especially natural with FNOs, where the input resolution directly determines the range of frequencies the model can represent. Beginning with coarse inputs implicitly restricts learning to low-frequency structures through the Nyquist--Shannon Theorem (see \Cref{sec: Nyquist}), while gradually increasing resolution introduces higher-frequency information in a controlled manner. By design, the FNO supports inputs and outputs defined on different resolutions. Consequently, multi-resolution training and inference can be implemented directly within the standard FNO pipeline and implementation in \texttt{NeuralOperator 2.0.0}, without any architectural modification or additional interpolation layers.

\hfill 

 \subsubsection{The Incremental FNO (iFNO)}

 \vspace{2mm}
 
 The \mydef{Incremental Fourier Neural Operator (iFNO)}~\citep{George2022Incremental} couples multi-resolution training with an adaptive expansion of the FNO's spectral capacity. Instead of fixing the number of Fourier modes and the data resolution throughout training, iFNO increases both progressively as learning advances. This dynamic evolution allows the model to align its representational complexity with the richness of the data, providing a flexible mechanism for efficient and stable operator learning.

Initially, iFNO is trained with low-resolution data and a small number of active Fourier modes. As optimization proceeds, two complementary mechanisms govern its evolution.

\vspace{1mm}

\paragraph{Incremental Frequency Expansion.} The first mechanism incrementally increases the number of active Fourier modes \(K\) within the spectral convolution layers, which determines the effective frequency bandwidth of the spectral convolution. Rather than being fixed, \(K\) is expanded adaptively according to the learning dynamics. Two adaptive options have been proposed to determine when to add more modes, and allow the model to expand gracefully, maintaining continuity in learned representations as new modes become active:

\begin{itemize}
\item Based on the training loss. If the loss stagnates over several epochs, indicating that the current spectral representation is saturated, higher-frequency modes are introduced to enrich the model's expressiveness. 

    \item  Based on the proportion of the total spectral energy represented by the current active modes. Let \(P_k = \sum_{i,j} |R_{k,i,j}|^2\) denote the energy associated with the \(k\)-th Fourier coefficient, where \(R_{k,i,j}\) are learnable weights in the spectral domain. Given the total number of potential modes $p$, the cumulative explained ratio is defined as \(g(K, \mathbf{P}) = \sum_{k=1}^{K}{P_k} / {\sum_{k=1}^{p} P_k}\). When \(g(K, \mathbf{P}) < \alpha\) for a chosen threshold \(\alpha \in (0,1)\), additional high-frequency modes are activated until the criterion is met. This ensures that spectral capacity increases only when necessary, keeping training efficient and well-regularized. 
\end{itemize}

\clearpage 

\vspace{-1.5mm}

\paragraph{Incremental Resolution Curriculum.} The second mechanism progressively increases the resolution of the training data. Training begins at the lowest resolution and transitions to finer levels in stages, reusing the learned parameters from each stage as initialization for the next. These transitions may be scheduled after fixed numbers of epochs or triggered dynamically when performance plateaus. It can also leverage data augmentation through downsampling. If high-resolution data are available, lower-resolution versions can be generated on demand to enrich the coarse stages of training. By systematically integrating both real and synthetic examples and different resolutions, iFNO promotes consistency across scales and improves generalization to unseen resolutions. From a computational standpoint, early training on coarse grids reduces memory and time requirements, since operations such as the Fourier transforms scale with the grid size. As resolution increases, the model already encodes coarse-scale behavior, enabling efficient fine-scale adaptation. This hierarchy of learning leads to faster convergence and smoother optimization trajectories.

\vspace{1mm}

\paragraph{Combined Incremental Strategy.} Together, the frequency expansion and resolution curriculum form a unified \mydef{incremental learning strategy}. The model begins with low-frequency, coarse-resolution representations and gradually expands both as training progresses. This harmonizes spatial and spectral growth: the increase in data resolution naturally complements the activation of higher-frequency modes, ensuring that each level of added complexity is meaningfully supported by the data. Empirical results in \cite{George2022Incremental} demonstrate that this incremental process leads to efficient convergence, high predictive fidelity, and robust generalization.

\vspace{1mm}

\paragraph{Implementation.}

The incremental FNO is implemented in \texttt{NeuralOperator 2.0.0} using the \codebox{max\_n\_modes} argument of the \codebox{FNO} class. The parameter \codebox{max\_n\_modes} can be set as a tuple of integers specifying the maximum number of modes that would eventually be used. The actual number of modes used is specified using the usual \codebox{n\_modes} argument of the \codebox{FNO} class. The \codebox{n\_modes} parameter can be updated dynamically during training, up to \codebox{max\_n\_modes}, to gradually use more modes for better expressivity. The \texttt{NeuralOperator 2.0.0} library also provides an \codebox{IncrementalDataProcessor} class in \codebox{neuralop/data/transforms/data\_processors.py}, and an \codebox{IncrementalFNOTrainer} class in \codebox{neuralop/training/incremental.py} leveraging the two adaptive schemes for determining when to increase the number of modes during training presented earlier. A tutorial on how to use them is provided at \href{https://neuraloperator.github.io/dev/auto\_examples/training/plot\_incremental\_FNO\_darcy.html}{neuraloperator.github.io/dev/auto\_examples/training/plot\_incremental\_FNO\_darcy.html}.

\vspace{3mm}

\subsection{Physics-Informed Neural Operators (PINOs)} \label{sec: PINO}

\vspace{1mm}

\subsubsection{Motivation}  \label{sec: PINO Motivation}

\vspace{1mm}

Data-driven neural operators have shown great promise in capturing nonlinear and high-dimensional dynamics, offering rapid and accurate alternatives to traditional numerical solvers. 
Their success, however, often depends on the availability of high-quality data that sufficiently represents the underlying physical processes~\citep{li2024physics}.  In some practical settings, such data can be expensive or difficult to obtain, particularly when generated through high-resolution numerical simulations or complex experiments. As a result, training datasets are frequently limited in size, sparsely sampled, or available only at coarse spatial and temporal resolutions. These data constraints pose a general challenge for data-driven modeling, as models trained under such conditions may struggle to extrapolate reliably or capture fine-scale physical behavior. 

A promising direction to address these challenges is to incorporate physical principles and conservation laws directly into the learning process. This can be achieved by constraining model architectures so that the learned surrogate adheres to the governing equations. Projection-based methods, for example, restrict the model outputs to physically admissible subspaces~\citep{duruisseaux_towards_2024,Harder2024,jiang2020}. Alternatively, structural properties of PDEs can be leveraged to enforce known physical behavior. For instance, incompressibility, $\nabla \cdot u = 0$, can be ensured by representing the velocity field as the curl of a potential, $u = \nabla \times v$, and learning $v$ directly~\citep{Mohan2023}, which guarantees a divergence-free solution. Similarly, Hamiltonian surrogates can preserve the underlying symplectic structure of the dynamics~\citep{BurbyHenon,Duruisseaux2023NPMap,DuruisseauxLieFVINs,Jin2020,Valperga2022}, ensuring (near-) energy conservation and long-term stability. \citet{Cheng2026} use the governing PDE to separate parameter-dependent terms, enabling one operator to generalize across parameter regimes. \citet{yao2026} incorporate partial physics knowledge into pretrained surrogates to improve targeted out-of-distribution counterfactual predictions from factual data alone. When the governing physics are well understood and can be encoded explicitly, these approaches can be very effective, often reducing overfitting and enabling more compact architectures with fewer parameters.

However, these structure-preserving methods are typically tailored to specific PDEs with well-understood dynamics and well-characterized solutions. They depend on detailed prior knowledge of physical properties that can be explicitly encoded into the model, which is often infeasible for complex high-dimensional systems where such information is incomplete. In these cases, physics-informed losses offer a more flexible and broadly applicable alternative. By incorporating the governing equations as soft constraints within the training objective, the model learns to satisfy the underlying physics approximately while retaining the expressive power of data-driven representations. These losses act as effective regularizers that promote physically consistent solutions, reduce overfitting with limited data, and enhance generalization across parameter regimes and unseen conditions. Importantly, this approach requires only knowledge of the governing equations.

\vspace{2mm}

\paragraph{Physics-Informed Neural Networks (PINNs).} Physics-Informed Neural Networks (PINNs)~\citep{Raissi_PINN_OG,Raissi2017Part1,Raissi2017Part2} are neural networks designed to approximate solutions to PDEs. Their parameters are optimized by minimizing deviations from known physics laws such as conservation laws and the governing PDEs themselves. A wide range of PINN variants have been developed and successfully applied to solve PDEs across various domains~\citep{Cai2022,Jagtap2020_2,Yu2022}. Notably, PINNs can be trained without relying on data, using only knowledge of the governing differential equations. 

More precisely, given a function \(a\), consider the problem of finding a solution $u(x,t)$ to the nonlinear PDE
\begin{align}\label{eq: PDE setup}
    \frac{du}{dt} &= \mathcal{R}_a(u) \quad  \text{in } D \times (0,T] , \qquad  u=g_a \quad  \text{on }  \partial D \times (0,T], \qquad   u(\cdot,0)=v_a  \quad  \text{on } D,
\end{align}
where $D \subset \mathbb R^d$ is a bounded open set, $\mathcal{R}_a$ is a nonlinear partial differential operator,  $g_a$ denotes the boundary conditions, and $v_a$ the initial condition. To solve this PDE, a PINN takes a finite grid of points in $D \times (0,T]$ for a fixed final time $T>0$ as input and produces an approximation~$u_\theta(x,t)$ to the solution at each grid point $(x,t) \in D\times(0,T]$ by minimizing the residuals of the differential equation in an appropriate function space norm. A typical choice is the $L^2$ loss, giving the loss function
\begin{align}
\label{eq:pinns-dynamic}
\mathcal{L}_{\text{pde}}(a, u_\theta) 
& \ = \  \Big\|\frac{du_{\theta}}{dt} - \mathcal{R}_a(u_{\theta})\Big\|^2_{L^2(T;D)}  \ + \  \alpha  \ \Big\|u_{\theta}|_{\partial D} - g_a \Big\|^2_{L^2(T; \partial D)} \ + \ \beta  \ \Big\|u_{\theta}|_{t=0} - v_a \Big\|^2_{L^2(D)} . 
\end{align}
Here $\alpha, \beta \in \mathbb R^+$ are constant hyperparameters that can be tuned to balance the contributions of the different terms. In practice, the losses are evaluated on a discrete grid of points and summed up (possibly with quadrature weights) to approximate the loss integrals. The result is then used as the loss function for gradient-based optimization to train the model. The optimization could also be formulated via the variational form~\citep{weinan2018deep}.

Despite their successes and versatility, PINNs are not without limitations. While they provide a flexible framework for embedding physics directly into learning, their optimization process is often challenging due to highly nonconvex and non-smooth loss landscapes~\citep{Wang2021,fuks2020limitations,Krishnapriyan2021}. These difficulties can lead to slow convergence, high sensitivity to hyperparameters, and suboptimal solutions, particularly in multi-scale or time-dependent dynamical systems where the underlying physics spans a wide range of spatial and temporal scales. Moreover, a standard PINN learns the solution corresponding to a single PDE instance and must be re-optimized for each new configuration or equation family, limiting its generalization capability across related problems.

\hfill

\subsubsection{The Physics-Informed Neural Operator (PINO)} \label{sec: PINO Details} 

\vspace{2mm}

The \mydef{Physics-Informed Neural Operator (PINO)}~\citep{li2024physics} framework addresses the challenges of fully data-driven methods and PINNs by combining the data efficiency of PINNs with the generalization capabilities of neural operators. PINOs embed physical knowledge through the governing PDEs, but instead of learning the solution to a single instance, they learn solution operators for entire families of PDEs simultaneously. This formulation allows for strong generalization across a wide range of PDE instances. PINO integrates available training data with the PDE residual loss evaluated at a high resolution to accurately approximate solution operators across diverse PDE families. By enforcing PDE constraints at finer spatial and temporal resolutions, it compensates for coarse or limited data and produces high-fidelity operator approximations. 

The combination of data-driven and physics-based learning also mitigates the optimization challenges that hinder purely physics-based approaches, making PINO well suited for complex and time-dependent PDEs. Empirically, incorporating the PDE loss substantially improves generalization and physical consistency relative to purely data-driven neural operators. As a result, PINO can achieve competitive or superior accuracy while requiring fewer data samples, and has been shown to outperform data-trained FNOs on high-resolution benchmarks~\citep{rosofsky2022applications}. While the original PINO framework~\citep{li2024physics} leverages the FNO architecture, the approach naturally extends to other architectures such as graph neural operators~\citep{lin2025mGNO}. 

Learning the operator in function space often yields a smoother optimization landscape than learning a single solution directly as in PINNs, since optimizing coefficients in a learned functional basis is typically easier than optimizing an entire solution function. In the PINO framework, training proceeds in two phases: 
\begin{itemize}

    \vspace{2mm}
    
    \item \mydef{Pretraining} (or operator learning): a neural operator is trained to approximate the ground-truth solution operator for a family of PDEs using available data, synthetically generated PDE instances, or a hybrid combination of both. PINO can also be trained on low-resolution data enhanced with high-resolution PDE constraints, enabling accurate extrapolation to finer scales.

    \vspace{2mm}
    
    \item \mydef{Instance-wise fine-tuning} (or test-time optimization): The pretrained neural operator can be further refined to solve a single PDE instance using only the PDE loss to minimize residual errors. An additional \emph{anchor loss} can be incorporated as a regularization term to stabilize optimization and ensure that the updated operator remains close to the pretrained model, e.g. via  $\|\mathcal{G}_{\theta_i}(a) - \mathcal{G}_{\theta_0}(a)\|_{L^2}$  where $\mathcal{G}_{\theta_i}$ and  $\mathcal{G}_{\theta_0}$ are the fine-tuned and pretrained operators, respectively. Because the learned operator parametrizes the solution in function space, fine-tuning focuses only on adjusting the operator outputs rather than re-optimizing the full model, enabling efficient adaptation to individual PDE instances.
\end{itemize}

    \vspace{2mm}
    
This two-phase process offers a trade-off between generalization and accuracy: the pretrained operator enables very fast inference across diverse PDE instances, while instance-wise fine-tuning can improve accuracy up to near-exact solutions while incurring higher computational costs. 

    \vspace{2mm}

Hybrid strategies for constraint enforcement are also possible, where certain constraints are embedded directly into the model architecture, while others are imposed through physics training losses. For instance, in incompressible Navier--Stokes problems, the momentum equations can be  embedded via a physics loss, while the incompressibility condition can be enforced explicitly using projection techniques from~\citep{jiang2020,duruisseaux_towards_2024}. Note that a \codebox{spectral\_projection\_divergence\_free} function is available in \codebox{neuralop/layers/spectral\_projection.py} for enforcing the divergence-free condition, with a tutorial made available at \href{https://neuraloperator.github.io/dev/auto_examples/layers/plot\_spectral\_projection.html}{neuraloperator.github.io/dev/auto\_examples/layers/plot\_spectral\_projection.html}.

PINO can also be applied to inverse problems, either by learning the forward operator and performing gradient-based inversion or by directly learning the inverse operator~\citep{li2024physics}. In both cases, enforcing the PDE constraint serves as a powerful regularizer that ensures physically consistent and stable inverse solutions, even when observations are limited or noisy.

\hfill 

\subsubsection{Computing Derivatives} \label{sec: Computing Derivatives} 

\vspace{2mm}

Effective training of neural operators with physics losses depends critically on the accurate and efficient computation of derivatives. Even small numerical inaccuracies in derivatives can be amplified through the physics losses, ultimately compromising the fidelity and stability of the predicted
solutions.

\paragraph{Finite Differences.}  A straightforward approach for estimating derivatives numerically is the use of \mydef{finite differences}, which approximate derivatives by evaluating differences between neighboring function values on a discrete grid. This method is simple to implement, computationally efficient, and memory-light, requiring only $\mathcal{O}(n)$ operations for a grid containing $n$ points. Owing to its simplicity, it remains a standard tool in numerical analysis and scientific computing. 

Nevertheless, finite-difference schemes inherit the same limitations as finite-difference numerical solvers: accurate results often demand fine spatial or temporal resolutions. On coarse grids, derivative estimates can degrade significantly, introducing large numerical errors. As a result, finite differences can become computationally burdensome or impractical when modeling multi-scale phenomena or rapidly varying dynamics. Note that higher-order finite difference schemes or stencils can sometimes help mitigate these issues. 

Higher-order finite difference formulas can be viewed as exact derivatives of a
local interpolating polynomial.  Instead of manipulating Taylor expansions term
by term, we fit a low-degree polynomial $p$ to the values of the function $f$ on a small
neighborhood and then differentiate that polynomial at the point of interest.
The resulting derivative is a linear combination of nearby samples with weights
chosen so that all Taylor terms up to a prescribed order are reproduced exactly. More precisely, let $f$ be smooth and let $y_j = f(x_j)$ denote its values at grid points $x_1,\dots,x_n$.  Around a target point $x_\bullet$ we approximate $f$ by a polynomial of degree at most $(n-1)$, 
\begin{equation}
    p(x) = \sum_{k=0}^{n-1} c_k \frac{(x-x_\bullet)^k}{k!}
\end{equation}
that interpolates the data on the chosen stencil, $p(x_j) = y_j$.  This defines a linear system $V c = y$ with entries $
    V_{jk} = (x_j - x_\bullet)^k/k!$. By construction $c_k = p^{(k)}(x_\bullet)$ so the $k$-th derivative of $f$ at $x_\bullet$ is approximated by $c_k$.  Intuitively, we replace $f$ near $x_\bullet$ by the unique polynomial that matches its nearby samples and then differentiate this polynomial, which turns the derivative into an explicit linear combination of the $y_j$.

For a uniform grid, $x_j = x_\bullet + s_j\,\delta x$, where $\delta x$ is the grid spacing and the integers $s_j$ encode the stencil offsets. The $k$-th order finite difference stencil then has the form
\begin{equation}
f^{(k)}(x_\bullet) \approx \frac{1}{\delta x^k} \sum_{j=1}^{n} w^{(k)}_j f(x_j),
\end{equation}
with weights $w^{(k)}_j$ that depend only on the integer offsets $s_j$ and can be precomputed once for each stencil pattern. This construction is purely local and depends only on the relative positions $s_j$ (and in particular not on the target point $x_\bullet$).  The same precomputed table of weights applies at every grid point that uses the same pattern of offsets.

If more than $n$ points are available, one can fit the polynomial in the least squares sense rather than by exact interpolation.  Let $X$ denote the resulting rectangular design matrix built from powers of $(x_j - x_\bullet)$ and define
\begin{equation}
    c = \arg\min_{\tilde c} \| X \tilde c - y \|_2^2
= (X^{\top} X)^{-1} X^{\top} y.
\end{equation}
The same scaling and extraction of the $k$-th coefficient then produces finite difference weights that are no longer exact for all polynomials of degree $(n-1)$ but are more robust to noise.  In both the interpolatory and least squares settings, higher-order finite difference stencils arise as derivatives of systematically constructed local polynomial fits.

\hfill 

\paragraph{Automatic Differentiation.} Derivatives can be evaluated pointwise using \mydef{automatic differentiation}, which applies the chain rule to the sequence of operations in a model. Libraries such as Autograd~\citep{Autograd} automate this process by constructing a computational graph during the forward pass and using reverse-mode differentiation to compute gradients efficiently in the backward pass. Automatic differentiation is often favored over finite-difference methods in PINNs~\citep{Baydin2017} because it yields highly accurate derivatives even at low resolutions, requires only a single function evaluation, and enables efficient computation of higher-order derivatives with minimal overhead. In contrast, finite differences demand multiple evaluations and are more susceptible to numerical errors. However, automatic differentiation also has limitations: all operations must be differentiable, and storing intermediate values in the computational graph can lead to significant memory usage in deep models. Moreover, when the physics loss involves deep compositions, vanishing or exploding gradients may arise during backpropagation, and for sufficiently coarse grids, finite-difference methods can sometimes offer comparable accuracy at lower cost. 

When using an FNO as the backbone of a PINO, the FNO directly outputs a numerical solution $u$ on a regular grid (as opposed to pointwise evaluations), we need to define a query function $u(x)$ which can be evaluated at any point, enabling autograd to compute $\nabla_x u(x)$ by backpropagating through some (or all) of the operator layers. This formulation is fully compatible with modern autograd frameworks and provides exact analytical derivatives within numerical precision. Integrating PINO with neural architectures that operate independently of regular grids and fast Fourier transforms, such as mollified GNOs and GINOs~\citep{lin2025mGNO}, is generally more straightforward. 

\hfill

\paragraph{Spectral Differentiation.} As discussed in \Cref{sec: Spectral Differentiation}, spectral differentiation provides a fast and memory-efficient approach to compute derivatives by operating in the frequency domain. This method is particularly effective for smooth and periodic solutions, offering very high accuracy even at moderate resolutions, and can be efficiently implemented using the FFT. A key advantage of spectral differentiation is that higher-order derivatives can be computed at essentially no additional cost, whereas finite-difference schemes require increasingly wide stencils and higher computational expense for the same purpose. Spectral differentiation is also significantly more memory-efficient than automatic differentiation, which must store the entire computational graph and intermediate activations during training. In high-dimensional PDEs or deep architectures, this often leads to excessive memory consumption and slow backward passes when evaluating physics-based losses. In contrast, spectral differentiation requires only the storage of the function values and their Fourier coefficients. When applied to periodic problems, spectral differentiation becomes the differentiation method of choice for PINO for regular-grid problems due to its combination of efficiency, accuracy, and simplicity. This approach enables the efficient computation of highly accurate derivatives, which makes it especially suitable for applications that demand high-precision solutions.

For non-periodic problems, one can embed the original non-periodic signal within a larger domain and define its extension so that it becomes periodic on the extended domain. Spectral differentiation can then be applied to the signal
on the larger domain. Simple padding approaches to construct the extension can sometimes be sufficient to obtain the desired PDE solutions. In demanding settings that require highly accurate solutions, padding methods may fail to yield sufficiently precise derivatives because the convergence rate of Fourier series depends on smoothness, while padding often introduces non-differentiable behavior in the function or in its low-order derivatives. In this case, Fourier continuation techniques with higher-precision arithmetic can prove very helpful and successful, as discussed in \Cref{sec: Non-periodicity} and done in FC--PINO~\citep{ganeshram2025fcpinohighprecisionphysicsinformed}. 

It should be noted, however, that the high accuracy and effectiveness of spectral differentiation and Fourier continuation methods primarily apply to data-free or noise-free data settings. When the underlying function is contaminated by noise, spectral differentiation can become ill-conditioned and lead to a significant loss of accuracy. To mitigate the amplification of errors associated with high-frequency noise, spectral differentiation is often combined with low-pass spectral filtering in noisy settings. Such spectral filtering reduces the impact of noise but inevitably degrades the accuracy of the resulting spectral derivatives. In these situations, alternative differentiation techniques, such as high-order finite difference schemes, can sometimes yield more reliable and accurate results.

\hfill

\subsubsection{Multi-Objective Optimization} \label{sec: Multi-objective optimization}

\vspace{2mm}

Introducing physics-based losses in the objective function typically requires the use of loss coefficients (e.g. $\alpha, \beta \in \mathbb{R}^+$ in \Cref{eq:pinns-dynamic}) to balance the contributions of the different terms in the total objective. Multi-objective optimization with such losses can be challenging due to complex and often ill-conditioned loss landscapes~\citep{Krishnapriyan2021}. The gradients of individual terms may differ greatly in magnitude and direction, leading to unbalanced updates and preventing simultaneous reduction of all components during training~\citep{Wang2021}. Manually tuning these coefficients becomes prohibitively expensive as the number of loss terms grows.

This motivated the development of adaptive weighting strategies that automatically balance relative contributions, such as \texttt{SoftAdapt}~\citep{Heydari2019} and \texttt{ReLoBRaLo} (Relative Loss Balancing with Random Lookback)~\citep{Bischof2021}. 

\texttt{SoftAdapt} dynamically balances multiple loss terms by monitoring their relative magnitudes and rates of change across training steps. Rather than relying on gradient information, it uses simple loss statistics to determine how much each objective should contribute to the overall optimization. Intuitively, the method assigns higher weights to loss components that are either decreasing more slowly or remain relatively large, allowing the model to focus on underperforming objectives and promote balanced progress among all tasks. The scaling coefficients are computed as
\begin{equation}
\lambda_i(t) = \frac{\exp\left(\tau \, [\mathcal{L}_i(t) - \mathcal{L}_i(t-1)]\right)}{\sum_{j=1}^k \exp\left(\tau \, [\mathcal{L}_j(t) - \mathcal{L}_j(t-1)]\right)}, \quad i \in \{1, \dots, k\},
\end{equation}where $\tau$ controls how strongly differences in loss progress influence the weighting.

\texttt{ReLoBRaLo} builds upon \texttt{SoftAdapt} by introducing a stochastic memory mechanism that allows the model to balance losses using both recent and long-term progress information. Instead of relying solely on consecutive loss changes, it occasionally compares each loss to its initial value through randomly chosen lookback intervals. This combination of short- and long-term feedback helps the model maintain steady improvement across objectives while avoiding overfitting to recent fluctuations. The scaling coefficients are computed as
\begin{equation}
\lambda_i(t) = \alpha \lambda_i^{\text{hist}} + (1 - \alpha) \lambda_i^{\text{bal}}(t, t-1),
\end{equation} \vspace{-4mm}
\begin{equation}
\lambda_i^{\text{hist}}(t) = 
\rho \lambda_i(t-1) + (1 - \rho) \lambda_i^{\text{bal}}(t, 0)
\ \ 
\text{and } \ \   \lambda_i^{\text{bal}}(t, t') = 
m \exp\!\left(\!\frac{\mathcal{L}_i(t)}{\tau \, \mathcal{L}_i(t')}\!\right) 
\Big/ 
\sum_{j=1}^m \exp\!\left(\!\frac{\mathcal{L}_j(t)}{\tau \, \mathcal{L}_j(t')}\!\right).
\end{equation}Here, $\alpha$ controls the influence of past weightings, $\tau$ regulates the sharpness of the softmax distribution, and $\rho$ (a Bernoulli random variable) determines whether the update references the most recent or the initial loss, thereby implementing the ``random lookback.'' This design allows \texttt{ReLoBRaLo} to adapt smoothly to changing training dynamics, improving convergence stability and efficiency without relying on gradient-based statistics.

\texttt{SoftAdapt} and \texttt{ReLoBRaLo} are available in \codebox{neuralop/losses/meta\_losses.py} in \texttt{NeuralOperator 2.0.0}, based on the implementation from \texttt{NVIDIA PhysicsNeMo}~\cite{physicsnemo}, and can be instantiated via

\vspace{1mm}

\begin{minted}[fontsize=\normalsize, bgcolor=gray!5, frame=single, linenos]{python}
train_loss = SoftAdapt(num_losses=num_losses, params=model.parameters())
train_loss = Relobralo(num_losses=num_losses, params=model.parameters())
\end{minted}

\vspace{2mm}

Individual loss components are then computed and stored in a dictionary which is passed to the aggregator that automatically computes both the total weighted loss and individual weights. This design allows easy integration into existing training loops without modifying the loss definitions themselves.

\vspace{1mm}

\begin{minted}[fontsize=\normalsize, bgcolor=gray!5, frame=single, linenos]{python}
loss_map = {"l2": L2loss, "ic": ICloss, "equation": eq_loss}
loss_values = {name: [] for name in list_of_loss_names}
# loop over training batches
    # loop over different loss terms
    for loss_name in list_of_loss_names:
        loss_values[loss_name].append(loss_map[loss_name](pred, target))
    # Aggregate losses adaptively
    total_loss, weights = train_loss(loss_values, step=epoch)
    total_loss.backward()
\end{minted}

\vspace{2mm}

\codebox{script/train\_burgers\_pino.py} provides an example script showing how to train PINO with \texttt{SoftAdapt} and \texttt{ReLoBRaLo}. Regarding the hyperparameters of \texttt{ReLoBRaLo}, the default argument values are typically fine. \codebox{alpha} controls the exponential moving average smoothing of weights, with higher values leading to slower and smoother adaptation. \codebox{beta} governs how often random lookback comparisons are used, where larger values imply longer memory and increased stability. \codebox{tau} adjusts the sharpness of the weighting distribution, focusing more strongly on the largest loss when smaller and promoting uniformity when larger.

\subsection{Uncertainty Quantification with Neural Operators (UQNO)} \label{sec: UQNO}

\vspace{2mm}

Although neural operators have demonstrated high predictive power, they typically output only deterministic estimates and offer no measure of confidence. Certain scientific applications of operator learning require uncertainty estimates that are both spatially resolved and statistically reliable. In many domains, practitioners must not only obtain accurate field predictions but also quantify confidence at every point in the domain. Such safety-critical contexts demand uncertainty representations that remain valid across the entire spatial field rather than only at isolated query points or global averages. 

Conventional deep-learning heuristics such as Monte Carlo dropout \cite{gal2016dropout} or ensembles provide approximate uncertainty but lack distribution-free coverage guarantees and often assume Gaussian behavior. Bayesian operator variants can, in principle, provide posterior uncertainty but are computationally prohibitive at realistic discretizations. Previous attempts at functional calibration focus either on individual spatial points or on global scalar summaries and typically assume homoscedastic noise, i.e. that prediction errors have the same variance everywhere, regardless of how easy or difficult a region of the input space is to model. 

While homoscedasticity simplifies the statistical treatment of uncertainty, real-world scientific and engineering problems often exhibit spatially or temporally varying sources of uncertainty. For instance, in a turbulent fluid simulation, errors are typically higher near boundaries or discontinuities, whereas smooth interior regions are more predictable. In image reconstruction or surrogate modeling of PDEs, regions with sharp gradients or fine-scale structures tend to accumulate larger model errors. Such settings require an uncertainty representation that adapts to the local complexity of the data, otherwise users face a trade-off between under-calibrated or overly conservative uncertainty bands \cite{guo2024ibuq}. Related theoretical work studies out-of-distribution risk bounds for neural operators and architectural factors affecting generalization, which is complementary to but distinct from calibrated uncertainty estimation \cite{benitez2024oodno}.

\vspace{2mm}

The \mydef{Uncertainty-Quantified Neural Operator (UQNO)}~\cite{ma2024uqno} provides a principled framework for uncertainty quantification in operator learning, by introducing a \emph{residual operator} that learns to predict the local magnitude of model errors and then applying a \emph{conformal calibration} step that adjusts the scale of these error estimates to ensure reliable coverage in function space. Together, these two processes yield uncertainty estimates that are adaptive, interpretable, and supported by formal statistical guarantees.

Given a neural operator \(\tilde{\mathcal{G}}\) trained to predict a solution \(u\) from an input function \(a\), UQNO defines a second neural operator \(\mathcal{E}(a)(x)\) that estimates the magnitude of the model's residual error at each location \(x\). This residual operator is trained with a generalized quantile loss, which encourages \(\mathcal{E}(a)(x)\) to approximate the upper quantiles of the empirical error distribution, thereby learning where and how strongly the neural operator \(\tilde{\mathcal{G}}\) tends to make errors. The resulting field \(\mathcal{E}(a)(x)\) provides heteroscedastic uncertainty estimates (i.e. estimates varying across the spatial domain), reflecting the intrinsic complexity of the underlying physics and the model's confidence in different regions. Regions near discontinuities, sharp gradients, or boundaries naturally receive higher uncertainty scores, while smooth interior regions with stable predictions receive lower ones. 

Although the residual operator $\mathcal{E}$ captures useful structure, its estimates are not automatically calibrated. There is no guarantee that the predicted uncertainty bounds will contain the true solution with a specified frequency. To ensure formal coverage guarantees, UQNO employs a split conformal prediction procedure that calibrates a single global scaling factor using a held-out dataset. Specifically, for each calibration example, it computes the ratio between the true error and the predicted uncertainty magnitude,
\begin{equation}
\mathcal{R}(a,x) = \frac{|u(a)(x) - \tilde{\mathcal{G}}(a)(x)|}{\mathcal{E}(a)(x)}.
\end{equation}This indicates how much the model's uncertainty predictions need to be inflated to achieve valid coverage. UQNO then determines a single scalar multiplier \(\lambda\) corresponding to a desired pair of tolerance parameters \((\alpha, \delta)\), which respectively control the allowable fraction of uncovered spatial points and the allowable fraction of entire functions that may violate the coverage criterion. The final uncertainty region is then defined as
\begin{equation}
C_{\lambda}(a)(x) = \bigl\{\, p \in \mathbb{R}^{d_u} : \|p - \tilde{\mathcal{G}}(a)(x)\|_2 \le \lambda\,\mathcal{E}(a)(x)\,\bigr\}.
\end{equation}This construction guarantees that, with probability at least \((1 - \delta)\) over unseen functions, at least a fraction \((1 - \alpha)\) of spatial points will be covered by the predicted uncertainty intervals. The coverage guarantee provided by UQNO is finite-sample in the sense that it holds for a limited amount of calibration data rather than only in the asymptotic regime of infinite data. It is also distribution-free because it does not rely on any assumed form of the underlying noise or data distribution. The only requirement is that the calibration and test samples are exchangeable, meaning they are drawn from the same statistical process. This mild assumption allows UQNO to provide rigorous and reliable uncertainty coverage even in complex scenarios where the noise is non-Gaussian, heteroscedastic, or otherwise difficult to model parametrically.

This combination of a learned residual field and a conformal scaling factor provides both adaptability and validity. The residual field \(\mathcal{E}(a)(x)\) captures the spatial distribution of uncertainty, reflecting where the model is intrinsically more uncertain, while the calibration factor \(\lambda\) enforces global statistical correctness. As a result, UQNO produces uncertainty bands that are sharp in confident regions yet wide enough in challenging areas to achieve the target coverage. The calibration step requires only simple quantile computations, and the residual operator can reuse the same architecture as the base model, which makes the approach scalable to large domains and high resolution data without the overhead of Bayesian sampling or model ensembles.

In practice, UQNO can be implemented as a two-stage pipeline. The first stage trains the base neural operator to predict the target solution field, followed by a second training phase for the residual operator that predicts the local magnitude of the errors. In the second stage, a small calibration dataset is used to determine the appropriate global scale factor \(\lambda\). The final model outputs both the predicted field and its corresponding uncertainty map, providing a complete picture of the model's confidence across the domain. The use of UQNO is exemplified in the \codebox{scripts/train\_uqno\_darcy.py} in \texttt{NeuralOperator 2.0.0} using the \codebox{UQNO} class from \codebox{neuralop/models/fno.py}.

\clearpage 
\section{Advanced Architectural Modifications}  \label{sec: Advanced Architectural Modifications}

\subsection{Tensor Fourier Neural Operators (TFNO)}
\label{sec: tfno}

\vspace{1mm}

As discussed earlier, FNOs learn mappings between function spaces by performing spectral convolutions on the Fourier coefficients of input signals. Each layer learns a complex-valued weight matrix in the frequency domain that maps input channels to output channels for each retained Fourier mode. In their standard form, these matrices are dense and independently parameterized, resulting in substantial redundancy, increased memory and computational cost, and a tendency to overfit due to over-parameterization. 

To address this, \citet{Kossaifi2023MGTFNO} introduced the \mydef{Tensor Fourier Neural Operator (TFNO)}, which factorizes the spectral weight matrices using low-rank tensor decompositions~\citep{kolda2009tensor}. This formulation provides a principled and computationally efficient means to remove redundancy across frequencies and channel dimensions, achieving significant parameter savings while maintaining model expressivity. In particular, the Tucker decomposition, which can be regarded as a multidimensional generalization of the singular value decomposition (SVD), offers a stable and interpretable framework for compressing and regularizing spectral operators without compromising their structural richness.

\begin{figure*}[t]
\centering
\includegraphics[width=0.8\textwidth]{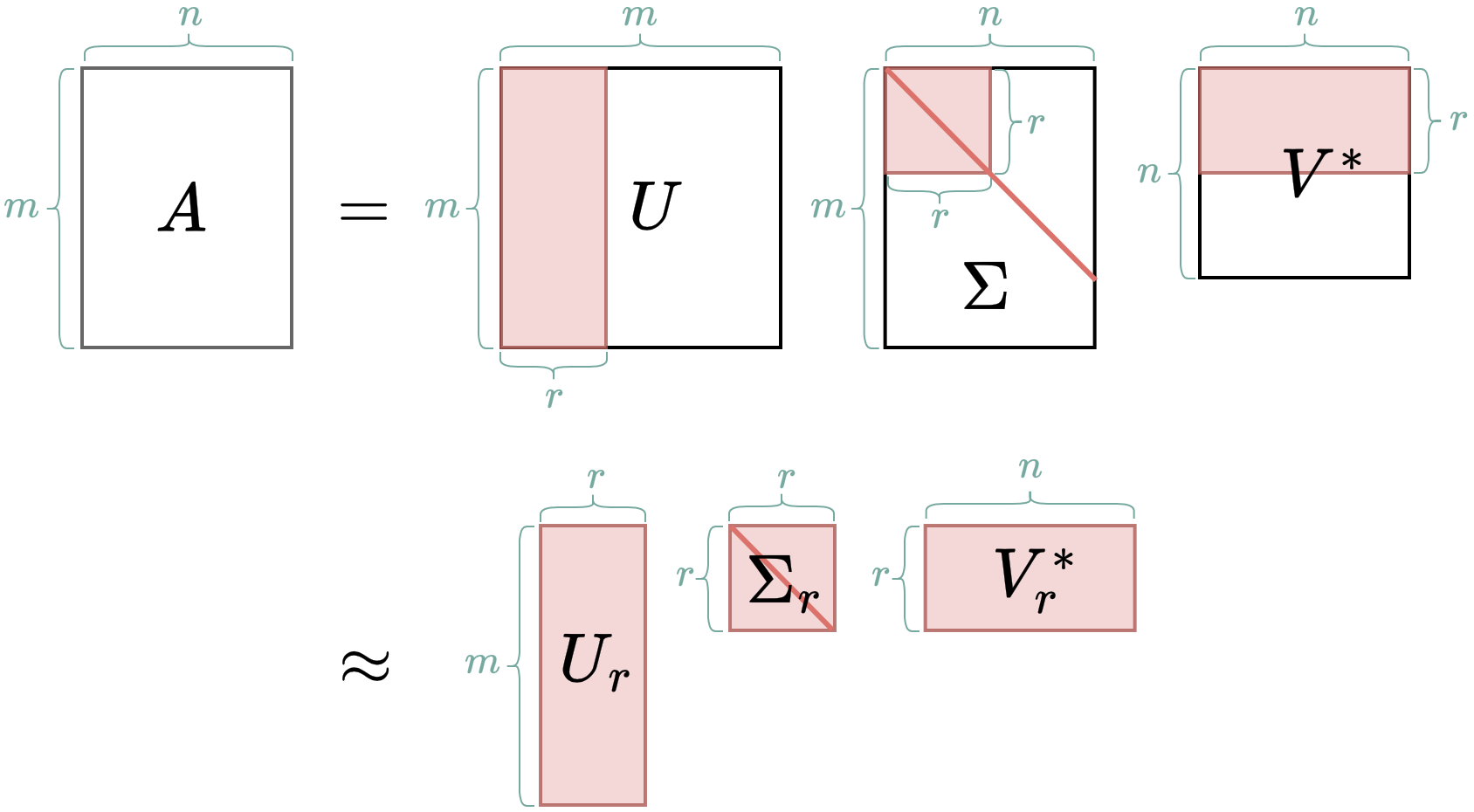}  \vspace{-0mm}
\caption{The Singular Value Decomposition (SVD) decomposes a matrix $A \in \mathbb{C}^{m \times n}$ as $U \, \Sigma \, V^*$, where $U \in \mathbb{C}^{m \times m}$ and $V \in \mathbb{C}^{n \times n}$ are unitary matrices, and $\Sigma \in \mathbb{R}^{m \times n}$ is a diagonal matrix whose positive entries are the singular values of $A$. The rank-$r$ SVD approximation to $A$ is obtained via $A \approx U_r \, \Sigma_r \, V_r^*$ by truncating the matrices $U, \Sigma ,$ and $V^*$ as depicted. \label{fig: SVD} } \vspace{3mm}
\end{figure*}

\paragraph{Singular Value Decomposition.}  The \mydef{Singular Value Decomposition (SVD)} is a foundational tool in numerical linear algebra that provides a compact and interpretable representation of a matrix. Given a matrix $A \in \mathbb{C}^{m \times n}$, the SVD factorizes it into three matrices:
\begin{equation}
    A = U \, \Sigma \, V^*
    \label{eq: svd}
\end{equation}
where $U \in \mathbb{C}^{m \times m}$ and $V \in \mathbb{C}^{n \times n}$ are unitary matrices containing the left and right singular vectors, respectively, and $\Sigma \in \mathbb{R}^{m \times n}$ is a diagonal matrix whose positive entries are the singular values of $A$. Each singular value represents the importance of a corresponding pair of singular vectors in describing the structure of the matrix. Truncating the decomposition by retaining only the largest $r$ singular values yields a \mydef{low-rank SVD approximation} that minimizes reconstruction error in the Frobenius norm sense. \Cref{fig: SVD} illustrates the SVD and how the low-rank SVD approximation leads to compression.

Intuitively, the SVD identifies orthogonal directions along which the matrix acts as a pure scaling operation. This property makes it a powerful tool for dimensionality reduction, noise filtering, and identifying dominant patterns in data. The SVD is also referred to as the Proper Orthogonal Decomposition (POD) in the context of reduced-order modeling, where the singular vectors define the dominant spatial modes of the system and the singular values measure the energy or significance associated with each mode. In practice, low-rank SVD approximations are widely used to compress models, denoise signals, and regularize learning systems by capturing only the most significant modes of variation. This same principle motivates the use of tensor factorizations such as the Tucker decomposition for higher-order data, where similar ideas extend naturally to multi-dimensional arrays representing more complex relationships.

\paragraph{Tucker Decomposition.} Tensors generalize vectors and matrices to higher orders.  An $N$-th order tensor $\mathcal{T} \in \mathbb{C}^{I_1 \times I_2 \times \cdots \times I_N}$ is an $N$-dimensional array where each mode $n$ has dimension $I_n$. Like matrices, tensors can be decomposed into smaller, low-rank components that capture their dominant interactions.  A widely used form of tensor decomposition is the Tucker decomposition, also known as the higher-order singular value decomposition (HOSVD)~\citep{kolda2009tensor}. The \mydef{Tucker decomposition} expresses a tensor as a core tensor multiplied by a matrix along each mode:
\begin{equation}
    \mathcal{T} \ \approx \  \Sigma_{core} 
    \  \times_1 \  U^{(1)} 
    \ \times_2 \ U^{(2)} 
   \  \cdots \ 
    \times_N \  U^{(N)}
    \label{eq: tucker}
\end{equation}
where $\Sigma_{core} \in \mathbb{C}^{R_1 \times R_2 \times \cdots \times R_N}$ is the \emph{core tensor}, $U^{(n)} \in \mathbb{C}^{I_n \times R_n}$ are the \textit{factor matrices}, and $\times_n$ denotes the $n$-mode product.  The ranks $(R_1, R_2, \ldots, R_N)$ of the factor matrices determine the level of compression.

In the special case where $N = 2$, the Tucker decomposition reduces to the standard SVD. In this case, the core tensor $\Sigma_{core}$ corresponds to the diagonal matrix of singular values $\Sigma$, and the factor matrices $U^{(1)}$ and $U^{(2)}$ correspond to the left and right singular vector matrices. Thus, Tucker decomposition naturally generalizes SVD to higher dimensions. This property allows methods based on low-rank matrix factorization to extend seamlessly to multi-way data structures such as the spectral weight tensors in FNOs.

In the standard Tucker decomposition, the factor matrices are typically chosen to be orthogonal, similar to the SVD case. Orthogonality of the factor matrices provides computational and numerical benefits: it allows efficient projections using simple matrix multiplications, enables reconstruction via transposition instead of inversion, and ensures numerical stability since orthogonal matrices have a condition number of one. \\ 

\begin{figure}[t]  \vspace{-1mm}
\centering   
\includegraphics[width=0.84\textwidth]{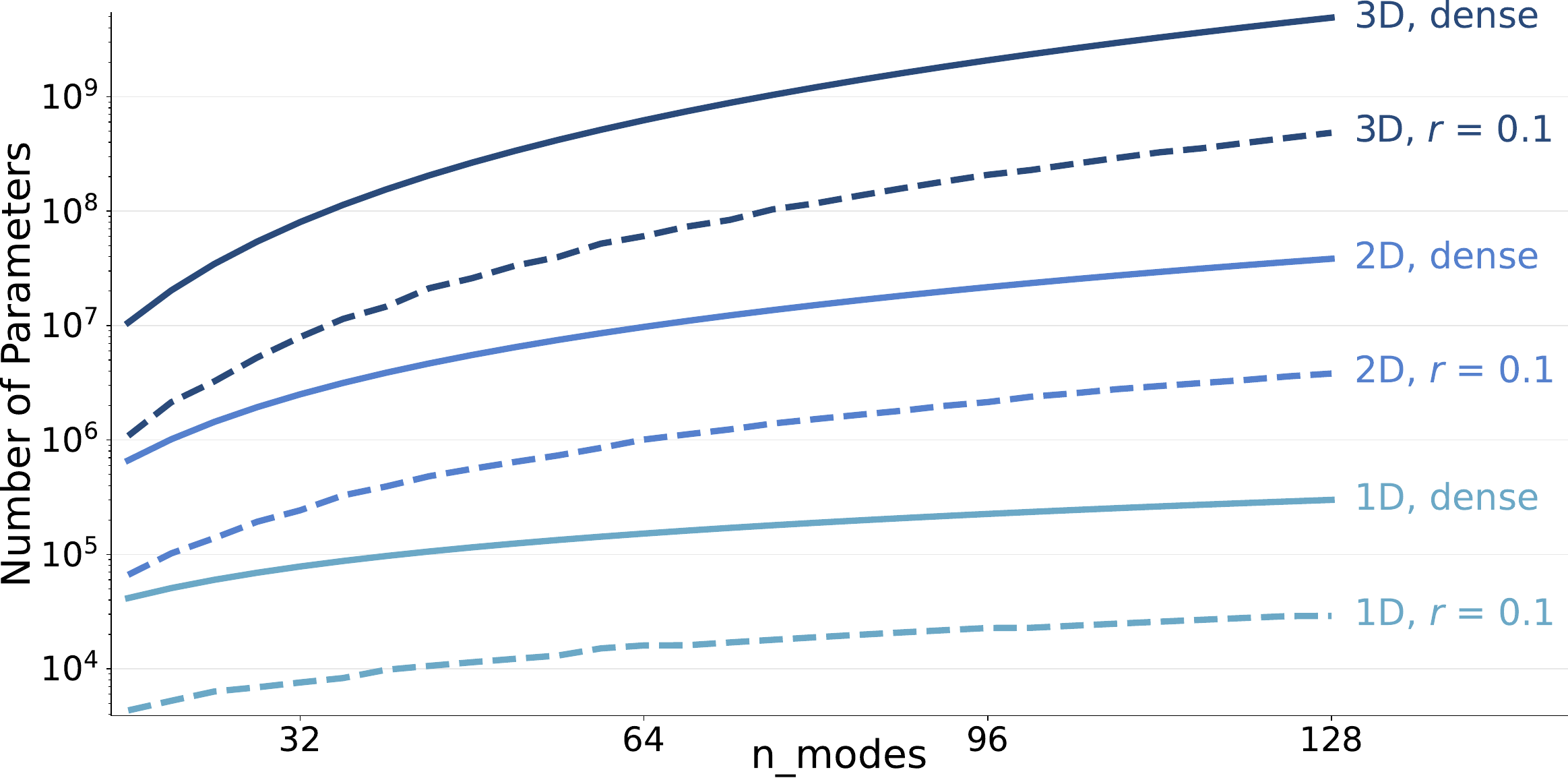}  \vspace{-2mm}
\caption{Evolution of the parameter count in the \codebox{SpectralConv} (with 48 input and output channels) as the number of modes increases, with and without Tucker factorization. Both settings show similar growth rates, while factorization with a rank of 0.1 maintains roughly 10$\times$ fewer parameters than the dense case. } \label{fig: TFNO} 
\end{figure}

\paragraph{Tucker Decomposition of Spectral Weights in FNOs.} Each set of spectral weights across all $K$ Fourier modes and channel pairs can be represented as a third-order tensor 
\begin{equation}
\mathcal{W} \in \mathbb{C}^{K \times I \times O},
\end{equation}
where $I$ and $O$ are the input and output channel dimensions, respectively.  Instead of storing $\mathcal{W}$ as a dense tensor with $\mathcal{O}(KIO)$ parameters, we approximate it via Tucker decomposition:
\begin{equation}
    \mathcal{W} \ \approx \
   \  \Sigma_{core}  
  \  \times_1 \ U^{(K)} 
    \ \times_2 \ U^{(I)} 
  \   \times_3 \ U^{(O)},
    \label{eq:weight_tucker}
\end{equation}
where $ \Sigma_{core} \in \mathbb{C}^{R_K \times R_I \times R_O}$ is a compact core tensor and $U^{(K)}, U^{(I)}, U^{(O)}$ are factor matrices along the Fourier, input-channel, and output-channel modes. This parameterization reduces the total number of parameters to
\begin{equation}
R_K R_I R_O + K R_K + I R_I + O R_O,
\end{equation}
which is typically $5\times$ to $20\times$ smaller than the original dense representation. \Cref{fig: TFNO} shows the evolution of the parameter count in the \codebox{SpectralConv} as the number of modes increases, both with and without Tucker factorization with rank 0.1.

This low-rank factorization preserves the essential structure of the spectral operator while removing redundancy. Each factor matrix captures a shared low-dimensional subspace that reflects a specific aspect of the data, such as spatial frequency, input channels, or output channels. The core tensor $ \Sigma_{core}$ encodes the interactions between these subspaces. In practice, this structured representation regularizes the learning process and improves generalization, especially in settings with limited data.

\paragraph{Implementation.} The TFNO is implemented in the \texttt{NeuralOperator 2.0.0} library and can be instantiated either directly through the \codebox{TFNO} class in \codebox{neuralop/models/fno.py}, or by using the \codebox{FNO} class with the parameter \codebox{factorization=``tucker"}. When \codebox{rank} is specified as a floating-point value between 0 and 1, it controls approximately the fraction of parameters of the dense weight tensor retained by the Tucker representation, rather than specifying the same fractional multilinear rank along every tensor dimension. During TFNO training, the Tucker core tensor and factor matrices $U^{(n)}$ are instantiated directly as trainable parameters and optimized jointly with the other model parameters. This should be distinguished from Tucker decomposition algorithms such as Higher-Order Orthogonal Iteration (HOOI), which compute a low-rank approximation of an already existing dense tensor through alternating SVD-based updates. Such decomposition routines can be used when initializing or converting a dense tensor to a Tucker representation, but they are not the standard mechanism by which the TFNO factors are obtained during training.

\hfill

\subsection{Localized Differential and Integral Kernels}
\label{sec: LocalNO}

The standard FNO relies on global spectral convolutions to efficiently capture long-range dependencies and smooth structures in function space. While this global formulation is powerful for representing spatially homogeneous systems, many physical phenomena exhibit localized interactions or multi-scale behavior where short-range effects are equally important. Examples include systems with discontinuities, sharp interfaces, heterogeneous materials, or localized forcing terms. In such cases, the global Fourier representation alone may not fully capture local variations or non-stationary spatial correlations. 

While the channel-mixing MLP and the skip connections restore and enrich high-frequency details, further additions to the FNO architecture have been developed in \citep{liu2024neural} that introduce \mydef{localized differential and integral kernel} components. These augmentations retain the computational and theoretical advantages of the FNO while expanding its expressive range to include both local and spatially varying nonlocal operators.

The goal of local integral kernel layers is to construct convolutional mappings that remain well-defined in the continuum limit. In particular, we seek formulations that do not collapse to pointwise operations as the grid resolution increases, but instead converge to either local integral or differential operators. 

\paragraph{Differential Layers.} Convolutional kernels can be viewed as discrete approximations of continuous integral operators. When the kernel is symmetric and its weights sum to one, the convolution acts as a local averaging operator. If the kernel is centered and appropriately scaled, it approximates a differential operator. This provides a principled way to endow neural operators with learnable local differential structure.

More precisely, given a function \( v  \) sampled on a regular grid with spacing \( h \), a discrete convolution with kernel \( K = (K_1, \ldots, K_S) \) can be written as
\begin{equation}
\operatorname{Conv}_K[v](y) = \sum_{i=1}^{S} K_i\, v(y + z_i),
\qquad
z_i = h\!\left(i - 1 - \frac{S-1}{2}\right).
\end{equation}
As the resolution increases and \(h \to 0\), this operator converges to a local integral that, for non-centered kernels, corresponds to averaging:
\begin{equation}
\lim_{h \to 0} \operatorname{Conv}_K[v](y) = \overline{K}\, v(y),
\qquad
\overline{K} = \sum_{i=1}^{S} K_i.
\end{equation}
By centering the kernel and scaling by \(1/h\), the convolution captures local differences, yielding a first-order differential operator in the continuous limit:
\begin{equation}
\lim_{h \to 0} \frac{1}{h}\, \operatorname{Conv}_{K - \overline{K}}[v](y)
    = \sum_{j=1}^{n} b_j\, \nabla v_j(y),
\end{equation}
where \(b_j\) depends on the first moments of \(K\). Higher-order derivatives follow analogously by scaling with \(1/h^k\), or can be approximated by composing first-order differential layers.

The resulting differential layer provides a continuous interpretation of local convolutions and serves as a natural complement to the global spectral components within neural operator architectures.

\hfill 

\paragraph{Local Integral Kernel Layers.} This paragraph focuses on kernels that preserve a fixed receptive field and represent local integral operators. A general neural operator layer can be expressed and approximated as
\begin{equation}
\int_{U(y)} k(x, y)\, v(x)\, dx \approx \sum_{x \in D_h \cap U(y)} k(x, y)\, v(x)\, q_x,
\end{equation}
where \( k(x, y) \) is a learned kernel and \( U(y) \) denotes a neighborhood around \( y \),  \( D_h  \) denotes a discretized mesh, and \( q_x \) are quadrature weights associated with the discretization. To ensure translation equivariance and enable efficient convolutional implementation, the kernel is often assumed to depend only on the relative displacement between points, \( k(x, y) = \kappa(x - y) \). This yields the localized convolution
\begin{equation}
\label{eq:localconv}
\mathrm{LocalConv}_\kappa[v](y) = \sum_{x \in D_h} \kappa(x - y)\, v(x)\, q_x,
\end{equation}
where \(\kappa\) has compact support defining the receptive field. This construction ensures that as the resolution increases and \(h \to 0\), the convolution converges to a local integral operator with finite support, rather than degenerating to a pointwise mapping. The resulting layer is both resolution-agnostic and naturally compatible with the continuous operator formulation.

This connects directly to the \mydef{discrete--continuous (DISCO)} convolution framework~\citep{ocampo2022scalable}, in which \(\kappa\) represents a locally supported, resolution-independent kernel shared across discretizations. In practice, the kernel can be parameterized via a small basis expansion, \begin{equation} \kappa(z) = \sum_{\ell=1}^{L} \alpha_\ell\, \varphi_\ell(z),\end{equation} where \(\{\varphi_\ell\}\) are basis functions (for instance, hat or radial basis functions) and \(\alpha_\ell\) are trainable coefficients. This parameterization allows the kernel to adapt smoothly to the underlying geometry while maintaining consistent behavior across resolutions. This framework can be generalized beyond Euclidean domains to general Lie groups~\citep{ocampo2022scalable,liu2024neural}, and paired with specialized architectures such as the Spherical FNO~\cite{bonev2023spherical}  for dynamical systems evolving on the sphere.

\begin{figure}[t!]
    \centering
    
    \includegraphics[width=0.87\linewidth]{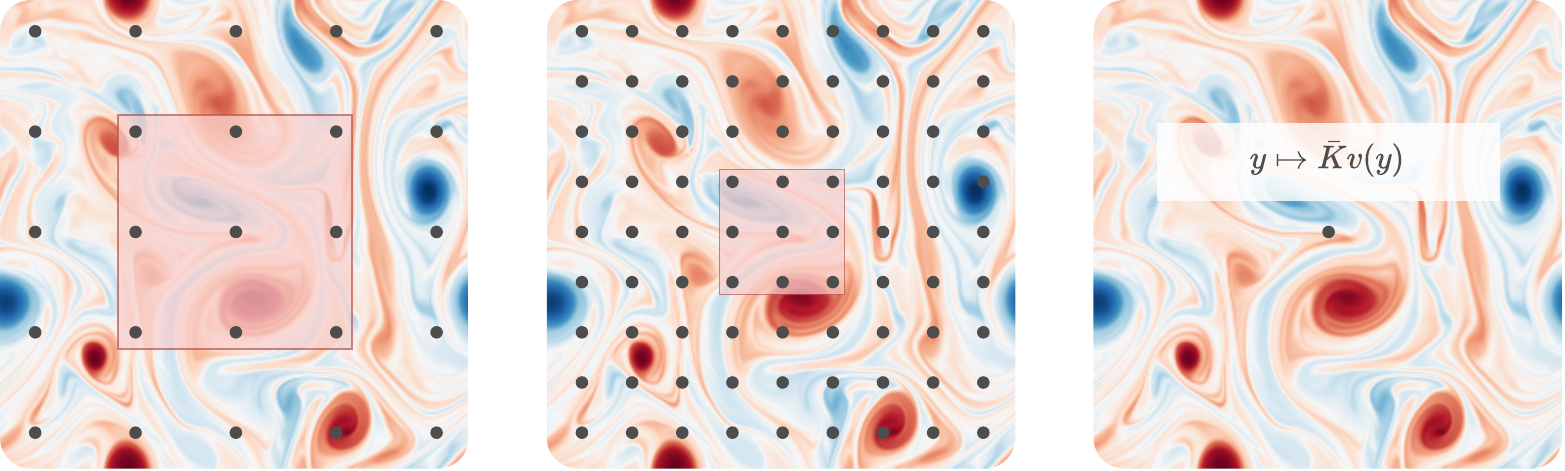}
    \\[-0em]
    {\large \textbf{Regular kernel}}
    \\[1.2em]
    
    \includegraphics[width=0.87\linewidth]{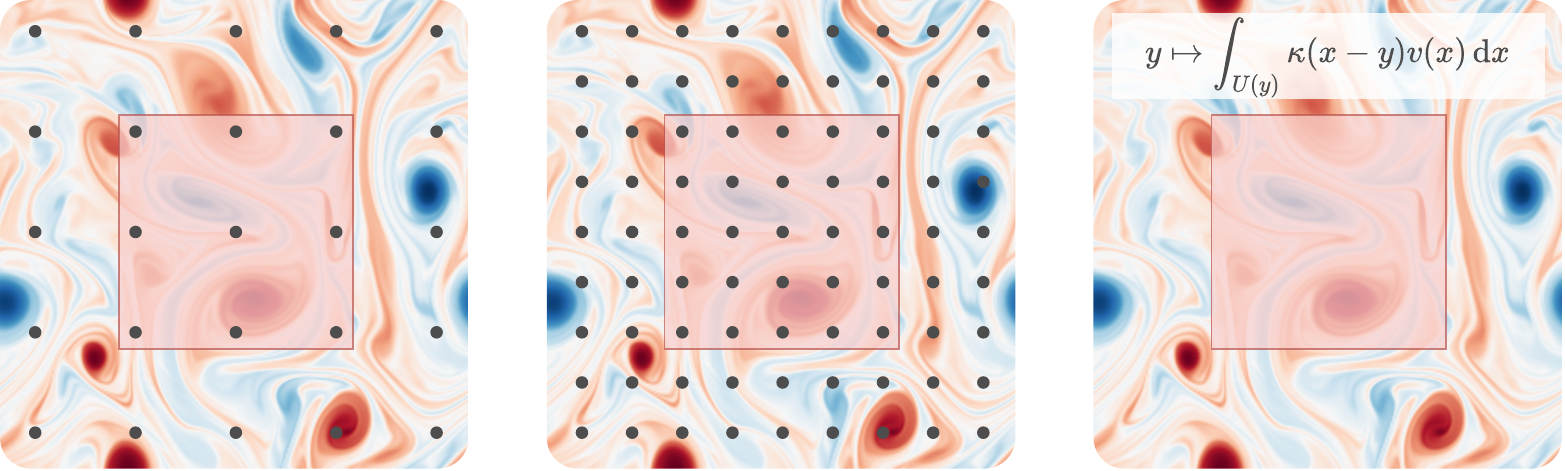}
    \\[-0em]
    {\large \textbf{Local integral kernel}}
    \\[1.2em]
    
    \includegraphics[width=0.87\linewidth]{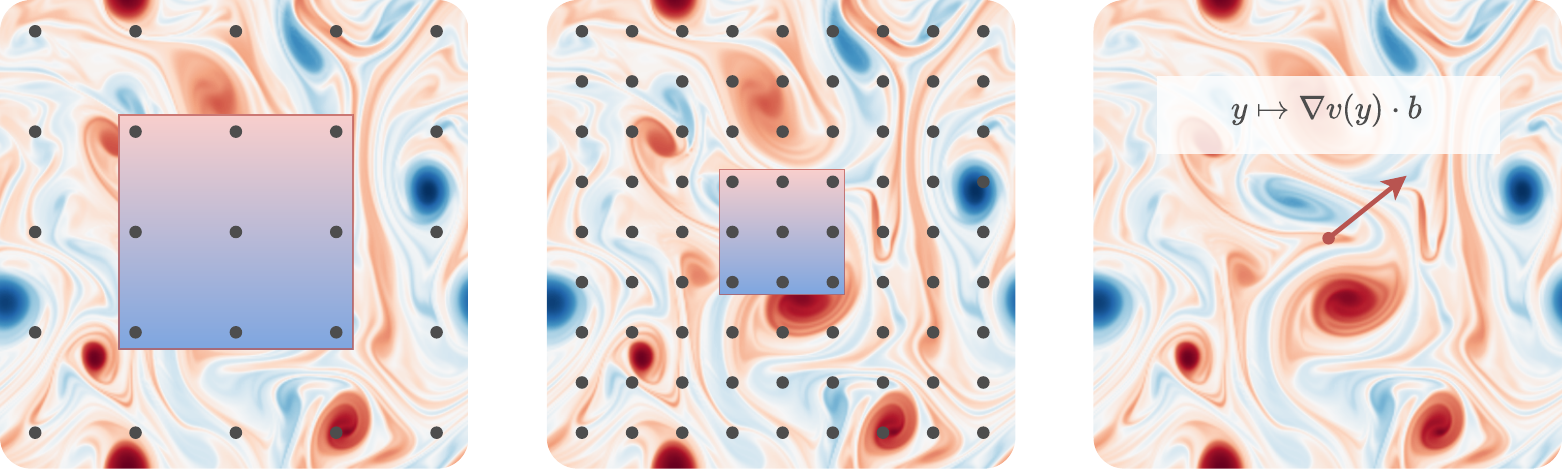}
    \\[-0em]
    {\large \textbf{Differential kernel}}
    \\[1.3em]
    
    {\Large
    $ h \hspace{1.4em} \xrightarrow{\mathmakebox[4.9em]{}} 
    \hspace{1.6em} h/2 
    \hspace{1.6em} \xrightarrow{\mathmakebox[4.9em]{}} 
    \hspace{1.4em} 0 $
    }
    \\[0em]
    
    \caption{
   Visualization of the limiting behavior of a convolution applied to a discretized function $v$ as the grid spacing $h$ is progressively refined, that is, as $h \to 0$ (figure extracted from~\citep{liu2023tipping}). 
    \textbf{(Top)} A regular convolution collapses to a pointwise linear operator. 
    \textbf{(Middle)} Using a kernel that can be evaluated at arbitrary resolutions keeps the receptive field unchanged, converging to a local integral operator. 
    \textbf{(Bottom)} Constraining the kernel while letting it collapse leads to convergence to a differential operator.}
    \label{fig:diffconv}
\end{figure}

\vspace{2mm}

\paragraph{Implementation.} These localized and integral kernel components are implemented in the \codebox{FiniteDifferenceConvolution} class from \codebox{neuralop/layers/differential\_conv.py} and in the \codebox{EquidistantDiscreteContinuousConv2d} class from \codebox{neuralop/layers/discrete\_continuous\_convolution.py} in 2D. These modules are then combined with the other components of the FNO within the \codebox{LocalNOBlocks} class in \codebox{neuralop/layers/local\_no\_block.py} and the \codebox{LocalNO} model in \codebox{neuralop/models/local\_no.py}. A tutorial displaying the effect of DISCO convolutions is also provided at \href{https://neuraloperator.github.io/dev/auto_examples/layers/plot_DISCO_convolutions.html}{neuraloperator.github.io/dev/auto\_examples/layers/plot\_DISCO\_convolutions.html}.

\begin{figure}[t]   \vspace{-1mm}
\centering   
\includegraphics[width=0.8\textwidth]{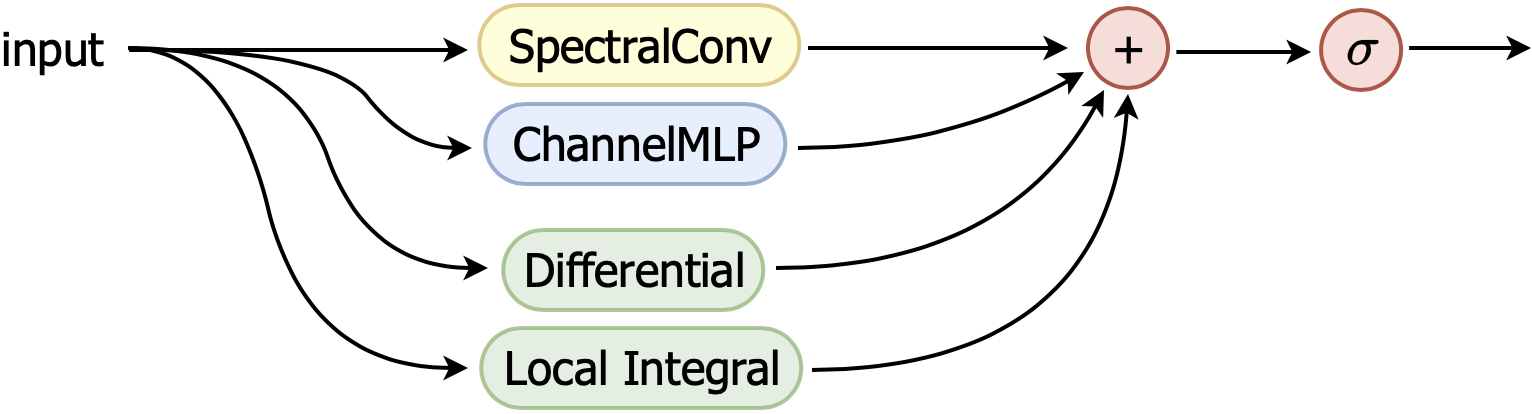}  \vspace{1mm}
\caption{Simplified diagram of the \codebox{LocalNOBlocks} forward. In addition to the \codebox{SpectralConv} and \codebox{ChannelMLP} components, the input is also passed through local integral and differential layers. In the actual \texttt{NeuralOperator 2.0.0} implementation, the forward is more sophisticated and contains additional skip connections, normalization layers, and nonlinearities, similarly to \Cref{fig: FNOBlock}.   
\label{fig: LocalNO} } 
\end{figure}

We now discuss hyperparameters specific to the \codebox{LocalNO} class in \texttt{NeuralOperator 2.0.0}. 

The \codebox{disco\_layers} parameter governs the inclusion of local integral (DISCO) kernels within the model. Each enabled layer augments the global Fourier convolution with a localized integral operator acting over a compact spatial neighborhood. The argument can be provided as a boolean or as a list of booleans of length equal to the total number of layers, allowing selective inclusion of these connections throughout the architecture. Setting \codebox{disco\_layers} to \texttt{True} for all or most layers typically enhances model expressivity on problems involving both local and nonlocal interactions, whereas selective activation supports ablation studies balancing predictive accuracy and computational cost.

The spatial extent and receptive field of the local integral kernel are controlled by \codebox{disco\_kernel\_shape} and \codebox{radius\_cutoff}. The former specifies the kernel shape (either as a single integer or as a list or tuple) while the latter defines an optional cutoff radius relative to the physical domain. Together, these parameters determine how far the local integral operator aggregates information. Smaller supports capture highly localized structure but may require deeper models to model wide-range interactions, whereas larger supports increase spatial coverage at the cost of potential over-smoothing. Structural properties of the local integral layer are further modulated by \codebox{disco\_groups} and \codebox{disco\_bias}. Grouped convolutions reduce inter-channel coupling and parameter count by assigning separate kernels to each channel group, while \codebox{disco\_bias} adds a learnable bias term to each local operator output, enhancing representational flexibility.

Analogous to \codebox{disco\_layers}, the \codebox{diff\_layers} parameter controls the inclusion of parallel differential kernel connections at each layer. These layers model localized derivative-based interactions that complement the integral kernels by introducing a differential inductive bias. The finite-difference stencil width is specified by \codebox{fin\_diff\_kernel\_size}, where smaller values yield standard central-difference approximations and larger values capture broader derivative structure. The parameter \codebox{mix\_derivatives} determines whether derivative information is shared across feature channels, enabling richer cross-channel coupling for multivariate systems. Now, \codebox{conv\_padding\_mode} sets the boundary handling scheme for all local convolutions, supporting options such as periodic, reflective, or zero padding to match the physical boundary conditions of the target problem.

It is worth noting that increasing the number of local layers, especially differential layers, can heighten sensitivity to the training resolution. When the resolution of the training data is insufficient to capture fine-scale dynamics, this often manifests as overfitting to the discretization scale and degraded performance in super-resolution tasks. In our experiments, positioning differential layers toward the earlier stages of the model, rather than near the output, alleviates this overfitting effect. Alternatively, ensuring that the training data are sampled at sufficiently high resolution to resolve the relevant small-scale structures can eliminate the problem entirely. Multi-resolution training strategies, as discussed in~\Cref{sec: ifno}, also provide an effective means of mitigating resolution-dependent artifacts and improving generalization across grid scales.

Finally, \codebox{domain\_length} encodes the domain sizes, and is used internally to compute spatial features and scale kernel operations consistently across discretizations. Proper configuration of \codebox{domain\_length} ensures geometric consistency between the learned operators and the coordinates of the physical system.

\clearpage

\subsection{Alternative and Complementary Operator Layers}
\label{sec:alternative_operator_layers}

\vspace{2mm}

Just as the localized convolutions in \Cref{sec: LocalNO} complement the Fourier spectral convolution, other operator components can provide additional ways to represent and process functions. They can run in parallel with the Fourier convolution pathway, be inserted between Fourier layers, or replace selected spectral convolutions, while retaining the surrounding channel mixing, nonlinearities, and skip connections. \Cref{fig: LocalNO} illustrates this modular construction for local integral and differential layers.

\hfill

\subsubsection{Alternative Global Transforms and Boundary-Adapted Bases}
\label{sec:alternative_global_transforms}

\vspace{2mm}

The basis expansions introduced in \Cref{sec: Basis Functions and Expansions} suggest choosing components that reflect known properties of the problem. Instead of describing every field through global sinusoidal patterns, one can use functions that satisfy prescribed boundary conditions, decay away from a localized disturbance, or represent characteristic transient responses. These choices make useful structure explicit before learning begins. Changing the basis does not automatically preserve the Fourier convolution theorem, however: the learned transformation may couple different coefficients rather than act independently on each mode.

\vspace{2mm}

\paragraph{Sine- and Cosine-Based Operators (SPFNO).}
The Semi-Periodic Fourier Neural Operator (SPFNO)~\cite{liu2023spfno} uses sine expansions for homogeneous Dirichlet conditions and cosine expansions for homogeneous Neumann conditions on intervals and tensor-product domains. Sine modes vanish at the endpoints, whereas cosine modes have vanishing endpoint derivatives, aligning the spectral representation with the corresponding boundary constraints. At the transform level, this is a simple boundary-adapted specialization of the FNO, where sine and cosine transforms can be implemented through FFTs of odd and even extensions. This offers a direct way to combine efficient Fourier transforms on uniform grids with boundary constraints.

\vspace{2mm}

\paragraph{Orthogonal Polynomial Neural Operator (OPNO).}
The OPNO~\cite{liu2024opno} combines spectral Galerkin ideas with learned transformations of polynomial coefficients. Its efficient implementation applies banded mixing to Chebyshev coefficients and reconstructs leveraging short combinations of Chebyshev polynomials satisfying homogeneous Dirichlet, Neumann, or Robin conditions. A final projection restores this boundary-adapted representation after nonlinear processing, enforcing the constraints on the output. Fast cosine transforms on Chebyshev--Gauss--Lobatto nodes make the construction practical, although uniformly sampled data require resampling for this implementation. This approach is particularly appealing for non-periodic diffusion and transport problems where boundary behavior and accurate spatial derivatives are important.

\vspace{2mm}

\paragraph{Spectral Neural Operator (SNO).}
The SNO~\cite{fanaskov2023spectral} learns nonlinear mappings directly between the coefficients of finite Fourier or Chebyshev expansions. Fourier modes describe oscillations, while Chebyshev modes describe polynomial variation of increasing degree. Rather than repeatedly alternating between physical-space values and spectral coefficients, the model processes these compact descriptions of the input and output functions. This is attractive for smooth profiles or trajectories that admit short expansions, and provides an output that can be evaluated or differentiated at new coordinates. 

\vspace{2mm}

\paragraph{Hermite Neural Operator (HNO).}
The HNO~\cite{bai2026hermite} leverages Hermite function expansions and learns transformations of their coefficients. These functions combine polynomial shapes and oscillations with a Gaussian envelope, so every finite expansion decays at infinity. This is well suited for localized dynamics on unbounded domains, including the heat diffusion and nonlinear Schr\"odinger problems studied in the paper. In particular, this representation incorporates far-field decay without imposing a boundary at a finite location.

\vspace{2mm}

\paragraph{Laplace Neural Operator (LNO).}
The LNO~\cite{cao2023lno} parameterizes its kernel in the Laplace domain through learned poles and residues. The real and imaginary parts of a pole determine an exponential growth or decay rate and an oscillation frequency, while its residue weights the corresponding response. Combining this kernel with a Fourier representation of the input produces both exponential transients and responses at the input frequencies. The layer can therefore represent short-lived disturbances alongside slower oscillatory behavior, making it a natural choice for relaxation, resonance, and driven dynamics involving several time scales.

\clearpage

\subsubsection{Piecewise and Multiresolution Representations}
\label{sec:piecewise_multiresolution_operators}

\vspace{2mm}

For shocks, interfaces, and localized disturbances, it can be helpful to represent abrupt changes or spatially concentrated activity directly. Piecewise representations provide complementary building blocks: Walsh functions describe global step-like patterns, wavelets distinguish localized features at different scales, and splines assemble functions from locally supported polynomial pieces. Their usefulness depends on how the shape, support, and smoothness of these components match the fields being modeled.

\vspace{1mm}

\paragraph{Walsh--Hadamard Neural Operator (WHNO).}
The WHNO~\cite{cavallazzi2026walsh} replaces sinusoidal modes with Walsh (step) functions, which alternate between $+1$ and $-1$ on subintervals. Their number of sign changes, or sequency, plays a role analogous to frequency. The transform forms signed combinations of field values, and learned weights modify retained low-sequency coefficients before reconstruction. These step-like patterns are attractive for discontinuous material coefficients and piecewise initial conditions. The basis functions remain global, so their distinguishing feature is their step-like shape rather than spatial localization.

\vspace{1mm}

\paragraph{Wavelet Neural Operator (WNO).}
The WNO~\cite{tripura2022wavelet} uses successive filtering and subsampling to separate a field into a coarse representation and localized details at several scales. Wavelets are shifted and rescaled patterns, so their coefficients describe both where a feature occurs and at what scale. The learned transformations act on selected subbands at the final decomposition level, and the inverse wavelet transform reconstructs the field. This organization is appealing for transported fronts, localized vortices, and evolving interfaces. The wavelet family, decomposition depth, and processed subbands determine which structures receive the most direct learned processing.

\vspace{1mm}

\paragraph{Multiwavelet-Based Operator Learning (MWT).}
The MWT~\cite{gupta2021multiwavelet} enriches the wavelet hierarchy by using several polynomial-based scaling and wavelet functions in each region, with constructions based on Legendre or Chebyshev polynomials. A coarse cell then carries a local polynomial profile rather than only an average, while detail coefficients describe what must be added under refinement. Fixed decomposition and reconstruction filters are combined with learned transformations of coarse and detail components that are reused across levels. This combines high-order local approximation with explicit scale separation, making it attractive for dispersive waves and fluid dynamics. ``Multiwavelet'' refers to the multiple basis functions within this construction, not simply to several independent wavelet transforms.

\vspace{1mm}

\paragraph{B-Spline Representations and Decoders.}
B-splines assemble functions from overlapping, locally supported polynomial pieces, with their degree and knot configuration controlling smoothness. Romagnoli et al.~\cite{romagnoli2024bspline} predict spline coefficients from an initial state to produce a continuous future trajectory, demonstrated on closed-loop quadrotor dynamics. Each coefficient influences only part of the trajectory, which can be evaluated between training times and differentiated wherever its regularity permits. This is useful for prediction and control when a continuous trajectory is more convenient than a fixed sequence of samples. The same representation principle is realized in L-SplineFNO~\citep{gopakumar2026learnablesplines}, whose analytic B-spline decoder evaluates predicted fields at arbitrary spatial locations and permits closed-form spatial differentiation.

\hfill 

\subsubsection{Structured Fourier Processing}
\label{sec:structured_fourier_processing}

\vspace{2mm}

An alternative strategy retains Fourier modes but changes the structure of their learned transformation. The following methods separate directional interactions, share transformations across frequencies, or emphasize selected frequency bands. Like the tensor compression in ~\Cref{sec: tfno}, they seek more economical or better structured spectral processing, but modify different aspects of the operation.

\vspace{1mm}

\paragraph{Factorized Fourier Neural Operator (F-FNO).}
The F-FNO~\cite{tran2023factorized} replaces a multidimensional spectral convolution with a sum of learned 1D spectral convolutions. In 2D, these operations mix information along rows and columns separately, while successive nonlinear layers couple their effects. At fixed channel width, the spectral parameter count grows with the sum rather than the product of the directional mode counts, which is attractive for higher-dimensional or higher-bandwidth problems. This factorizes the learned interactions, not merely the algorithm used to compute a multidimensional FFT. In contrast, TFNO compresses the weight tensor of a multidimensional spectral convolution through low-rank tensor decompositions.

\vspace{2mm}

\paragraph{Adaptive Fourier Neural Operator (AFNO).}
The AFNO~\cite{guibas2021adaptive} processes Fourier coefficients using nonlinear transformations within blocks of channels, with weights shared across frequencies. Soft-thresholding shrinks the transformed coefficients and sets sufficiently small components to zero, producing a sparse spectral correction. The result combines global mixing with a compact learned parameterization, which is especially useful when large spatial fields must be processed repeatedly. FourCastNet~\cite{pathak2022fourcastnet} provides a concrete dynamical application, using AFNO-based mixing for autoregressive global weather prediction.

\vspace{4mm}

\paragraph{Gabor-Filtered Fourier Neural Operator (GFNO).}
The GFNO~\cite{qi2024gabor} modulates Fourier-domain weights with learnable Gabor filters whose frequency responses emphasize selected spectral bands. These filters organize the learned transformation around characteristic oscillation scales, providing an additional way to control which parts of the spectrum receive emphasis. This is a useful design to explore when wave or flow dynamics concentrate important information in particular frequency ranges. The modification acts on spectral weights rather than introducing a separate local convolution pathway.

\hfill

\vspace{4mm}

\subsubsection{Modified Pathways and Multiscale Architectures}
\label{sec:hybrid_multiscale_operators}

\vspace{2.5mm}

Complementary components can be combined within individual blocks or organized across the network. Parallel pathways process the same latent field in different ways before combining their outputs, for example pairing global spectral communication with local corrections or input-dependent interactions. Hierarchical architectures instead move information through compact intermediate representations while retaining paths for finer detail.

\vspace{4mm}

\paragraph{U-FNO.}
The U-FNO~\cite{wen2022} adds a U-Net pathway alongside the Fourier convolution and pointwise pathways in selected blocks. The U-Net coarsens and reconstructs its input, using skip connections to preserve local detail while processing patterns over several scales. This combination is particularly well motivated by multiphase flow, where a broad pressure field coexists with sharp saturation fronts. The original U-Net pathway uses grid-based convolutions, so a resolution-consistent adaptation must also account for its physical receptive fields and resampling operations.

\vspace{4mm}

\paragraph{Wavelet--Laplace Neural Operator (WLNO).} The
WLNO~\cite{abid2026wlno} combines a Laplace pole-residue branch with a parallel, single-level Haar-wavelet branch. In two dimensions, the Haar decomposition separates local averages from horizontal, vertical, and diagonal details; independent pointwise channel transformations process these subbands before reconstruction. A trainable scalar controls the contribution of the resulting wavelet correction. This pairs a representation of characteristic transient responses with explicit access to local spatial differences.

\vspace{4mm}

\paragraph{U-Shaped Neural Operator (U-NO).} The
U-NO~\cite{rahman2022u} arranges operator layers into an encoder--decoder that contracts the latent domains while increasing channel dimensions, then reverses this process to reconstruct the output. These intermediate fields are learned representations, not cropped portions of the physical solution. Long skip connections bypass the compact intermediate stages, reducing memory demands while supporting deeper networks for large spatial and space--time prediction problems, including fluid evolution. Unlike U-FNO, the U-shaped organization belongs to the operator network itself rather than to an auxiliary pathway inside selected blocks.

\vspace{4mm}

\paragraph{Multigrid Neural Operator (MgNO).}
The MgNO~\cite{he2024mgno} parameterizes linear operator layers through multigrid cycles with learned smoothing, restriction to coarser grids, coarse corrections, and prolongation back to finer grids. The intuition comes from classical multigrid: local smoothing addresses fine-scale error, while broad error components become easier to correct on a coarse grid. Nonlinearities between these layers produce the full nonlinear model. This supplies local processing and long-range communication without Fourier transforms.

\clearpage

\subsection{FNOs on Structured Non-Euclidean Domains}
\label{sec: SFNO}

\vspace{3mm}

\subsubsection{General Setting}

\vspace{4mm}

The FNO employs the classical Fourier transform on regular Euclidean grids, where translation symmetry allows convolution to be represented as pointwise multiplication in the frequency domain. Many systems of interest evolve on domains that are not Euclidean but possess intrinsic symmetry or curvature, such as spheres or more general groups and homogeneous spaces. The FNO framework can be extended to such settings by generalizing the Fourier transform so that it respects the structure of the underlying domain.

\vspace{2mm}

A central idea is to identify a set of basis functions that capture the symmetries and periodicity of the domain. These functions play the role of the classical Fourier modes on $\mathbb{R}^{n}$ and enable a generalized Fourier transform that projects a function into a spectral representation. The definition of this transform depends on the domain, and the correspondence between convolution and multiplication holds when an appropriate notion of convolution is available and is compatible with the transform. 

\vspace{2mm}

On structured domains with group symmetry, such as compact Lie groups, these bases arise from irreducible unitary representations. The Peter--Weyl theorem guarantees that the matrix coefficients of these representations form an orthonormal basis of square-integrable functions on the group, which generalizes the Euclidean Fourier basis and captures the fundamental modes consistent with the group symmetry. The associated group Fourier transform projects functions onto these blocks, and group convolution becomes blockwise multiplication in the spectral domain. The study of harmonic analysis and representations on groups has a long and rich history, and detailed expositions can be found in~\citep{HewittRoss1970v2,Folland2016AHA,BrockerTomDieck1985,Helgason2000GGA}.

\vspace{2mm}

Symmetries play a central role in the analysis of physical systems and provide a natural organizing principle for constructing operator-learning frameworks on curved or structured domains. Formally, they can be described through group actions $\Phi: G \times M \to M,\ (g, x) \mapsto \Phi(g, x)$ acting on the configuration manifold~$M$. Respecting such actions allows for the identification of conserved quantities via Noether's theorem and the removal of redundant degrees of freedom, leading to simplified and physically consistent models. In the machine-learning context, these symmetries act as constraints that restrict the learned operators to physically meaningful submanifolds, improving robustness and regularization. This viewpoint naturally extends the notion of convolution. On a general manifold, convolution can be defined through integration over the group action, $(\kappa * u)(x) = \int \kappa(\Phi(x,y))\,u(y)\,dy$, which generalizes the translational convolution by replacing shifts with the appropriate symmetry transformations. This establishes a correspondence between generalized convolutions and harmonic transforms that respect the manifold's intrinsic structure.

\vspace{2mm}

Operating in a symmetry-aligned spectral representation yields two practical benefits. First, learned operators remain consistent with intrinsic symmetries, which supports equivariance or invariance to the relevant group actions. Second, the representation supports resolution independence because spectral coefficients correspond to intrinsic modes rather than discretization-dependent samples. Efficiency and scalability depend on available fast transforms and accurate quadrature for the domain of interest. On the sphere there exist fast spherical harmonic transforms, which enable practical spectral learning at substantial bandwidths. For selected compact groups there are generalized fast Fourier transforms that reduce cost compared with direct transforms. On general meshes or arbitrary manifolds such fast transforms are typically unavailable, and building spectral bases may require large computations, which can dominate the cost and limit scalability.

\hfill 

Overall, FNOs can be extended to structured non-Euclidean domains by replacing the classical Fourier basis with appropriate harmonic bases. This preserves symmetry and supports efficient global coupling when fast transforms are available. We now present the unit sphere as a concrete example, where spherical harmonics provide the natural orthonormal basis for a Spherical Fourier Neural Operator~\citep{bonev2023spherical}.

\clearpage

\subsubsection{The Spherical Fourier Neural Operator (SFNO)}
\label{sec: SFNO_def}

\vspace{2mm}

The unit sphere $\mathbb{S}^{2} = \{ x \in \mathbb{R}^{3} : \|x\| = 1 \}$ is a simple example of a compact non-Euclidean manifold with intrinsic rotational symmetry. Its geometry is characterized by the group of rotations $\mathrm{SO}(3)$ acting transitively on its points. Each rotation moves a point on the sphere without changing distances or angles, preserving the surface's curvature and intrinsic structure. In contrast to Euclidean domains, translations are not defined globally on the sphere. On the sphere, the natural transformations are rotations, which form a continuous compact group. This difference motivates a generalization of Fourier analysis that replaces the notion of spatial shifts with group actions, allowing functions defined on $\mathbb{S}^{2}$ to be analyzed through their spectral components aligned with spherical symmetry.

\begin{figure}[t]  \vspace{-4mm}
\centering   
\includegraphics[width=0.9\textwidth]{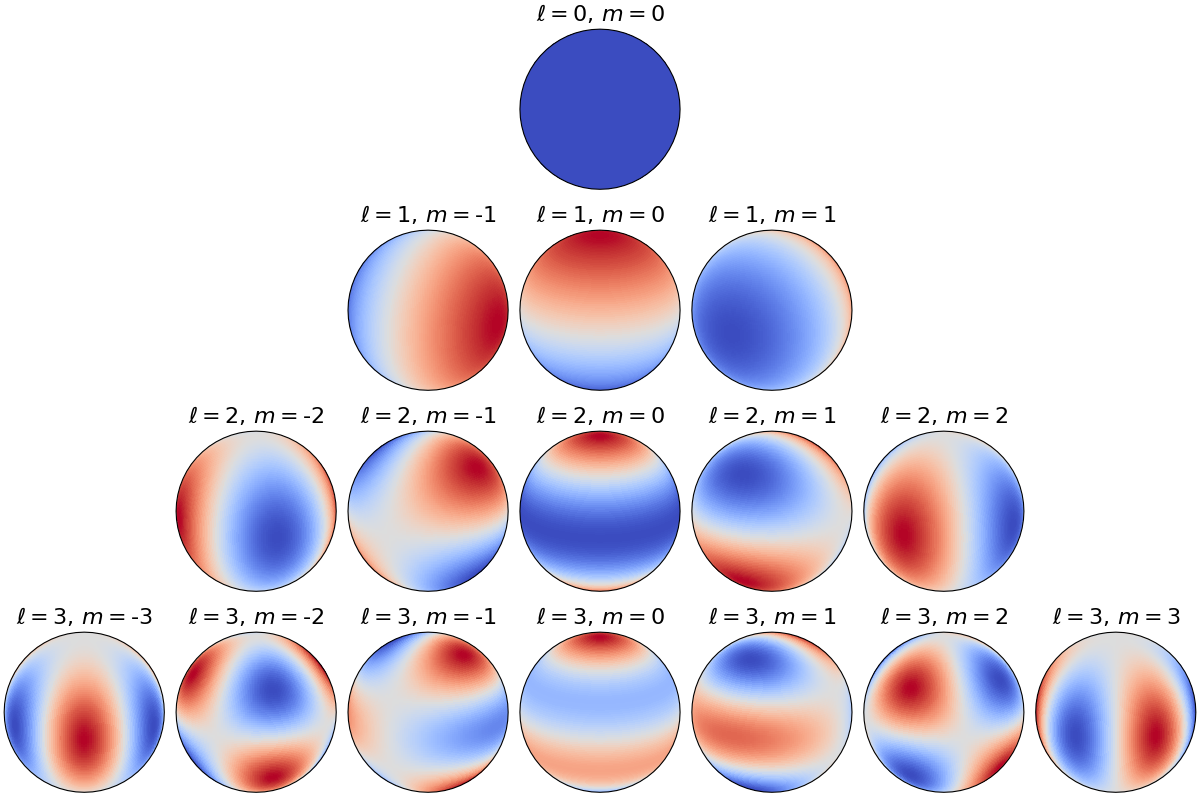}  \vspace{-1mm}
\caption{Examples of spherical harmonics $Y_{\ell}^{m}$, obtained using \texttt{torch-harmonics}.   
\label{fig: harmonics} }  \vspace{5mm}
\end{figure}

\paragraph{Basis functions.} \mydef{Spherical harmonics} provide the natural analogue of complex exponentials on $\mathbb{R}^{n}$ for the sphere. They form an orthonormal basis $\{Y_{\ell}^{m}\}$ for square-integrable functions on $\mathbb{S}^{2}$, where the integer degree $\ell \ge 0$ controls angular frequency and the order $m \in [-\ell, \ell]$ enumerates modes within each band. \Cref{fig: harmonics} displays the first few spherical harmonics.

\paragraph{Fourier transform.} Recall that for a sufficiently well-behaved function $f$ on $\R^d$, the continuous Fourier transform is defined as
\begin{equation}
\hat{f}(\xi) = \frac{1}{(2\pi)^{d/2}} \int_{\R^d} f(x) \, e^{-i \, x \cdot \xi} \, dx,
\end{equation}
where $\xi \in \R^d$ denotes the frequency variable. This can be adapted to $\mathbb{S}^{2}$ using the spherical harmonics. The \mydef{Spherical Harmonic Transform} $\mathcal{F}_{\mathbb{S}^{2}}$ projects a square-integrable function $u $ on $\mathbb{S}^{2}$ onto this basis via
\begin{equation}
\hat{u}(\ell,m)=\int_{\mathbb{S}^{2}}  u(\Omega) \ \overline{Y_{\ell}^{m}(\Omega)} \ d\Omega,
\end{equation}
with volume form $d\Omega=\sin\theta\,d\theta\,d\phi$ on $\mathbb{S}^{2}$. This generalizes the classical Fourier transform, and aligns frequencies with spherical symmetry. 

The inverse spherical harmonic transform can then be obtained via
\begin{equation}
u(\theta,\phi)=\sum_{\ell\in\mathbb{N}}\ \sum_{|m|\le \ell}\ \hat{u}(\ell,m)\,Y_{\ell}^{m}(\theta,\phi).
\end{equation}

\vspace{3mm}

\paragraph{Convolution.} Recall that for functions $f$ and $g$ on $ \R^d$, their convolution is defined by
\begin{equation}
(f * g)(x) = \int_{\R^d} f(y) \, g(x - y) \, dy.
\end{equation}
Convolution expresses how one function interacts with another by sliding across it, multiplying where they overlap, and integrating the result. It describes how one function smooths, filters, or modulates another.  The \mydef{spherical convolution} can be defined by replacing translations by $y$ with rotations by $R \in \mathrm{SO}(3)$, as
\begin{equation}
(\kappa \star u)(x)=\int_{R\in \mathrm{SO}(3)} u(Rn)\,\kappa(R^{-1}x)\,dR,\quad \forall x\in \mathbb{S}^{2},
\label{eq:sph-conv}
\end{equation}
where $n=(0,0,1)^{\top}$ is the north pole. 

\vspace{1mm}

\paragraph{Convolution Theorem.} Analogously to the classical $\widehat{f * g} = \hat{f} \, \hat{g}$ convolution theorem, the spherical harmonic transform and spherical convolution admit the convolution theorem\begin{equation}
\mathcal{F}_{\mathbb{S}^{2}}[\kappa \star u](\ell,m)
=
c_\ell \cdot  \mathcal{F}_{\mathbb{S}^{2}}[u](\ell,m) \cdot \mathcal{F}_{\mathbb{S}^{2}}[\kappa](\ell,0),
\label{eq:sph-conv-theorem}
\end{equation}
for some constant $c_\ell$ depending on $\ell$ only. 

\vspace{3mm}

\textbf{The Spherical FNO.} The \mydef{Spherical Fourier Neural Operator (SFNO)}~\citep{bonev2023spherical} is defined analogously to the FNO, except that the Fourier integral operator is now expressed using the spherical harmonic transform, i.e.  $\mathcal{K}[u](x) =\mathcal{F}^{-1}\left(  R_\theta\cdot \mathcal{F}(u) \right)(x)$ is replaced by 
\begin{equation}
\mathcal{K}[u](x) =\mathcal{F}_{\mathbb{S}^{2}}^{-1}\left[R_\theta \cdot \mathcal{F}_{\mathbb{S}^{2}}[u] \right](x).
\end{equation}

\vspace{3mm}

\paragraph{Implementation.} The Spherical FNO is implemented in \texttt{NeuralOperator 2.0.0} as the \codebox{SFNO} class from \codebox{neuralop/models/sfno.py}, by replacing the \codebox{SpectralConv} class of the FNO with the \codebox{SphericalConv} class from \codebox{neuralop/layers/spherical\_convolution.py}. The implementation relies on the \texttt{torch-harmonics} library, which provides GPU-optimized and differentiable \texttt{PyTorch} implementations of spherical harmonic transforms, spherical convolutions, and related tools for learning on spherical geometries. In particular, the spherical harmonic transform adopts the ``direct'' algorithm~\citep{schaeffer2013}, which is well suited to parallel computation on modern hardware, and supports backpropagation through the spectral coefficients. For a more advanced and flexible implementation of SFNO, as well as of spherical harmonic and spherical-geometry operations beyond the \texttt{NeuralOperator} interface, we refer the reader to the \texttt{torch-harmonics} package, available at \href{https://github.com/NVIDIA/torch-harmonics}{github.com/NVIDIA/torch-harmonics}. A tutorial on training an SFNO to solve the spherical shallow-water equations is available at \href{https://neuraloperator.github.io/dev/auto_examples/models/plot_SFNO_swe.html}{neuraloperator.github.io/dev/auto\_examples/models/plot\_SFNO\_swe.html}.

\hfill

\subsection{Integrating FNOs in Geometric Encoder-Decoder Structures for General Point Clouds} \label{sec: Point Clouds}

\vspace{1.5mm}

\subsubsection{Motivation}

\vspace{2mm}

Many physical systems in science and engineering occur in geometrically complex environments where structured grids are inefficient or unavailable. Examples include fluid flow around vehicles and airfoils, stress fields in composite materials, and wave propagation in heterogeneous geological media. In such cases, PDEs must be solved on unstructured or adaptive meshes to accurately resolve fine geometric details and boundary effects. Uniform discretizations often waste computation on regions with smooth solutions, while unstructured grids allow for efficient allocation of resolution where it is most needed.

FNOs have achieved remarkable success as data-driven surrogates for PDE solution operators, owing to their use of global convolutions implemented through the Fast Fourier Transform (see \Cref{sec: DFT} and \Cref{sec: fft}). This spectral representation yields excellent computational efficiency and global receptive fields on regular domains. However, the reliance on the FFT confines the efficient implementation of FNOs from \texttt{NeuralOperator 2.0.0} to rectangular domains with uniform sampling. When applied to irregular geometries, direct embedding into a rectangular grid introduces inefficiencies and boundary distortions, while interpolating between non-uniform and uniform meshes leads to significant numerical errors, especially in nonlinear systems. Furthermore, discrete Fourier transforms on irregular meshes are computationally expensive and lose discretization convergence, since the underlying sampling measure no longer approximates the continuous Fourier integral accurately. These limitations restrict the application of FNOs to many real-world physical systems where the computational domain is irregular or varies with design parameters. Extending neural operators to such domains requires architectures that incorporate geometric awareness while preserving the spectral efficiency of the Fourier representation. 

A widely adopted approach to overcome these geometric constraints is to use a geometric encoder-decoder architecture that connects irregular physical domains with regular computational grids. In this framework, the encoder maps the input geometry, which may be represented as a mesh, point cloud, or signed distance function, into a latent space equipped with a uniform grid. The FNO is then applied efficiently in this latent space, leveraging the FFT to capture global dependencies and nonlocal interactions with high expressivity and computational efficiency. The decoder subsequently projects the latent representation back to the physical domain, reconstructing the solution field on the original irregular geometry. This design effectively combines the geometric expressivity and spatial adaptivity of geometry-based encoder-decoder architectures with the spectral efficiency and representational power of the FNO. By jointly learning the deformation and operator dynamics in an end-to-end manner, such models achieve both geometric generality and spectral precision, providing an efficient and accurate framework for solving PDEs on complex, unstructured domains.

\vspace{1mm} 

\hfill 

\subsubsection{The Geometry-Aware Fourier Neural Operator (Geo-FNO)} \label{sec: Geo-FNO}

\vspace{2mm}

\begin{figure}[t]  
\centering   
\includegraphics[width=\textwidth]{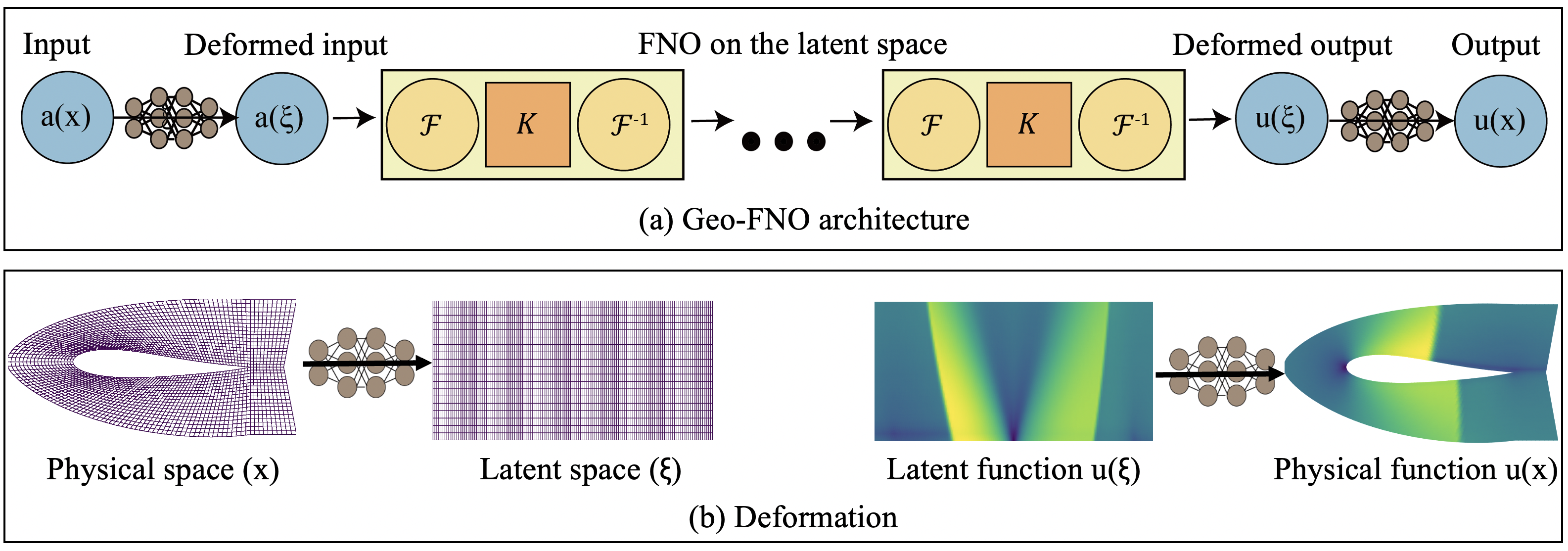}  \vspace{-6mm}
\caption{The Geometry-Aware Fourier Neural Operator (Geo-FNO) architecture (extracted from~\citep{li2023fourier}).   
\label{fig: Geo-FNO} }  \vspace{3mm}
\end{figure}

The \mydef{Geometry-Aware Fourier Neural Operator (Geo-FNO)}~\citep{li2023fourier} provides a concrete realization of the geometric encoder-decoder paradigm for extending FNOs to irregular and unstructured domains. Geo-FNO proceeds as depicted in \Cref{fig: Geo-FNO} by learning a smooth deformation map $\varphi_a$ that encodes an irregular physical domain into a regular latent grid where the FNO can be applied efficiently using FFTs. The resulting field is then decoded back to the physical geometry through the inverse mapping $\varphi_a^{-1}$.

\vspace{1mm} 

More precisely, Geo-FNO introduces a deformation map $\varphi_a : D_c \to D_a$, where \(D_a \subset \mathbb{R}^d\) is the physical domain associated with the geometry parameter \(a\), and \(D_c = [0,1]^d\) is a fixed computational domain equipped with a uniform mesh. The deformation \(\varphi_a\), assumed to be a diffeomorphism (both \(\varphi_a\) and its inverse \(\varphi_a^{-1}\) are smooth and invertible), defines a correspondence between points \(x \in D_a\) and coordinates \(\xi \in D_c\), allowing the function \(u : D_a \to \mathbb{R}^m\) to be pulled back to the computational space as $u_c(\xi) = u(\varphi_a(\xi))$. The deformation \(\varphi_a\) acts as a geometric encoder that embeds the irregular physical geometry into a regular computational domain where the FNO can be applied efficiently using the FFT. It can be prescribed analytically for structured meshes or parameterized via \(\varphi_a^{-1}(x) = x + f(x,a) \), where \(f\) is a neural network that learns geometry-dependent coordinate transformations. 

\vspace{1mm} 

Geo-FNO then relies on a geometric Fourier transform that generalizes the standard spectral transform to deformed domains. The forward and inverse geometric Fourier transforms are defined respectively using transformation of coordinates formulas as
\begin{equation}
(\mathcal{F}_a u)(k) = \int_{D_c} u_c(\xi) e^{-2\pi i \langle \xi, k \rangle} d\xi
= \int_{D_a} u(x) e^{-2\pi i \langle \varphi_a^{-1}(x), k \rangle} \left| \det \nabla_x \varphi_a^{-1}(x) \right| dx,
\end{equation}
and
\begin{equation}
(\mathcal{F}_a^{-1} \hat{u})(x) = \sum_k \hat{u}(k) e^{2\pi i \langle \varphi_a^{-1}(x), k \rangle}.
\end{equation}

These transforms effectively pull back the function from the physical domain to the latent computational domain, perform the standard Fourier operations in the latent grid, and push the result forward to the physical space. In practice, both transforms are implemented in their discrete form using sums over mesh points, which raises the computational cost from the \(\mathcal{O}(N \log N)\) scaling of the FFT to \(\mathcal{O}(N^2)\).

The deformation map is used only for the geometric Fourier transform in the first layer and the inverse geometric Fourier transform in the last layer. All intermediate layers operate entirely within the uniform latent grid using standard FFT-based spectral convolutions. This design maintains the efficiency and global receptive field of the original FNO while extending it to arbitrary, unstructured geometries. After processing in the latent domain, the inverse deformation maps the predicted field back to the physical coordinates, completing the geometric decoder step. 

Through this construction, Geo-FNO combines the spectral expressivity and global modeling capacity of the FNO with the geometric flexibility of deformation-based methods. The learned deformation aligns the latent representation with the underlying geometry, allowing the model to concentrate computational resolution in regions of high physical variation. As a result, Geo-FNO achieves accurate and discretization-consistent operator learning across diverse geometries. \\

\subsubsection{The Geometry-Informed Neural Operator (GINO)} \label{sec: GINO}

\vspace{4mm}

The \mydef{Graph Neural Operator (GNO)}~\citep{li2020neural} implements kernel integration with graph structures and is applicable to complex geometries and irregular grids. The GNO kernel integration shares similarities with the message-passing implementation of graph neural networks (GNNs)~\citep{battaglia2016interaction}. However, GNO defines the graph connectivity and vertices using a ball or neighborhood in continuous physical space, while  GNN assumes a fixed set of neighbors in a discrete graph. The GNN nearest-neighbor connectivity violates discretization convergence and degenerates into a pointwise operator at high resolutions, leading to a poor approximation of the operator. In contrast, GNO adapts the graph based on points within a physical space, allowing for universal approximation of operators. 

Specifically, the GNO acts on an input function $v$ as follows,
\begin{align}
\label{eq:gno}
 \mathcal{G}_\mathrm{GNO}(v)(x) & \coloneqq \int_D \mathds{1}_{\text{B}_r(x)}(y) \ \kappa(x,y)\ v(y) \, \mathrm{d}y,
\end{align}
where $D\subset\mathbb{R}^d$ is the domain of $v$, $\kappa$ is a learnable kernel function, and  $\mathds{1}_{\text{B}_r(x)}$ is the indicator function over the ball $\text{B}_r(x)$ of radius $r>0$ centered at $x\in D$. The radius $r$ is a hyperparameter, and the integral can be approximated with a Riemann sum, for instance.

\hfill

\begin{figure}[t]  \vspace{0mm}
\centering   
\includegraphics[width=\textwidth]{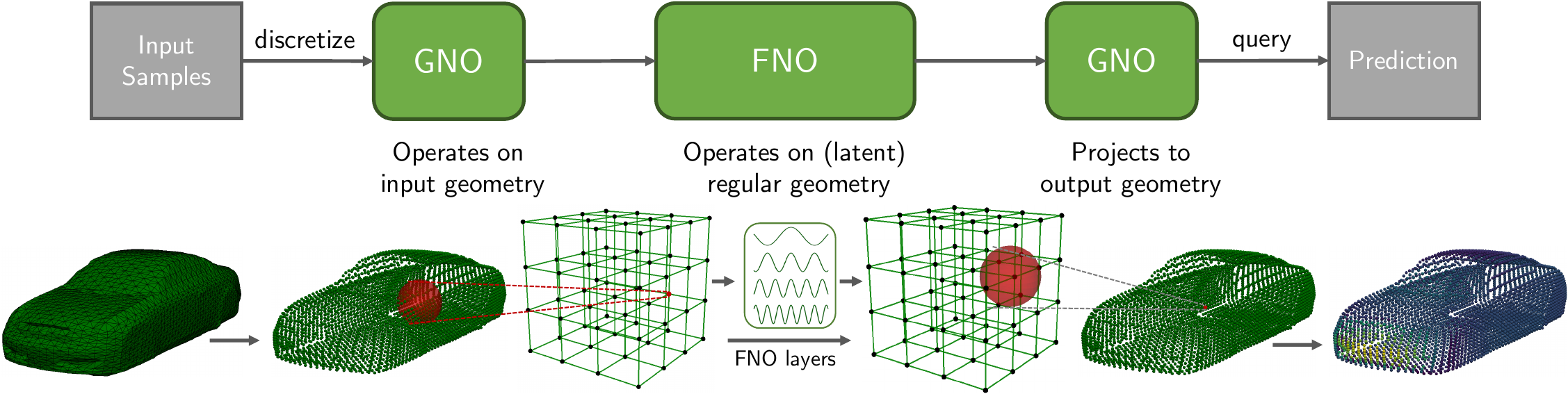}  \vspace{-2mm}
\caption{The Geometry-Informed Neural Operator (GINO) architecture (extracted from~\citep{li2024geometry}).   
\label{fig: GINO} }  \vspace{4mm}
\end{figure}

The \mydef{Geometry-Informed Neural Operator (GINO)}~\citep{li2024geometry} provides another concrete realization of the geometric encoder-decoder paradigm for extending FNOs to irregular and unstructured domains, by combining an FNO with GNOs, as depicted in~\Cref{fig: GINO}. More precisely, the input is passed through three main neural operators,
\begin{equation}
     \mathcal{G}_{\mathrm{GINO}} \ = \  \mathcal{G}_{\mathrm{GNO}}^{decoder} \ \circ \  \mathcal{G}_{\mathrm{FNO}}
   \  \circ \  \mathcal{G}_{\mathrm{GNO}}^{encoder} .
 \end{equation}
 
First, a GNO encodes the input given on an arbitrary geometry into a latent space with a regular geometry. The encoded input can be concatenated with a signed distance function evaluated on the same grid if available. Then, an FNO is used as a mapping on that latent space for efficient global integration. Finally, a GNO decodes the output of the FNO by projecting that latent space representation to the output geometry. As a result, GINO can represent mappings from one complex geometry to another. \\

\paragraph{Implementation.} The GNO and GINO architectures are available in \texttt{NeuralOperator 2.0.0} through the \codebox{GNOBlock} and \codebox{GINO} classes in \codebox{neuralop/layers/gno\_block.py} and \codebox{neuralop/models/gino.py}. 

The specification of the neighbor search radii, \codebox{in\_gno\_radius} and \codebox{out\_gno\_radius}, controls the spatial extent of interaction in the input and output graph domains. These radii define the size of the receptive field for each node, thereby determining the level of nonlocality. Larger radii enable to integrate information from more neighboring points, which improves the modeling of smooth global dependencies, but also increases computational costs. Conversely, smaller radii promote locality and efficiency but may limit expressivity. The choice of weighting function \codebox{gno\_weighting\_function}, further refines how neighboring points contribute to the aggregation process. This is based on the mollified GNO ($m$GNO)~\citep{lin2025mGNO} extension of the GNO which replaces the indicator function in GNO by a smooth decaying weighting function to allow for differentiability.

Within each geometric layer, the feature transformations are governed by \codebox{in\_gno\_transform\_type} and \codebox{out\_gno\_transform\_type}, which specify whether the integral transformation applied to the graph messages is linear or nonlinear. The linear setting corresponds to an affine aggregation, providing computational simplicity and numerical stability, whereas the nonlinear and kernel-only variants introduce activation functions that increase expressivity and allow the model to approximate more complex relationships. The expressivity of GINO is further shaped by the channel-wise MLP embedded in its graph layers and its parameters \codebox{in\_gno\_channel\_mlp\_hidden\_layers} and \codebox{out\_gno\_channel\_mlp\_hidden\_layers} defining the number of hidden layers used for feature mixing at each node. Deeper or wider configurations increase representational power but also raise computational costs. 

Finally, two implementation-related parameters, \codebox{gno\_use\_open3d} and \codebox{gno\_use\_torch\_scatter}, directly impact computational efficiency. When enabled, \texttt{Open3D}~\cite{Zhou2018} accelerates neighbor searches through optimized spatial indexing, which is particularly beneficial for large meshes or three-dimensional geometries. Similarly, \texttt{torch-scatter}~\cite{pytorch-scatter} enables efficient parallel aggregation of messages during the integral transform step, providing substantial speedups on modern GPUs. We recommend keeping both options active unless compatibility issues arise, as their deactivation can significantly slow down training and inference.

\hfill

\subsubsection{The Optimal Transport Neural Operator (OTNO)} \label{sec: OTNO}

\vspace{3mm}

The \mydef{Optimal Transport Neural Operator (OTNO)}~\cite{li2025geometricoperatorlearningoptimal} offers yet another principled framework for learning solution operators of PDEs defined on complex geometries. Its central insight is to recast geometric embedding as an optimal transport (OT) problem, transforming irregular domains into simple, uniform latent spaces while preserving their essential geometric structure. 

At a conceptual level, OTNO treats each geometric instance as a continuous density function defined over a manifold embedded in Euclidean space. The geometry is then encoded by solving an OT problem that maps this density to a uniform reference density in a regular latent domain (e.g. a sphere or torus). Within the latent space, efficient FNOs or Spherical FNOs can be leveraged with their fast FFT-based spectral convolutions. A notable feature of OTNO is that the latent space can have a lower intrinsic dimension than the physical domain itself. For example, two-dimensional surfaces in three-dimensional spaces can be mapped onto two-dimensional latent manifolds where computations are carried out, substantially improving efficiency by avoiding unnecessary volumetric representation while preserving the essential geometric and physical structure of the problem.

The transport can be formulated as a \emph{map} (Monge formulation) or as a \emph{plan} (Kantorovich formulation). The Monge map produces a bijective, continuous deformation that directly pushes points from the latent to the physical domain while the Kantorovich plan provides a probabilistic coupling that softly aligns the two domains. Both formulations enable smooth geometric correspondences without resorting to explicit interpolation or shared graph connectivity. This procedure constructs an instance-specific transformation that regularizes geometric complexity while preserving the underlying measure. Unlike Geo-FNO, which uses a shared global deformation map, OTNO performs a distinct transport per instance, yielding adaptive and faithful embeddings across a wide range of shapes. Finally, the OT decoder reverses the process, transporting the latent predictions back to the original surface using the same OT correspondence. In the Monge case, decoding involves applying the inverse of the transport map, while in the Kantorovich case, it corresponds to averaging with respect to the learned transport plan.

\begin{figure}[t]  \vspace{8mm}
\centering   
\includegraphics[width=\textwidth]{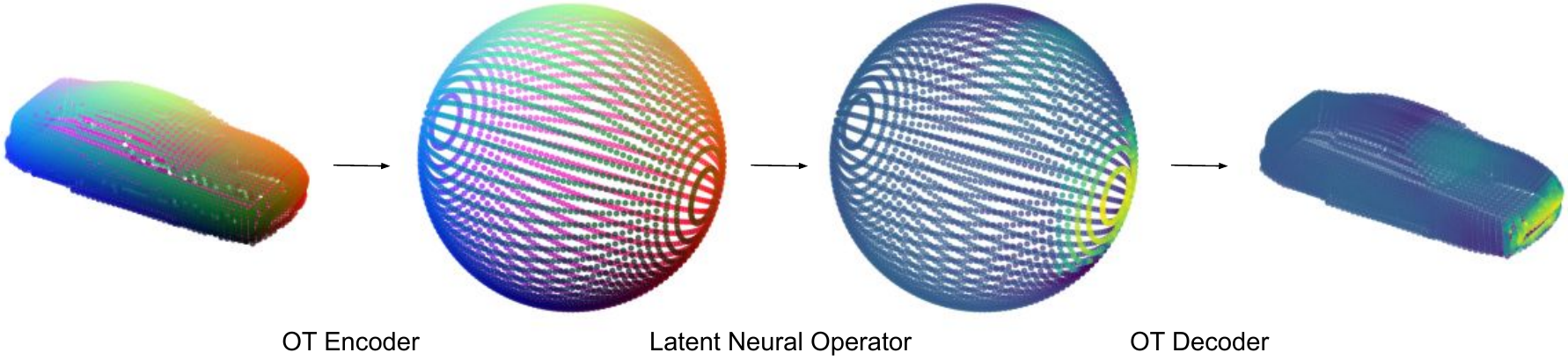}  \vspace{-3mm}
\caption{The Optimal Transport Neural Operator (OTNO) architecture (extracted from~\cite{li2025geometricoperatorlearningoptimal}).   
\label{fig: OTNO} }  \vspace{6mm} 
\end{figure}

The OTNO architecture will be made available in a future release of \texttt{NeuralOperator} through the \codebox{OTNO} class in \codebox{neuralop/models/otno.py}.

\hfill \\

\section*{Conclusion} 
\addcontentsline{toc}{section}{Conclusion}

\vspace{2mm}

The Fourier Neural Operator has established itself as a foundational architecture for learning mappings between function spaces, providing an efficient and theoretically principled framework for modeling complex physical systems. By leveraging the spectral representation of functions, it achieves global expressivity, computational efficiency, and resolution invariance that distinguish it from traditional neural networks and domain-specific surrogates. However, its effectiveness in practice depends critically on a sound understanding of its theoretical underpinnings and careful attention to implementation details.

In this work, we have provided a comprehensive and practical guide to FNOs, bridging the gap between mathematical foundations and computational realization. We have clarified the operator-theoretic and signal-processing principles that give rise to their spectral formulation, explained the design and behavior of each architectural component, and discussed how these insights translate into practical implementation within the \texttt{NeuralOperator 2.0.0} framework. By emphasizing clarity, intuition, and reproducibility, this guide seeks to enable practitioners to apply FNOs effectively, interpret their behavior confidently, and extend them to new scientific and engineering challenges.

Looking forward, the principles outlined here can inform the development of more advanced operator-learning architectures, including multi-resolution, adaptive, and physics-informed variants. As the field continues to evolve, a rigorous understanding of spectral methods and their computational realizations will remain essential to unlocking the full potential of neural operators in data-driven scientific computing.

\hfill \\

\section*{Acknowledgments}
\addcontentsline{toc}{section}{Acknowledgments}
A. Anandkumar is supported in part by Bren endowed chair, ONR (MURI grant N00014-18-12624), and by the AI2050 senior fellow program at Schmidt Sciences.

We also thank all who provided valuable feedback on this manuscript, with special thanks to Maurice Hanisch.

\newpage 

\bibliography{bib}
\bibliographystyle{style}

\end{document}